\documentclass[runningheads]{llncs}

\usepackage{eccv}

\usepackage{xcolor}
\usepackage{colortbl}
\usepackage{multirow}

\usepackage{soul}
\usepackage{pifont}

\definecolor{gold}{rgb}{1.0, 0.874, 0}
\definecolor{silver}{rgb}{0.77,0.77,0.77}
\definecolor{brown}{rgb}{0.95, 0.678, 0.4}

\newcommand{\xmark}{\ding{55}}%

\definecolor{firstcolor}{rgb}{1, 0.6, 0.6}
\definecolor{secondcolor}{rgb}{1, 0.8, 0.6}

\usepackage{eccvabbrv}

\usepackage{graphicx}
\usepackage{booktabs}
\usepackage{multirow}
\usepackage[accsupp]{axessibility}  

\usepackage[percent]{overpic}
\usepackage{pifont}
\usepackage{algorithm}
\usepackage{algpseudocode}
\usepackage{multirow}
\usepackage{float}
\usepackage{array}
\usepackage{tikz}
\usepackage{bm}
\usepackage{bbding}
\usepackage{pifont}
\usepackage{wasysym}
\usepackage{amssymb}
\usepackage{pifont}
\usepackage{tabularx}
\usepackage{booktabs}
\usepackage[export]{adjustbox}
\usepackage{enumitem}
\usepackage{gensymb}
\usepackage{xcolor}
\usepackage{arydshln}
\definecolor{higher}{HTML}{9FC4D9} 
\definecolor{lower}{HTML}{f9b0c7}
\definecolor{salmon}{HTML}{EDF4F2}
\definecolor{mygreen}{RGB}{0,128,0}

\usepackage{hyperref}

\usepackage{orcidlink}

\begin{document}

\title{Diffusion Models for Monocular Depth Estimation: Overcoming Challenging Conditions} 

\titlerunning{Overcoming Depth Challenges with Diffusion Models}

\author{Fabio Tosi \orcidlink{0000-0002-6276-5282} \and
Pierluigi Zama Ramirez \orcidlink{0000-0001-7734-5064} \and
Matteo Poggi \orcidlink{0000-0002-3337-2236}}

\authorrunning{F.~Tosi et al.}

\institute{University of Bologna, Bologna, Italy \\
\email{\{fabio.tosi5, pierluigi.zama, m.poggi\}@unibo.it} \\
\url{https://diffusion4robustdepth.github.io/}}

\maketitle

\begin{figure}
    \centering
    \renewcommand{\tabcolsep}{1pt}
    \begin{tabular}{ccccc}
        \scriptsize nuScenes\cite{caesar2020nuscenes} 
        & \scriptsize DrivingStereo \cite{yang2019drivingstereo} 
        & \scriptsize Booster\cite{zamaramirez2022booster} 
        & \scriptsize ClearGrasp\cite{sajjan2020clear} 
        & \scriptsize Trans10K\cite{xie2020segmenting} 
        \\
        
        \includegraphics[trim=2.2cm 0cm 2.1cm 0cm,clip,height=0.14\textwidth]{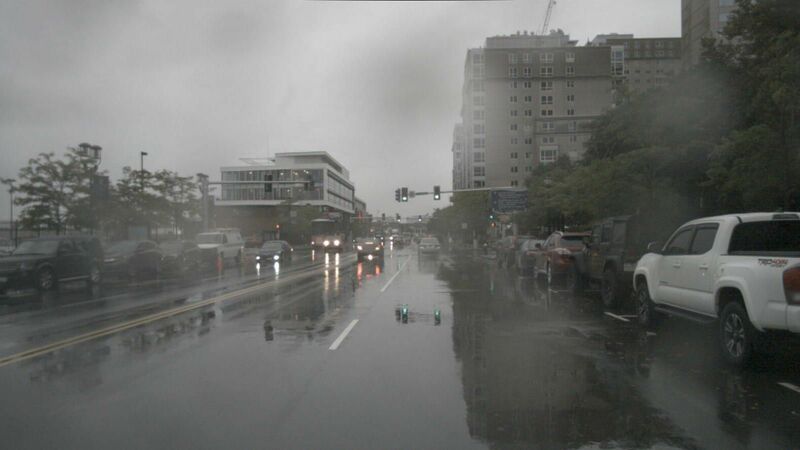} &
        \includegraphics[trim=4.4cm 0cm 12.2cm 0cm,clip,height=0.14\textwidth]{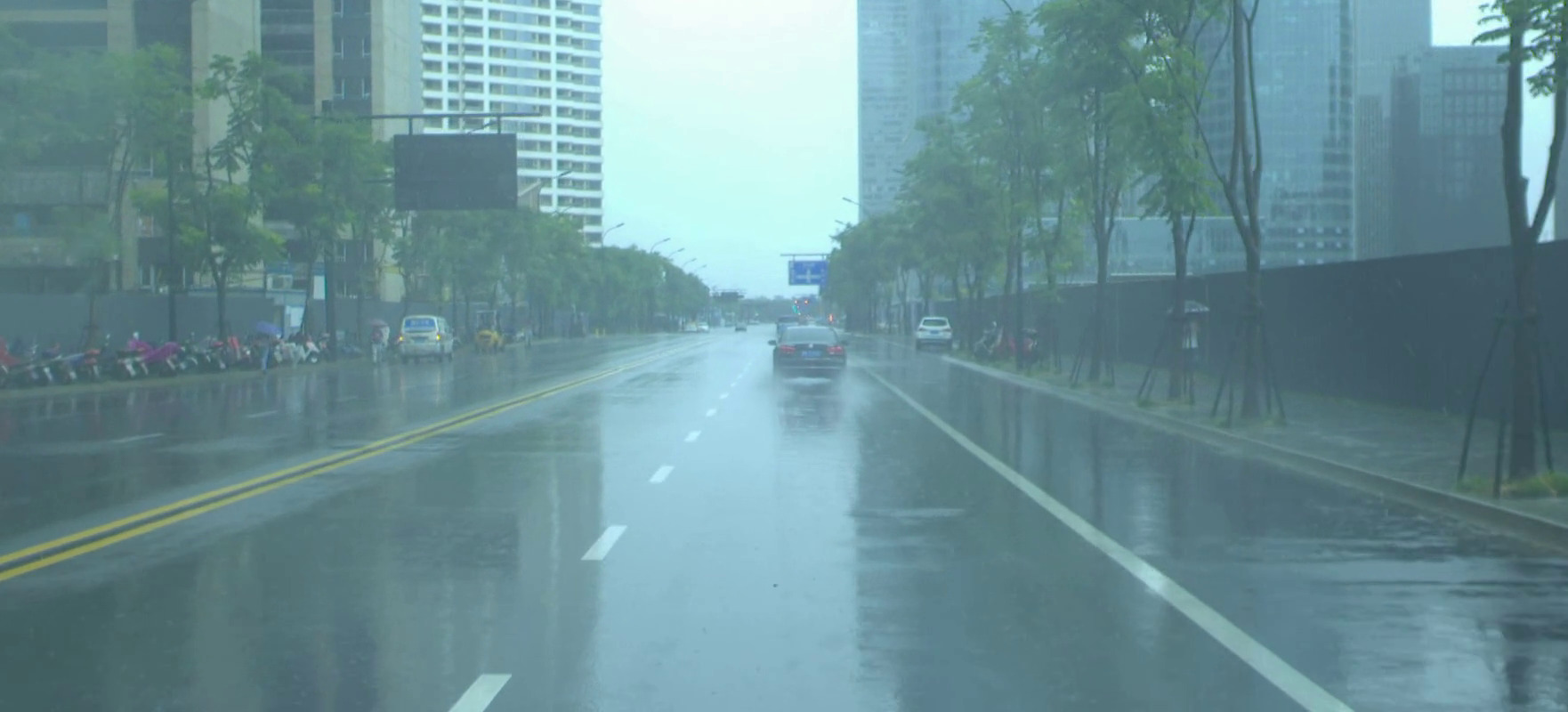} &
        \includegraphics[height=0.14\textwidth]{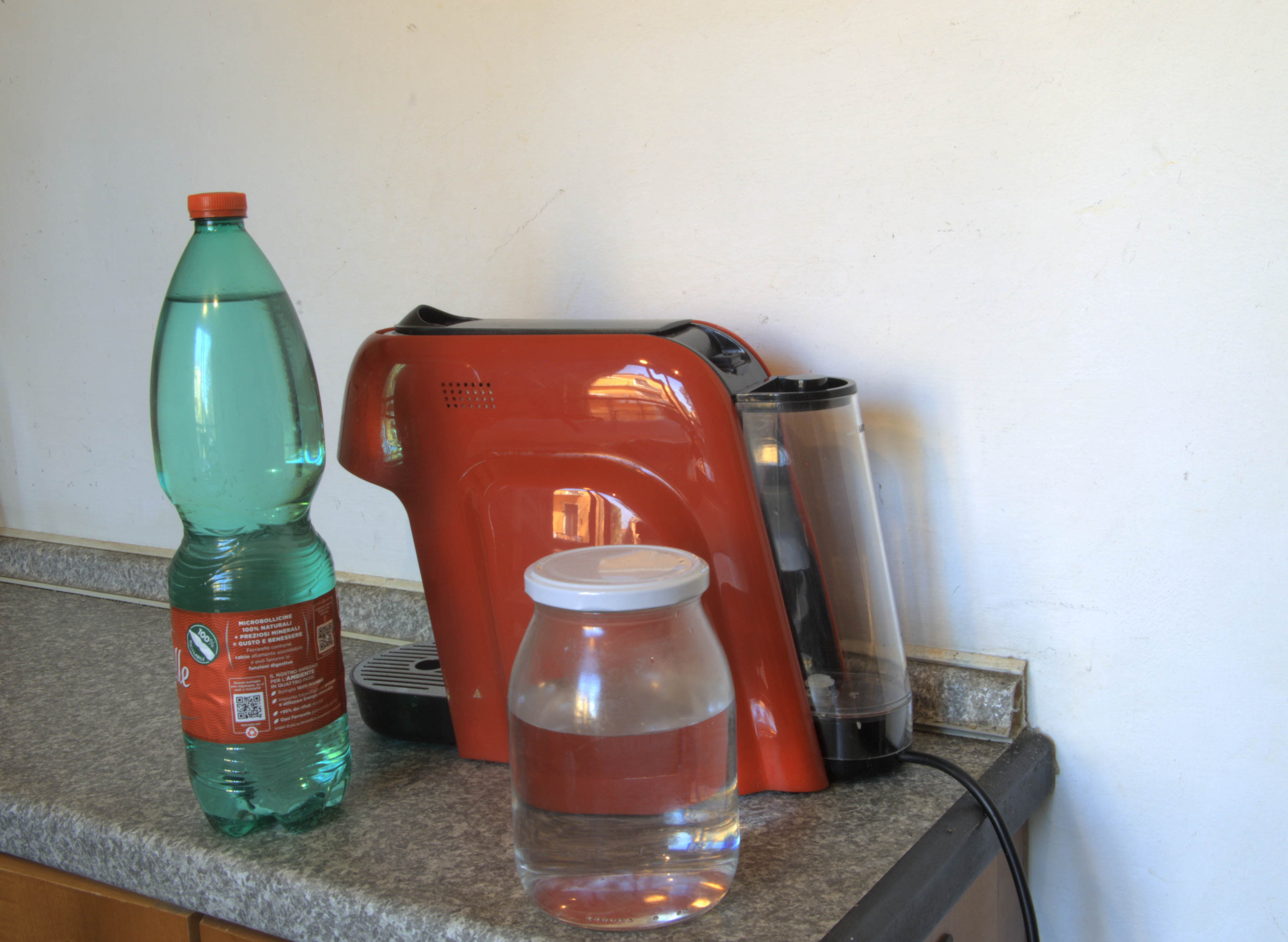} &
        \includegraphics[trim=1.8cm 0cm 4.2cm 0cm,clip,height=0.14\textwidth]{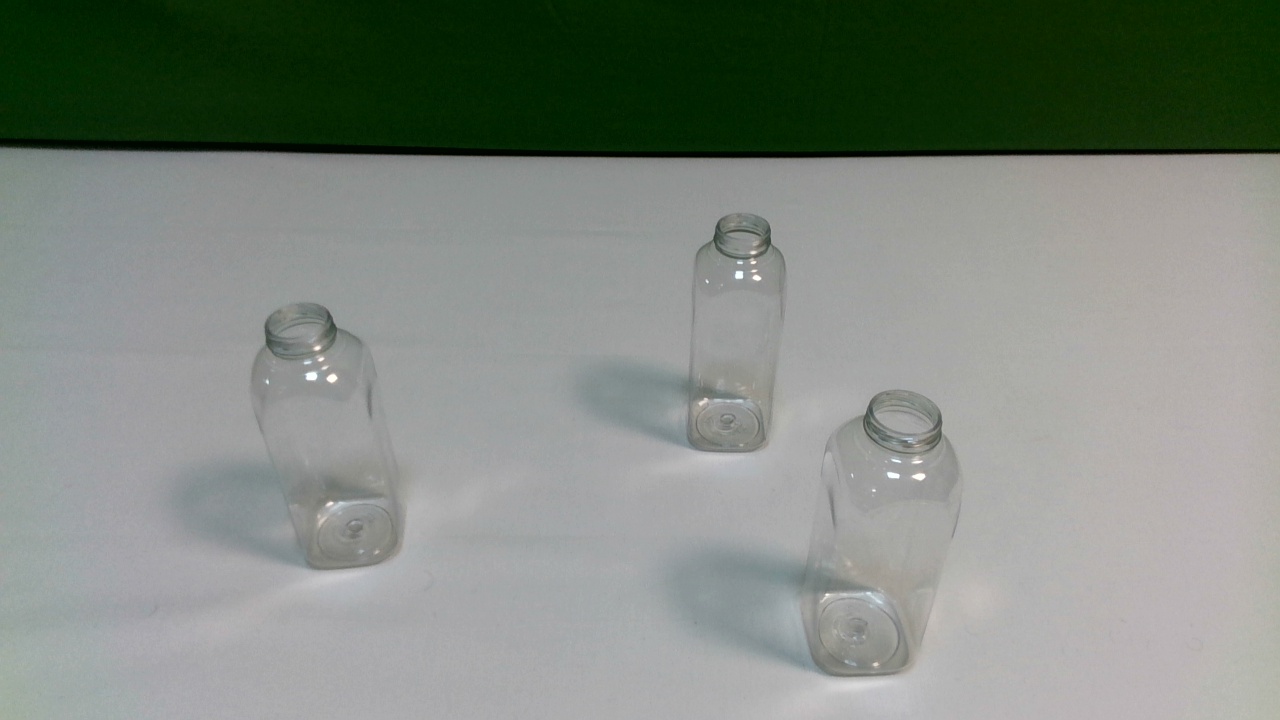} &
        \includegraphics[trim=0cm 6.5cm 0cm 0cm,clip,height=0.14\textwidth]{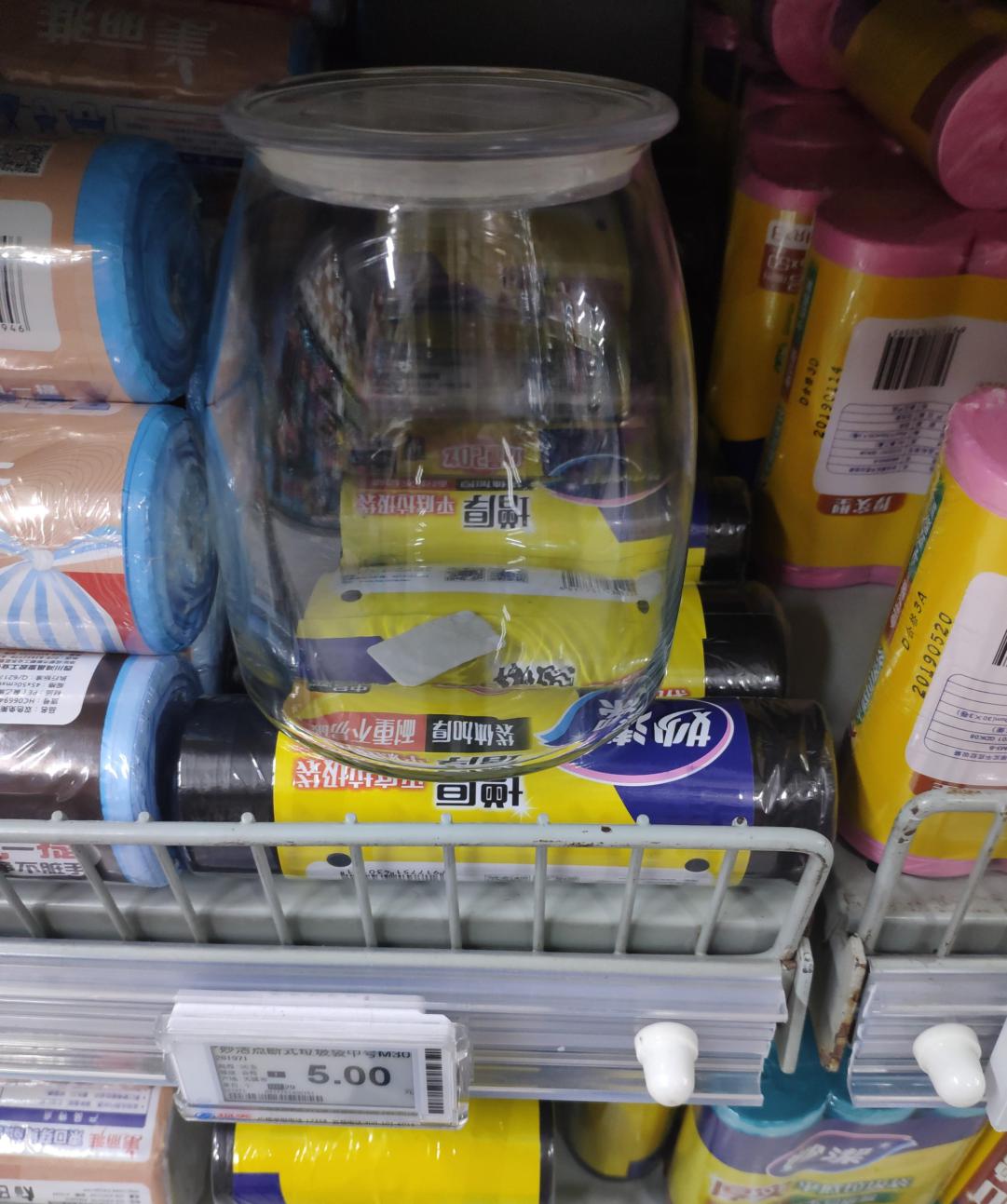} \\

        \includegraphics[trim=1.5cm 0cm 1.5cm 0cm,clip,height=0.14\textwidth]{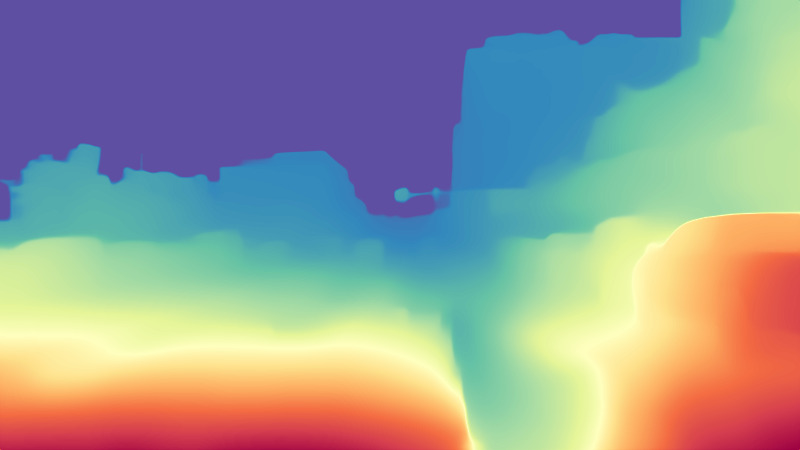} &
        \includegraphics[trim=2.5cm 0cm 9cm 0cm,clip,height=0.14\textwidth]{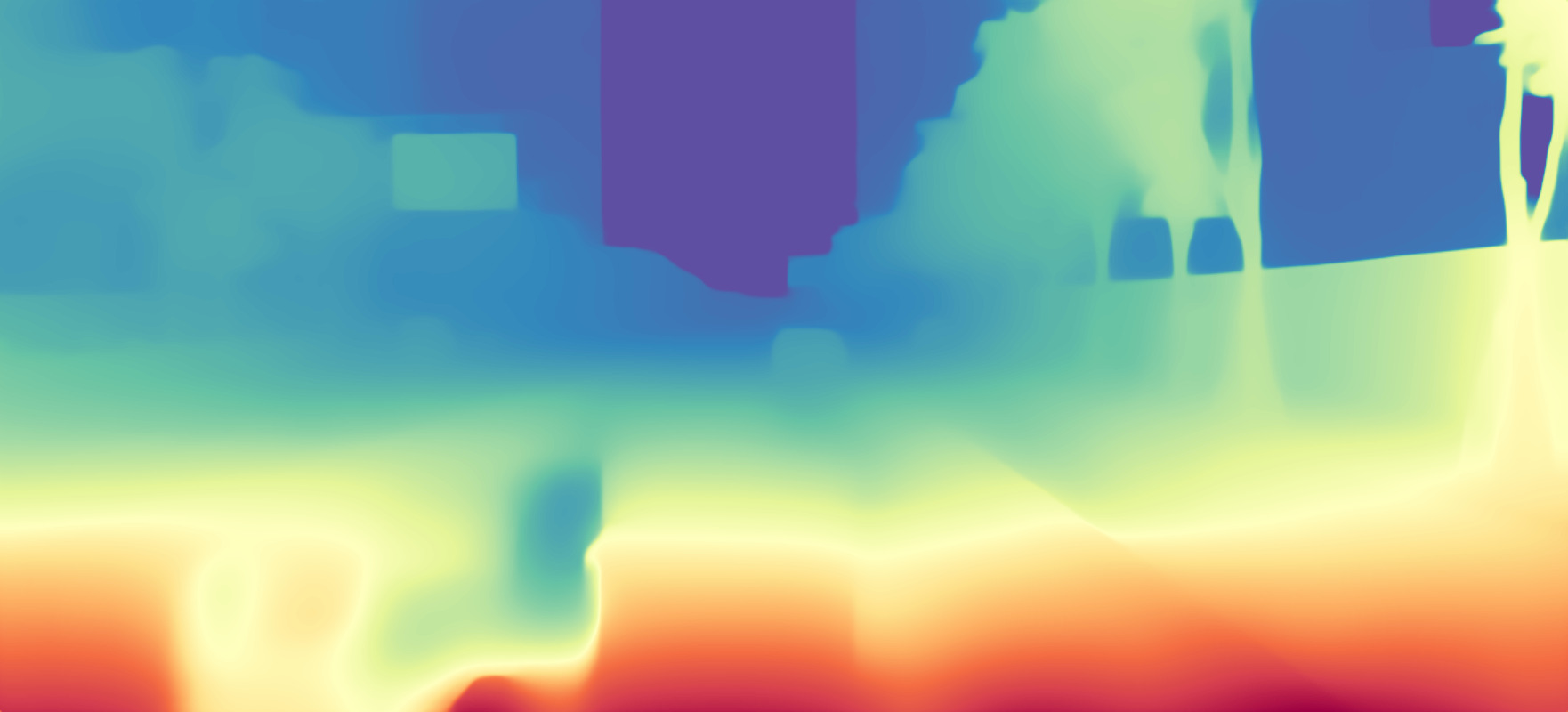} &
        \includegraphics[height=0.14\textwidth]{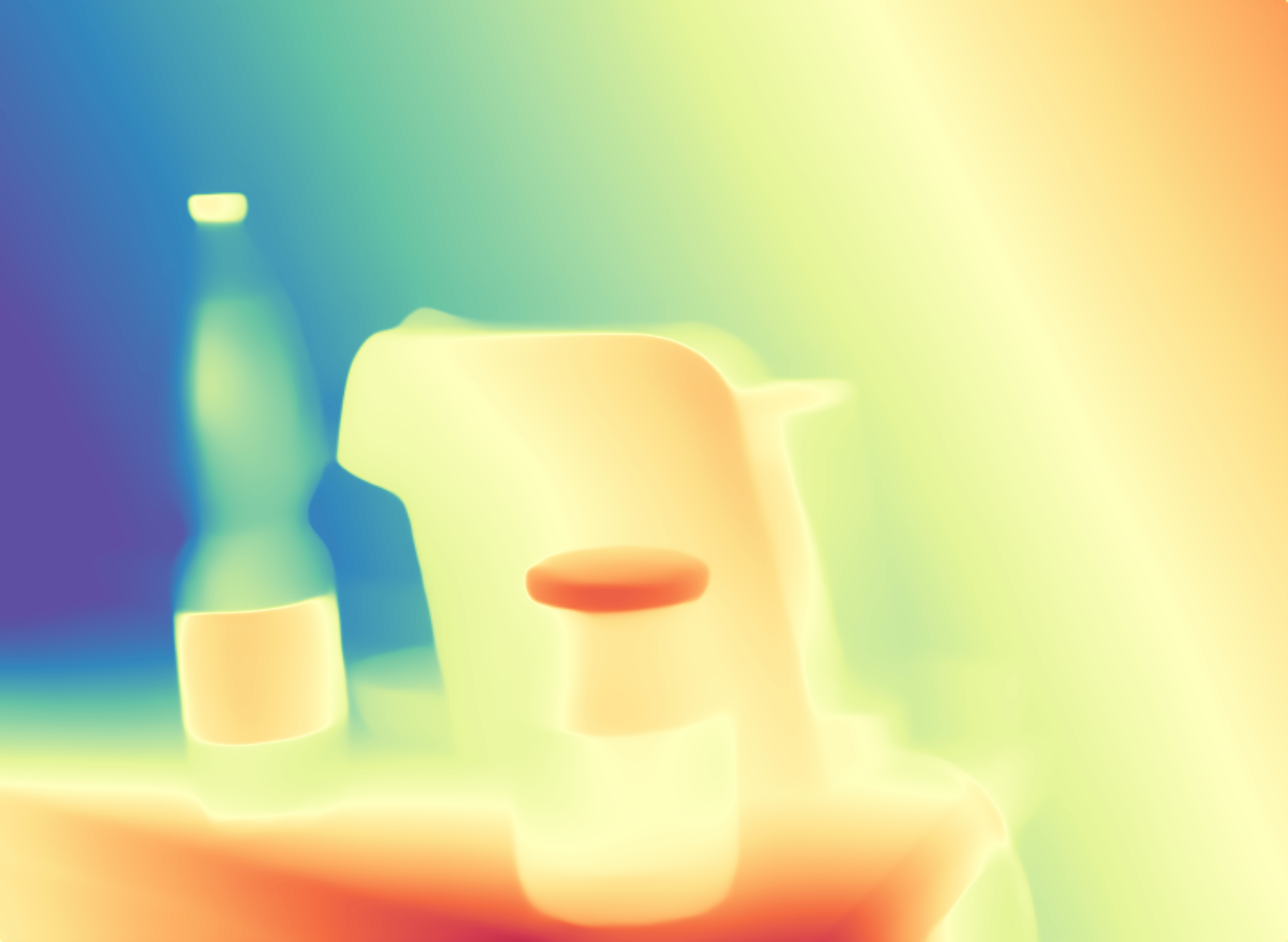} &
        \includegraphics[trim=1.0cm 0cm 3.0cm 0cm,clip,height=0.14\textwidth]{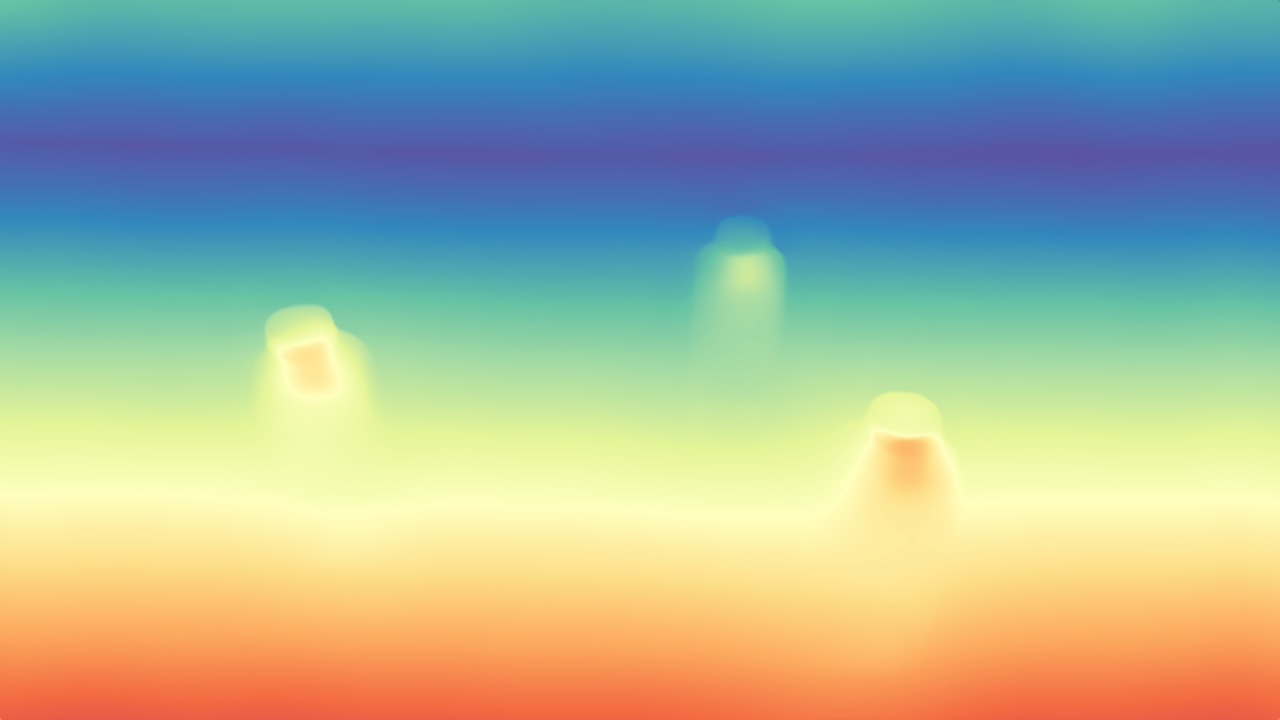} &
        \includegraphics[trim=0cm 5cm 0cm 0cm,clip,height=0.14\textwidth]{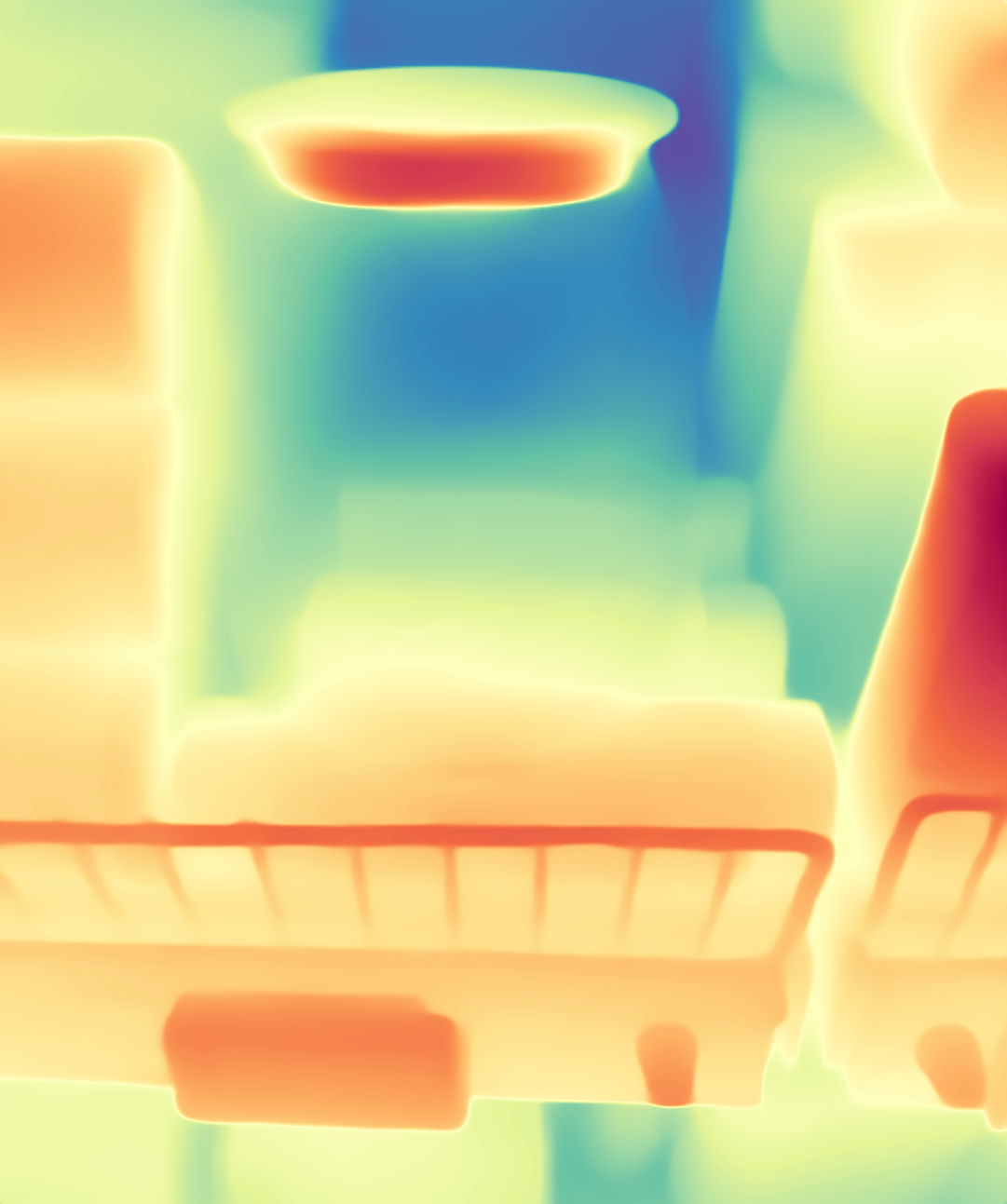} \\
        
        \includegraphics[trim=1.5cm 0cm 1.5cm 0cm,clip,height=0.14\textwidth]{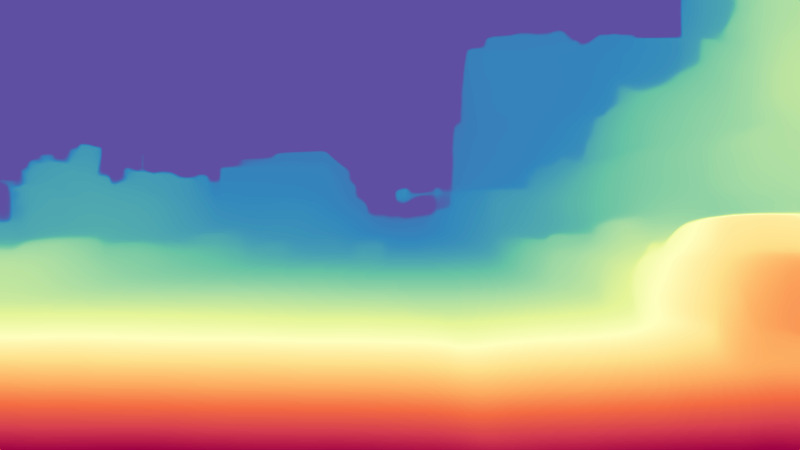} &
        \includegraphics[trim=2.5cm 0cm 9cm 0cm,clip,height=0.14\textwidth]{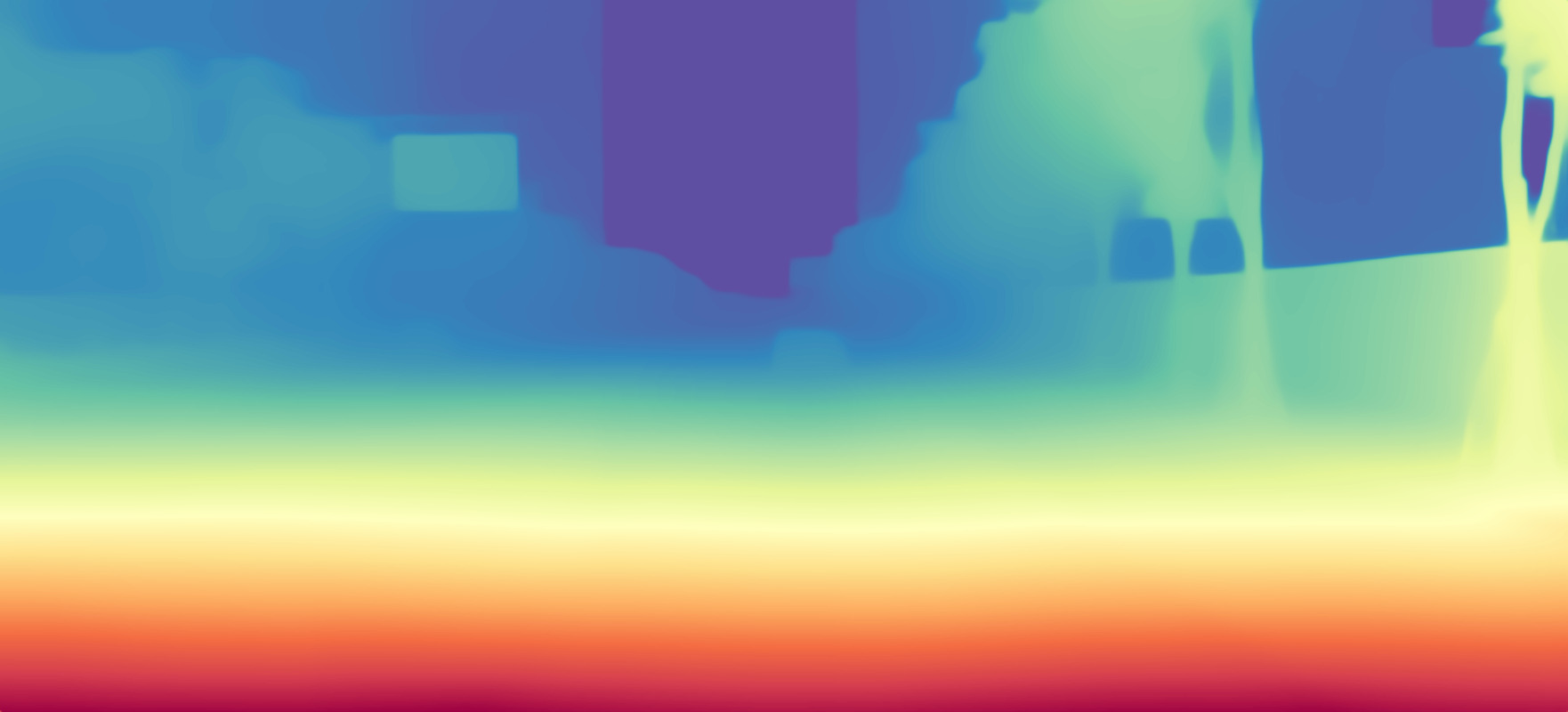} &
        \includegraphics[height=0.14\textwidth]{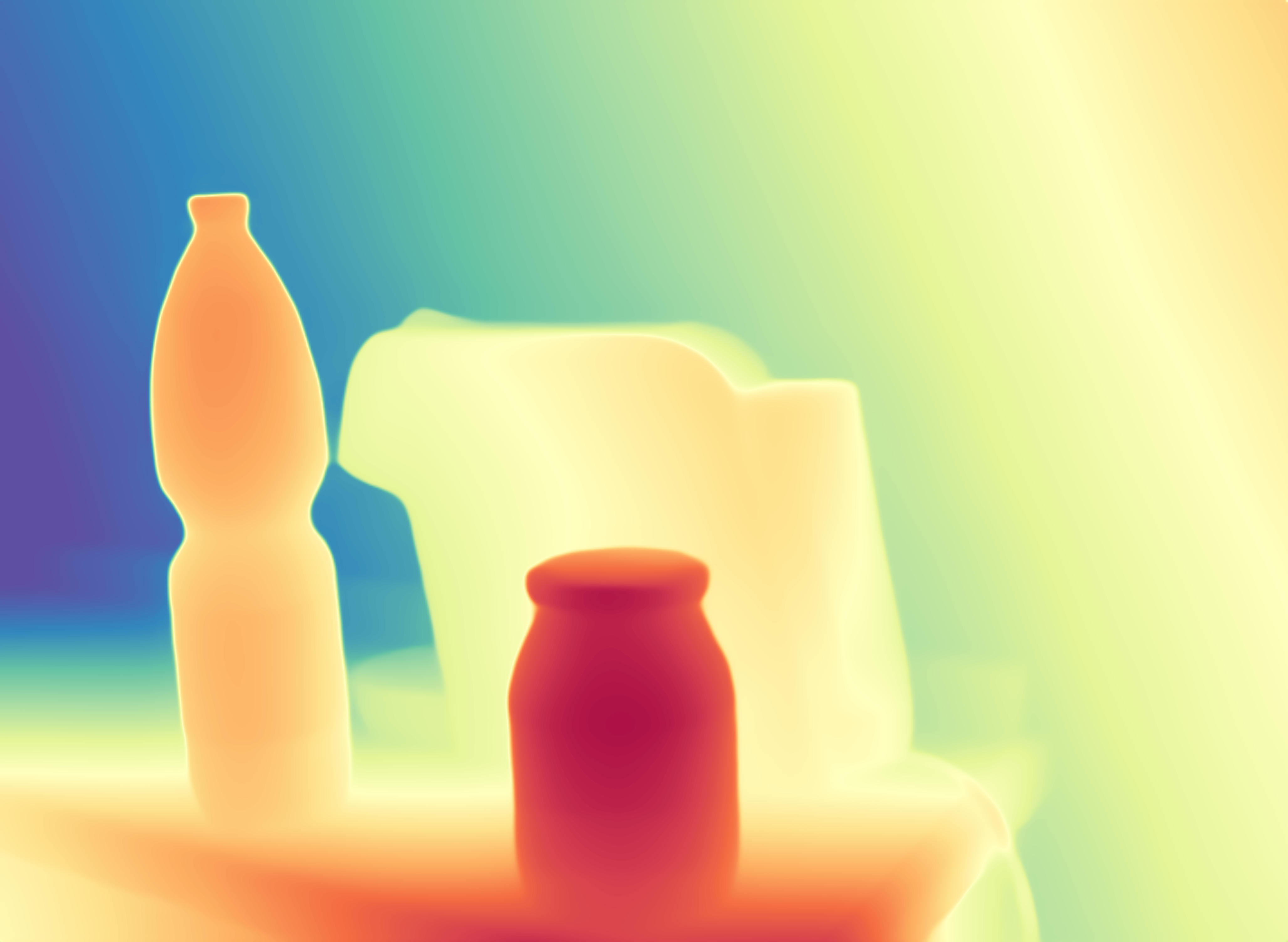} &
        \includegraphics[trim=1.0cm 0cm 3.0cm 0cm,clip,height=0.14\textwidth]{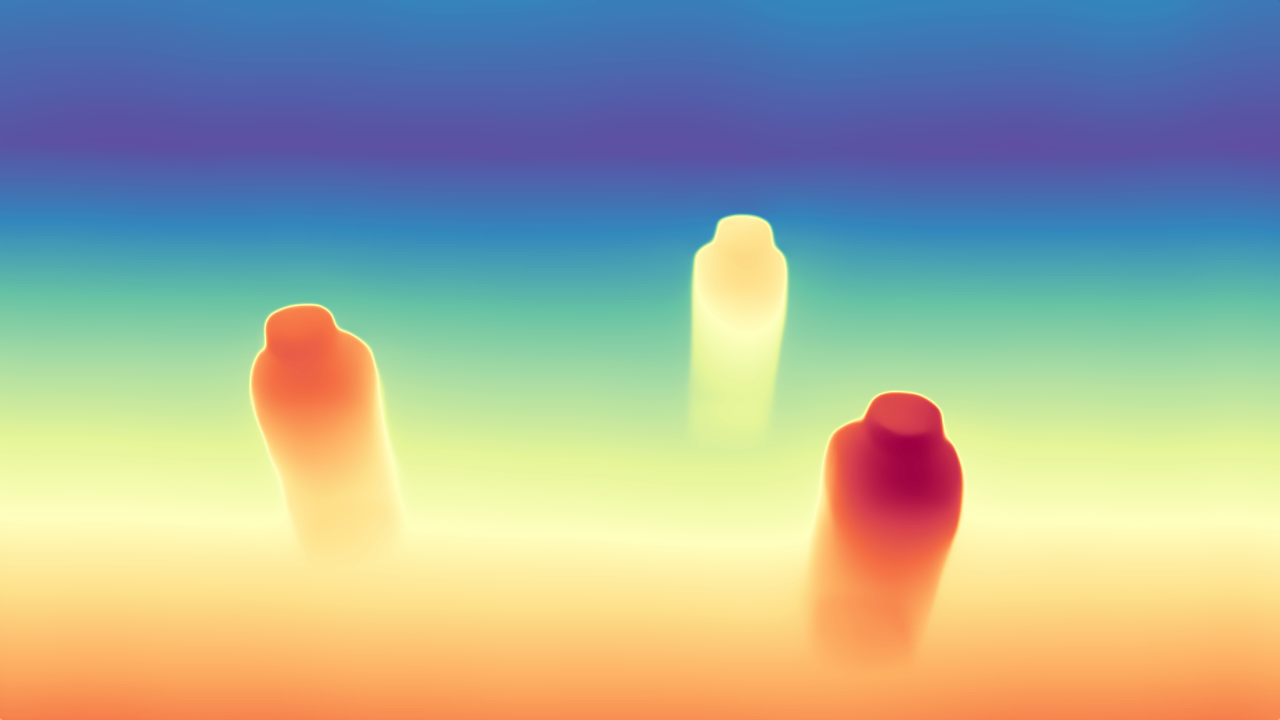} & 
        \includegraphics[trim=0cm 5cm 0cm 0cm,clip,height=0.14\textwidth]{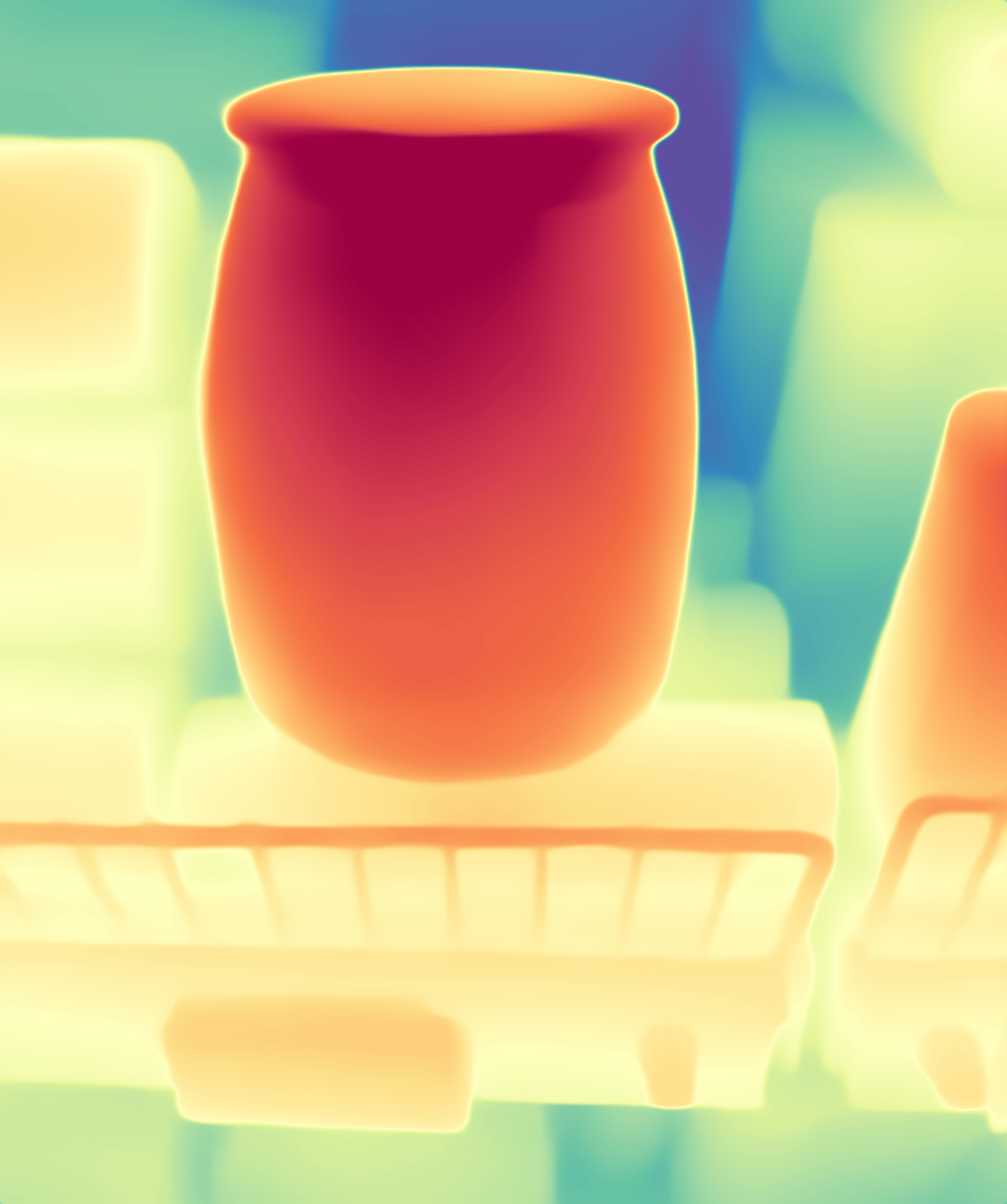} \\

    \end{tabular}
    \caption{
    \textbf{Framework Results.} From top to bottom: source image, depth predictions from the original Depth Anything \cite{yang2024depth}, and results from our fine-tuned version.}
    \label{fig:teaser}
\end{figure}

\begin{abstract}

We present a novel approach designed to address the complexities posed by challenging, out-of-distribution data in the single-image depth estimation task. Starting with images that facilitate depth prediction due to the absence of unfavorable factors, we systematically generate new, user-defined scenes with a comprehensive set of challenges and associated depth information. This is achieved by leveraging cutting-edge text-to-image diffusion models with depth-aware control, known for synthesizing high-quality image content from textual prompts while preserving the coherence of 3D structure between generated and source imagery.
Subsequent fine-tuning of any monocular depth network is carried out through a self-distillation protocol that takes into account images generated using our strategy and its own depth predictions on simple, unchallenging scenes. 
Experiments on benchmarks tailored for our purposes demonstrate the effectiveness and versatility of our proposal.

\end{abstract}


\section{Introduction} 

Monocular depth estimation, a key computer vision task, has significantly advanced due to recent breakthroughs in deep learning techniques. This has wide-ranging applications, from enhancing robotics and augmented reality to improving autonomous driving safety and precision, where relying on multiple images for depth estimation may not be feasible due to resource or spatial constraints. However, while being practical, it contends with the challenge of inferring depth from a single image, a problem acknowledged for its ill-posed and severely underconstrained nature.
Typically, addressing this challenge often involves training monocular depth networks through supervised methods \cite{lee2019big, yin2019enforcing, bhat2021adabins, wu2022toward, chen2021attention, patil2022p3depth, yuan2022neural} using annotations from active sensors or self-supervised techniques that exploit stereo image pairs \cite{monodepth17} or monocular video sequences \cite{zhou2017unsupervised} at training time.

State-of-the-art models, such as DPT \cite{Ranftl2021} and the newer Depth Anything \cite{yang2024depth}, instead, combine insights from a large number of datasets, each with depth annotations extracted using different techniques. This extensive training protocol equips these models to excel in a wide range of real-world scenarios. Nevertheless, it is crucial to stress that even these models, while excelling in numerous settings, face significant challenges when dealing with data falling far from the distribution observed during training -- such as, for instance, adverse conditions (\eg, rain and nighttime), or objects featuring non-Lambertian surfaces. These challenges arise mainly from insufficient high-quality annotated data for robust model training, compounded by the limitations of existing vision-based depth extraction techniques as well as active sensors (\eg, LiDAR, ToF, Kinect, etc.), which struggle in complex environments such as rain, snow, or materials with specific reflectivity properties. As a result, depth estimates in such settings tend to be unreliable, yielding severe implications for subsequent applications reliant on accurate 3D information. 
Tipically, current approaches tend to address these challenges independently. Some focus solely on resolving the issue of poor illumination and adverse weather \cite{gasperini_morbitzer2023md4all, gasperini2021r4dyn, wang2021regularizing}, while others tackle the problem of non-Lambertian surfaces \cite{costanzino2023iccv}. These disjointed approaches underscore the need for a unified methodology - a single framework capable of addressing all adverse scenarios simultaneously, providing a more robust and general solution.
 
In this work, we introduce diffusion models \cite{dhariwal2021diffusion, kingma2021variational}, originally designed for image synthesis, as a pioneering strategy to address the demanding challenges posed by those images that fall in the long tail of the data distribution usually considered to train depth estimation models.

Building upon principles of text-to-image diffusion models with multi-modal controls \cite{zhang2023adding, mou2023t2i}, we aim to create a diverse collection of highly realistic scenes that accurately replicate the 3D structure of a specific reference setting,  but are intentionally enriched with various adverse factors. Importantly, these conditions are purely arbitrary and can be customized with user-defined text prompts based on the specific application of interest. 

More specifically, our approach begins by selecting images that initially depict scenes devoid of the complexities associated with adverse conditions. These samples can be obtained either from an existing real-world dataset \cite{neuhold2017mapillary, Geiger2012CVPR, Cordts2016Cityscapes}, through custom collections, or even generated using generative models \cite{stable-diffusion-xl, openai-dall-e-2}. With the preselected images, we employ any readily available monocular depth estimation network to provide an initial 3D representation of the scenes. Importantly, such a model can be pre-trained on different large-scale datasets or tailored to a specific domain based on the application requirements.

Subsequently, we apply text-to-image diffusion models to transform the initial unchallenging images into more complex ones while preserving the same underlying 3D scene structure (\ie depth). 
After combining complex and simple imagery, the pre-trained depth network used for the 3D data generation enters the fine-tuning phase. In this stage, we expose the model to the composed dataset, providing it with challenging training images and their corresponding depth maps obtained in the initial step. This fine-tuning process refines the ability of the monocular network to infer depth, enabling it to better handle adverse settings, as clearly shown in Fig. \ref{fig:teaser}.

We summarize our main contributions as follows:

\begin{itemize}
    \item We pioneer the use of diffusion models as a novel solution to address the challenges of single-image depth estimation, particularly in scenarios involving adverse weather conditions and non-Lambertian surfaces.
    \item By distilling the knowledge of diffusion models, our approach improves the robustness of existing monocular depth estimation models, especially in challenging out-of-distribution settings.
    \item Our approach tackles adverse weather and non-Lambertian challenges at once, demonstrating the potential to address multiple challenging scenarios simultaneously while achieving competitive results compared to specialized solutions \cite{gasperini_morbitzer2023md4all,costanzino2023iccv} that rely on additional training information.
\end{itemize}
\section{Related work}

\begin{figure*}[t]
    \centering
    \includegraphics[ width=0.92\linewidth]{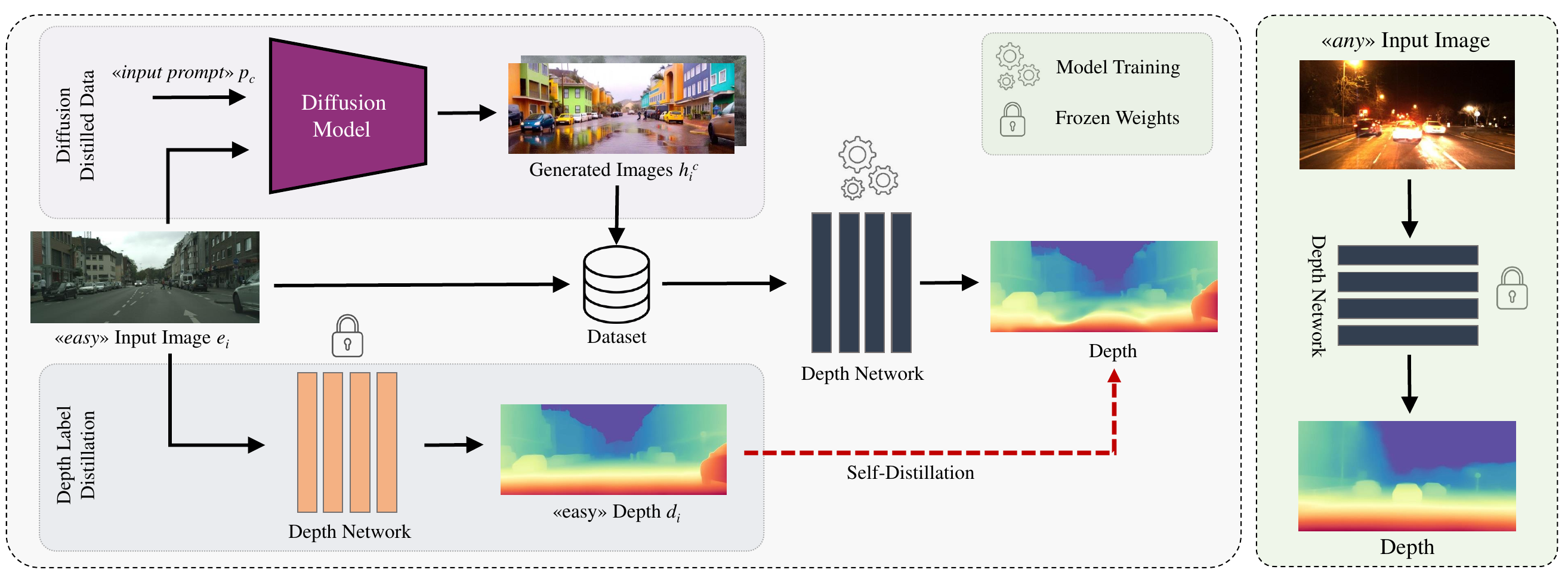}
    \caption{\textbf{Method Overview.} \textbf{Left}: Image generation and self-distillation. \textit{Diffusion Distilled Data} (upper): Easy image ($e_i$) and text prompt ($p_c$) input to conditional diffusion models generate adverse scenes ($h_i^c$). \textit{Depth Label Distillation} (lower): Pre-trained network estimates depth ($d_i$) from easy image ($e_i$). Pairs ($e_i, h_i^c$) used for fine-tuning with scale-and-shift-invariant loss. \textbf{Right}: Fine-tuned network handles diverse inputs in testing, from simple to complex scenarios.}
    \label{fig:framework}
\end{figure*}

\subsection{Monocular Depth Estimation}

Monocular depth estimation has undergone a profound shift with the emergence of deep learning \cite{zhao2020monocular}, making previous traditional methodologies \cite{saxena2005learning, saxena2008make3d, hoiem2005automatic} outdated. Initially, CNN-based methods relied on supervised training with depth data \cite{eigen2014depth, eigen2015predicting, li2015depth, liu2015learning}. Subsequently, the focus shifted to self-supervised techniques that tap into diverse sources, including stereo pairs \cite{garg2016unsupervised, monodepth17, poggi2018learning, pilzer2018unsupervised, aleotti2018generative, luo2018single, watson2019self, tosi2019learning, choi2021adaptive, peng2021excavating} or video sequences \cite{zhou2017unsupervised, mahjourian2018unsupervised, wang2018learning, bian2019unsupervised, guizilini20203d, gordon2019depth, casser2019depth, zhao2022monovit, sun2021unsupervised}. Within this context, several frameworks emerged, employing multi-task approaches by incorporating data like optical flow \cite{zou2018df, yin2018geonet, ranjan2019competitive, tosi2020distilled}, semantic segmentation \cite{zama2019geometry, guizilini2020semantically, klingner2020self}, and more. Additionally, other approaches include predicting depth uncertainty \cite{poggi2020uncertainty, hornauer2022gradient}.

Simultaneously, beyond supervised LiDAR-based techniques  \cite{lee2019big, yin2019enforcing, bhat2021adabins, wu2022toward, chen2021attention, patil2022p3depth, yuan2022neural}, recent approaches have explored techniques to mix multiple datasets \cite{Ranftl2021, Ranftl2022, eftekhar2021omnidata, guizilini2023towards, yin2023metric3d, bhat2023zoedepth, yang2024depth}, each enriched with diverse annotations from stereo or multi-view stereo followed by manual post-processing operations.

\textbf{Adverse Weather.} Despite significant advances \cite{spencer2023monocular,spencer2023second,spencer2024third}, existing monocular networks struggle under adverse weather conditions. DeFeatNet \cite{spencer2020defeat} addressed low visibility, but some had dedicated day-night branches based on GAN \cite{vankadari2020unsupervised, zhao2022unsupervised}, used extra sensors such as radar \cite{gasperini2021r4dyn}, or faced daytime trade-offs \cite{vankadari2023sun}. Adverse weather like rain also posed problems, with few solutions needing separate encoders for each condition \cite{zhao2022unsupervised}. Recently, \cite{gasperini_morbitzer2023md4all} introduced a novel GAN-based approach that addresses these issues, enabling standard models to perform robustly in diverse conditions without compromising their performance in ideal settings. Our approach, based on conditional diffusion models, relaxes the constraints of GAN-based methods as we use a single foundational model to tackle multiple challenges. This eliminates the need for separate GANs for each condition (e.g., night, rain). By combining this with text prompts, we generate potentially unlimited challenging samples from a single easy image. Furthermore, unlike other GAN-based methods that require paired easy/challenging samples, our approach needs no prior knowledge or real challenging samples beforehand. 

\textbf{Transparent and Specular Surfaces.} Estimating depth from a single image for transparent or mirror (ToM) surfaces presents a unique and complex challenge \cite{ramirez2023booster,Ramirez2023b, Ramirez_2024_CVPR}. To our knowledge, Costanzino et al. \cite{costanzino2023iccv} offer the only dedicated approach to this problem. While bypassing ground truth depth, their method relies on segmentation maps or pre-trained semantic networks specialized for these materials. They generate pseudo-labels by inpainting ToM objects in images and processing them with a pre-trained monocular depth model \cite{Ranftl2022}. These labels then enable fine-tuning of existing monocular or stereo networks to effectively handle challenging non-Lambertian surfaces.

\subsection{Image Diffusion}

Image Diffusion Models (IDMs), initially introduced by Sohl-Dickstein et al. \cite{sohl2015deep}, have gained widespread adoption in image generation \cite{dhariwal2021diffusion, kingma2021variational}. Subsequently, numerous enhancements have been proposed, improving both computational efficiency \cite{rombach2022high} and generation conditioning \cite{alembics-disco-diffusion, nichol2021glide}. Latent Diffusion Models (LDMs) \cite{rombach2022high} have notably reduced computational costs by incorporating denoising in the latent space. In terms of scale, Stable Diffusion \cite{stable-diffusion-v1-5,stable-diffusion-xl} represents a large-scale implementation of LDMs. Notably, common conditioning techniques involve cross-attention \cite{avrahami2022blended, brooks2023instructpix2pix, gafni2022make, hertz2022prompt, kawar2023imagic, kim2022diffusionclip, nichol2021glide, parmar2023zero, ramesh2022hierarchical}, and the encoding of segmentation masks into tokens \cite{avrahami2023spatext, gafni2022make}. Moreover, various conditioning schemes have been proposed to enable the generation of visual data conditioned by diverse factors such as text, images, semantic maps, sketches, and other representations \cite{bar2023multidiffusion, bashkirova2023masksketch, huang2023composer, mou2023t2i, zhang2023adding, voynov2023sketch}. 
In addition to image generation, diffusion models have exhibited remarkable capabilities in optical flow and monocular depth estimation \cite{saxena2023monocular, saxena2023surprising, ke2024repurposing}. Our approach stands out for using established conditioned diffusion models to tackle diverse challenges, including rain, night scenes, and non-Lambertian surfaces.
\section{Method}

This section overviews our framework, illustrated in Fig. \ref{fig:framework}, to improve monocular depth estimation in adverse settings. Assuming the absence of images depicting both \textit{easy} and \textit{challenging} conditions in a domain, our approach converts \textit{easy} samples to \textit{challenging} ones using diffusion models with depth-aware control. Subsequently, we fine-tune a pre-trained monocular depth network through self-distillation and a scale-and-shift-invariant loss using the generated data.

\subsection{Background: Diffusion Models}

Diffusion models have significantly influenced generative modeling in computer vision. These probabilistic generative models exhibit a distinct capability to generate highly realistic images from random noise.
This transformation involves two crucial phases—\textit{forward} and \textit{reverse} diffusion—formally outlined as follows: 

\textbf{Forward Diffusion.} This phase consists of the progressive degradation of an image by adding scaled Gaussian noise. It is defined as \(x_t = x_{t-1} + \epsilon_{t-1}\), where \(x_t\) represents the image at time step \(t\), and \(\epsilon_{t-1}\) denotes the noise increment from the preceding step. This process results in a progressively noised image, converging towards an isotropic Gaussian distribution.

\textbf{Reverse Diffusion.} Conversely, the reverse diffusion phase aims to restore the original image from its noised counterpart. For this purpose, the process begins with the noise \(x_t\) and generates progressively less noisy samples \(x_{t-1}, x_{t-2}, \ldots\) until it reaches the original image \(x_0\). A diffusion model is trained to generate \(x_{t-1}\) from \(x_t\) by predicting the noise component, denoted as \(\epsilon\). This prediction is performed by a neural network represented as \(\mathcal{N}_{\theta}(x_t, t)
\).

Through iterative application, the reverse diffusion mechanism equips the model to approximate the underlying data distribution adeptly. The prevalent architecture for noise-predicting networks involves the UNet \cite{ronneberger2015u} framework, trained using Mean Squared Error (MSE) loss at each temporal interval.
Diffusion models are characterized by their iterative approach, which provides stability in training and generation, unlike other generative models such as GANs \cite{creswell2018generative}.

\subsection{Conditional Diffusion Models}

Conditional Diffusion Models (CDMs), on the other hand, transform generative modeling by incorporating various conditions for image generation, ranging from textual cues to advanced visual information such as depth maps, segmentation maps, gradients, normals, and key points. Notably, ControlNet \cite{zhang2023adding}, 
a neural network capable of learning to
condition large diffusion models trained on billion samples from hundreds of times fewer data,
plays a key role in enabling controllable image generation. Seamlessly integrating diverse input conditions, from traditional textual prompts to complex visual data, ControlNet comprises two sets of weights within a pretrained diffusion model: a \textit{trainable copy} (\(\theta_{\text{train}}\)) and a \textit{locked copy} (\(\theta_{\text{locked}}\)). The trainable copy adapts dynamically to task-specific datasets during learning, fine-tuning parameters based on input conditions. In contrast, the locked copy retains knowledge from generic datasets, forming a robust foundation for image generation.
Central to ControlNet is the \textit{zero convolution} layer, denoted as \(\mathcal{L}_{\text{zero-conv}}\). Initialized with zero weights and biases, this layer plays a pivotal role during training. The convolution weights gradually evolve from zeros to optimized parameters, ensuring adaptability without introducing new noise to deep features. 

Following the principles established by ControlNet, several other works have emerged, exploring similar approaches to enhance the controllability and flexibility of pre-trained diffusion models \cite{zavadski2023controlnet,hu2023cocktail,mou2023t2i, mou2024t2i,zhao2024uni}. These studies aim to further improve the alignment between internal model knowledge and external control signals, enabling more precise and versatile image generation while maintaining the benefits of large-scale pre-training.

Leveraging these advancements in text-to-image diffusion models, we focus on generating complex scenarios from textual cues, employing 3D data information presented as depth maps derived from images devoid of challenges. 

\subsection{Diffusion-Distilled Data}

Our core goal is to curate a training dataset that addresses the scarcity of real-world challenging data with associated depth and is explicitly designed to improve the robustness of monocular depth estimation networks in adverse scenarios. Drawing inspiration from \cite{gasperini_morbitzer2023md4all}, we introduce paired images, denoted as $(e_i, h_i^c)$. 
In this context, $e_i$ represents an \textit{easy} sample belonging to the set $E$ of images that do not pose challenges for depth estimation models. These samples depict images captured under optimal environmental conditions, making them ideal for robust training and testing of monocular networks. They may include well-lit daytime scenes with excellent visibility, as well as scenes portraying surfaces and objects with Lambertian material characteristics. On the other hand, the collection of paired \textit{challenging} samples, denoted as $h_i^c \in H$, is meticulously designed to replicate a diverse array of adverse scenarios faithfully. Here, $H$ represents the set of the difficult samples for a specific condition of interest ($c \in C$). The set of conditions $C$ includes complex surfaces, non-Lambertian objects, and adverse weather conditions. Typically, training a monocular network on these scenarios is challenging due to their limited availability, and annotating them is very costly and often impractical, even with the use of active sensors.
In response to this, we employ recently developed text-to-image diffusion models, such as ControlNet \cite{zhang2023adding} and T2I-Adapter \cite{mou2023t2i}, guided by external textual prompts and, crucially, depth. These models systematically transform the \textit{easy} samples $e_i$ into \textit{challenging} counterparts $h_i^c$. The transformation process involves using the $e_i$ as input, alongside the depth map $d_i$ from $e_i$, and a corresponding text prompt $p_c$, which essentially describes the target scenario for a specific condition. 
The two conditioning inputs serve distinct purposes: i) modeling tasks such as simulating day-to-night transitions or transforming opaque objects into transparent or highly reflective surfaces, and ii) preserving the inherent 3D structure of the original \textit{easy} image $e_i$. This ensures that the depth $d_i$ remains consistent between the source and generated $h^c_i$ images.
This process allows for the distillation of hard samples on which depth estimation models struggle, yet obtaining reliable depth labels for free, \ie by predicting depth on \textit{easy} images.

\begin{figure*}[t] 
    \centering
    \begin{subfigure}[t]{0.32\textwidth}
        \centering
        \includegraphics[width=\textwidth]{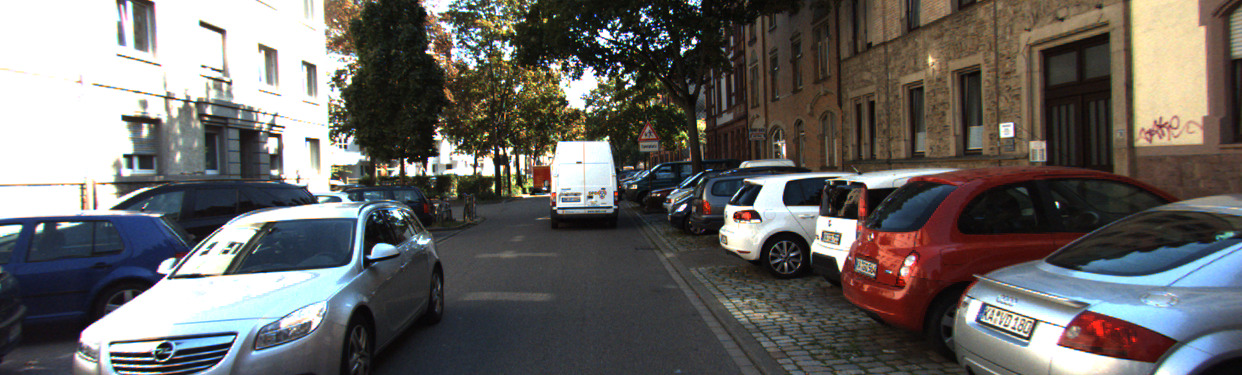}
        \caption{RGB Image ($e_i$)}
    \end{subfigure}
    \begin{subfigure}[t]{0.32\textwidth}
        \centering
        \includegraphics[width=\textwidth]{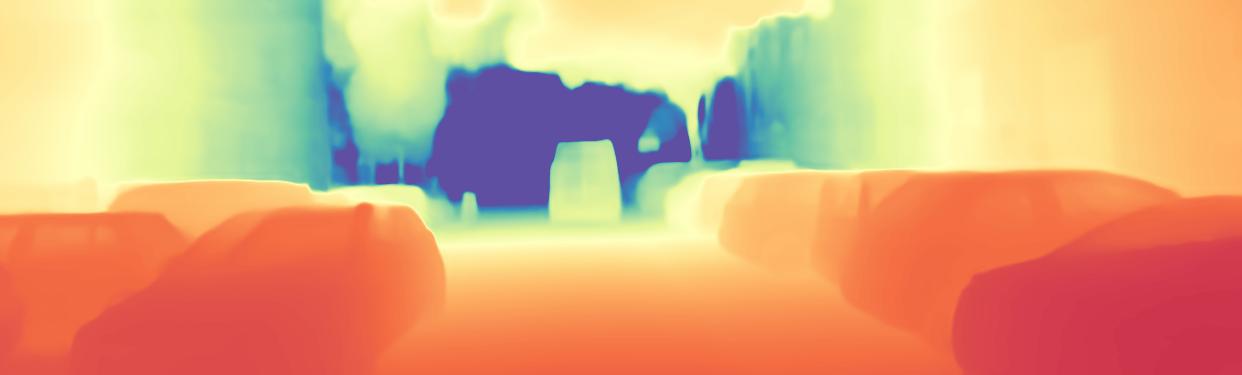}
        \caption{Depth Map ($d_i$) \cite{bhat2023zoedepth}}
    \end{subfigure}
    \begin{subfigure}[t]{0.32\textwidth}
        \centering
        \includegraphics[width=\textwidth]{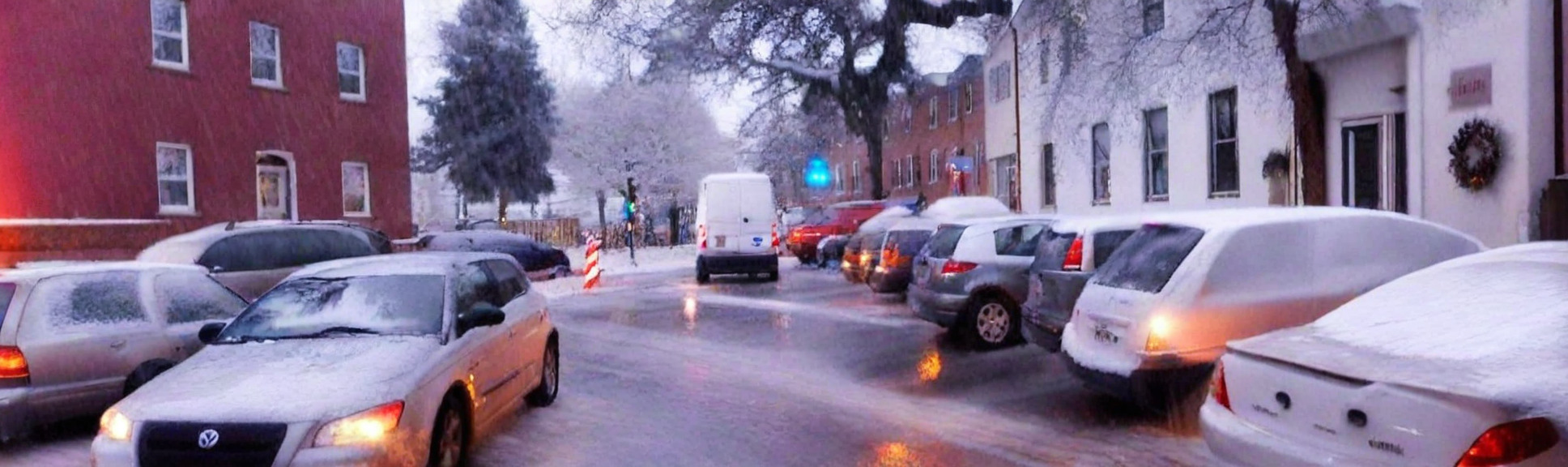}
        \caption{\centering\scriptsize"\textit{Snowy town street with cars}"}
    \end{subfigure}

    \medskip 

    \begin{subfigure}[t]{0.32\textwidth}
        \centering
        \includegraphics[width=\textwidth]{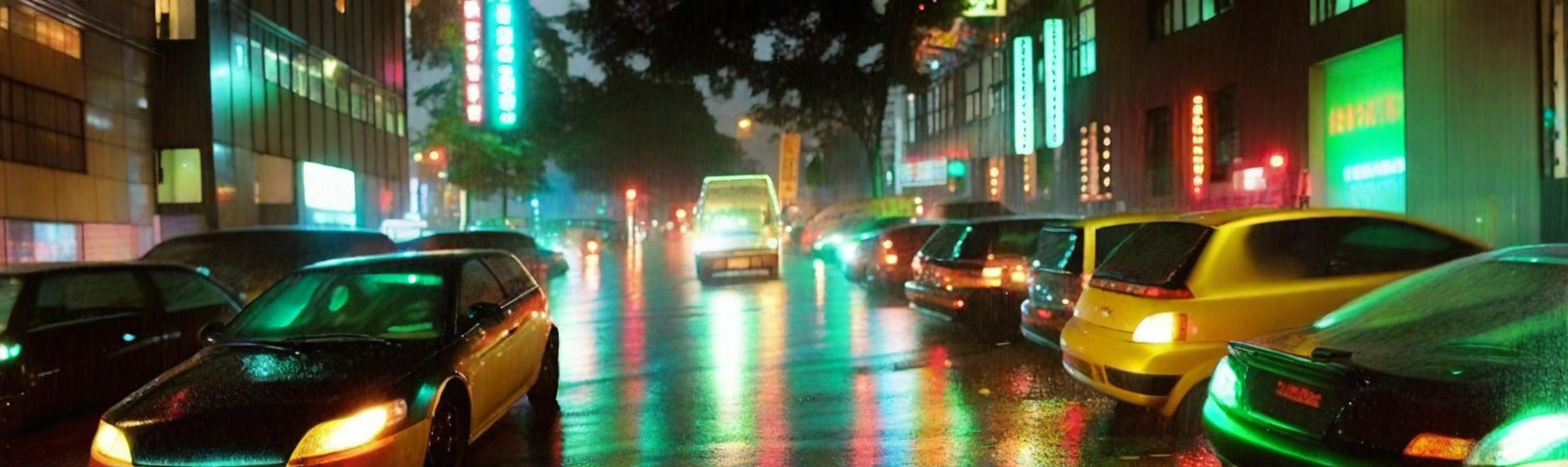}
        \caption{\centering\scriptsize"\textit{Rainy city street at night with colorful neon lights.}"}
    \end{subfigure}
    \begin{subfigure}[t]{0.32\textwidth}
        \centering
        \includegraphics[width=\textwidth]{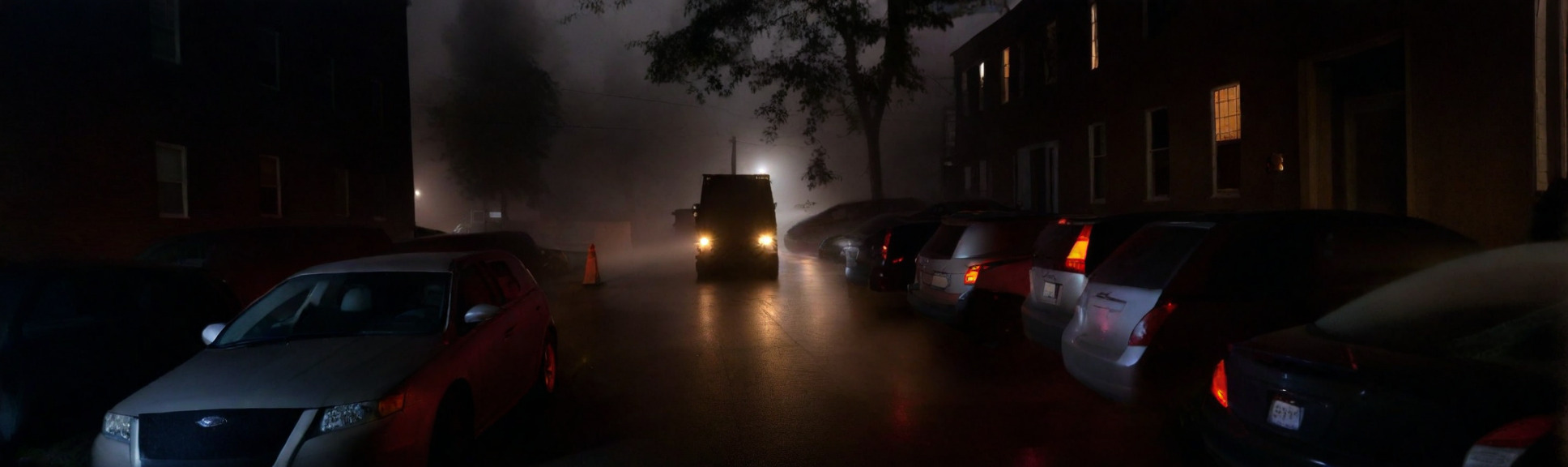}
        \caption{\centering\scriptsize"\textit{Foggy night street with cars and glowing headlights.}"}
    \end{subfigure}
    \begin{subfigure}[t]{0.32\textwidth}
        \centering
        \includegraphics[width=\textwidth]{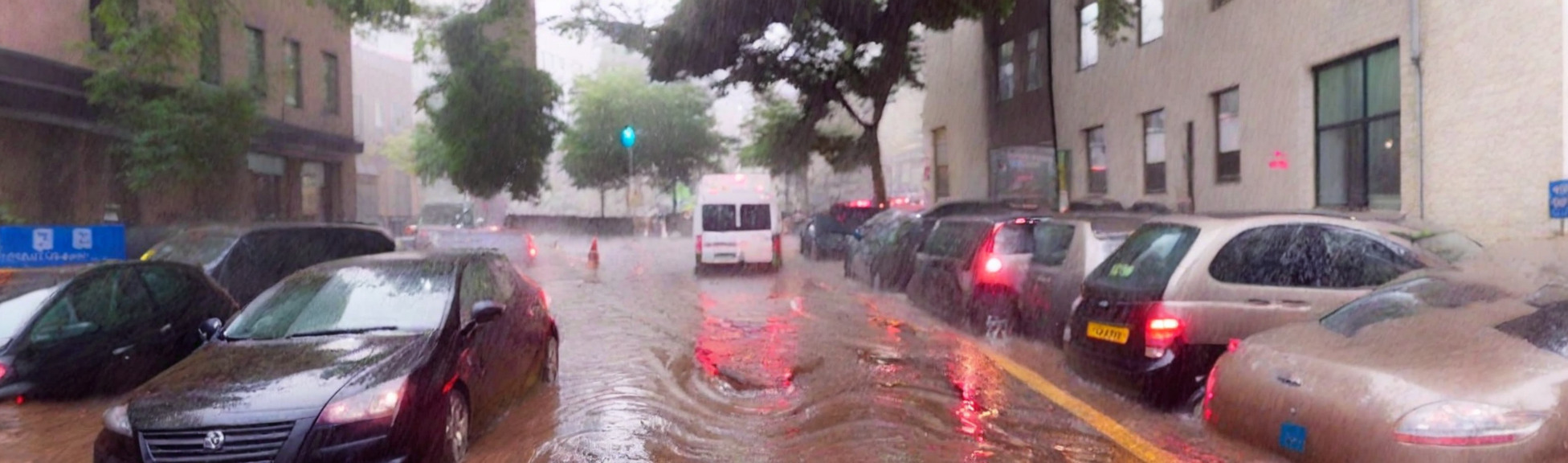}
        \caption{\centering\scriptsize"\textit{Flooded urban street with cars in heavy rain.}"}
    \end{subfigure}

    \caption{\textbf{Generated Images -- Weather Conditions.} (a-b): RGB and depth maps from KITTI 2015 \cite{Menze2015CVPR}. (c-f): images generated by a diffusion model \cite{mou2023t2i}, conditioned by the depth map from (b) and text prompts indicated in each subfigure.}
    \label{fig:example_text_prompts}
\end{figure*}

Playing with text-prompts, as shown in Fig. \ref{fig:example_text_prompts}, allows for a coverage of a potentially infinite number of complexities, ranging from adverse weather to non-Lambertian objects and more.
This process uses only diffusion model prompts to generate samples realistically depicting desired real-world conditions.

\subsection{Self-Distillation Training}

After generating challenging images using text-to-image diffusion models with depth-aware control, we fine-tune existing pre-trained models to improve their robustness to complex, out-of-distribution data. It is worth noting that, pre-training methods and fine-tuning models can be arbitrary, including supervised techniques using LiDAR-derived depth, photometric losses from video, stereo sequences, or other approaches. This makes our approach model agnostic and applicable to various existing and future diffusion or monocular models.

By employing distillation in a teacher-student paradigm, the pre-trained depth estimation model acts as the teacher, providing depth labels for both \textit{easy} and \textit{challenging} images generated, which are paired. Subsequently, the student network, instantiated from the same pre-trained teacher depth estimation network, undergoes a fine-tuning process using the scale-and-shift-invariant loss \cite{Ranftl2021,Ranftl2022} defined as $L_{\text{ssi}}(\hat{d}, \hat{d}^*) = \frac{1}{2M} \sum_{i=1}^{M} \rho(\hat{d}_i - \hat{d}_i^*)$.

This loss function compares scaled and shifted predictions $\hat{d}$ with corresponding inverse depth labels $\hat{d}^*$ from the teacher. Here, $\rho$ quantifies the absolute difference between predictions and provided annotations.
\section{Experimental Results}

\subsection{Evaluation Datasets \&   Protocol}

\textbf{Autonomous Driving Datasets}. Our study draws on a diverse selection of datasets for thorough evaluation. The \textbf{nuScenes} dataset \cite{caesar2020nuscenes}, known for its diverse weather conditions and integration of LiDAR data, consists of 1000 scenes. We adopt the split recommended in \cite{gasperini2021r4dyn, gasperini_morbitzer2023md4all}, which yields 15,129 training images and 6,019 validation images categorized by night and rainy weather conditions. The \textbf{RobotCar} dataset \cite{maddern20171}, which captures over 1,000 km in central Oxford, offers a collection of 20M images, LIDAR, GPS, and INS ground truth. Following \cite{gasperini_morbitzer2023md4all}, we use 16,563 training samples and 1,411 test images, with 709 showing nighttime scenes. We further leverage  \textbf{DrivingStereo} dataset \cite{yang2019drivingstereo}, originally designed for stereo but adapted for monocular research, with a specific focus on 500 frames representing rainy weather conditions. 
Furthermore, in our experiments, we utilize images from \textbf{KITTI 2012} \cite{Geiger2012CVPR}, \textbf{KITTI 2015} \cite{Menze2015CVPR}, \textbf{Apolloscape} \cite{huang2019apolloscape}, \textbf{Mapillary} \cite{neuhold2017mapillary}, and \textbf{Cityscapes} \cite{Cordts2016Cityscapes} for training purposes only.

\textbf{Non-Lambertian Datasets}. 
We select datasets based on their availability of ground truth depths for non-Lambertian objects. 
The \textbf{Booster} \cite{zamaramirez2022booster} dataset contains 228 and 191 images for training and testing, respectively. Images feature indoor scenes with non-Lambertian objects such as mirrors or glasses. The training set provides disparity and segmentation maps employed to evaluate approaches in our experiments. Each pixel in segmentation maps is categorized into 4 classes based on the type of surface material. Following \cite{costanzino2023iccv}, we define 2 macro-categories --  "ToM" (Transparent or Mirror) for classes 2-3, "Other" materials for labels 0-1.
The \textbf{ClearGrasp} \cite{sajjan2020clear} dataset, instead, comprises a synthetic and a real-world split. We use the latter in our experiments, made of 286 RGB-D images of transparent objects and their ground truth geometries, 
together with binary masks for ToM or Other objects.

\textbf{Evaluation Metrics.} Our evaluation in  driving scenarios, following \cite{gasperini_morbitzer2023md4all}, employs three standard metrics \cite{eigen2014depth}: \textit{AbsRel}, defined as $\frac{1}{n} \sum_{ij} \left| \frac{d_{\text{{gt}}} - d_{\text{{p}}}}{d_{\text{{gt}}}} \right|$; $\textit{RMSE} = \sqrt{\frac{1}{n} \sum_{ij} (d_{\text{{gt}}} - d_{\text{{p}}})^2}$; $\delta < \tau$ is the percentage of pixels with $\max \left( \frac{|d_{\text{{p}}}|}{d_{\text{{gt}}}}, \frac{|d_{\text{{gt}}}|}{d_{\text{{p}}}} \right) < \tau$, where $n$ is the total number of valid depth ground truth points, while $d_{\text{{gt}}}$ and $d_{\text{{p}}}$ represent ground truth and predicted depths at a given pixel, respectively. 
We follow the metrics used in \cite{costanzino2023iccv} on non-Lambertian datasets.
We employ AbsRel, $\delta_\tau$ with $\tau$ being 1.05, 1.15, and 1.25, RMSE and the mean absolute error (MAE).
We report results on all valid pixels (\textit{All}) or for those belonging to either ToM or other objects to assess the impact of our strategy on diverse surfaces.
As the predictions by monocular networks are up to an unknown scale factor, we rescale them according to the LSE criterion from \cite{Ranftl2022}.
For uniformity, our experiments use the evaluation frameworks by \cite{gasperini_morbitzer2023md4all} and \cite{costanzino2023iccv}, respectively. In tables, we identify the 
\textbf{top} and 
\underline{next best} results across macro-categories.

\begin{figure}[t]
    \centering
    \renewcommand{\tabcolsep}{1pt}
    \begin{tabular}{cccccc}

        \includegraphics[width=0.16\textwidth]{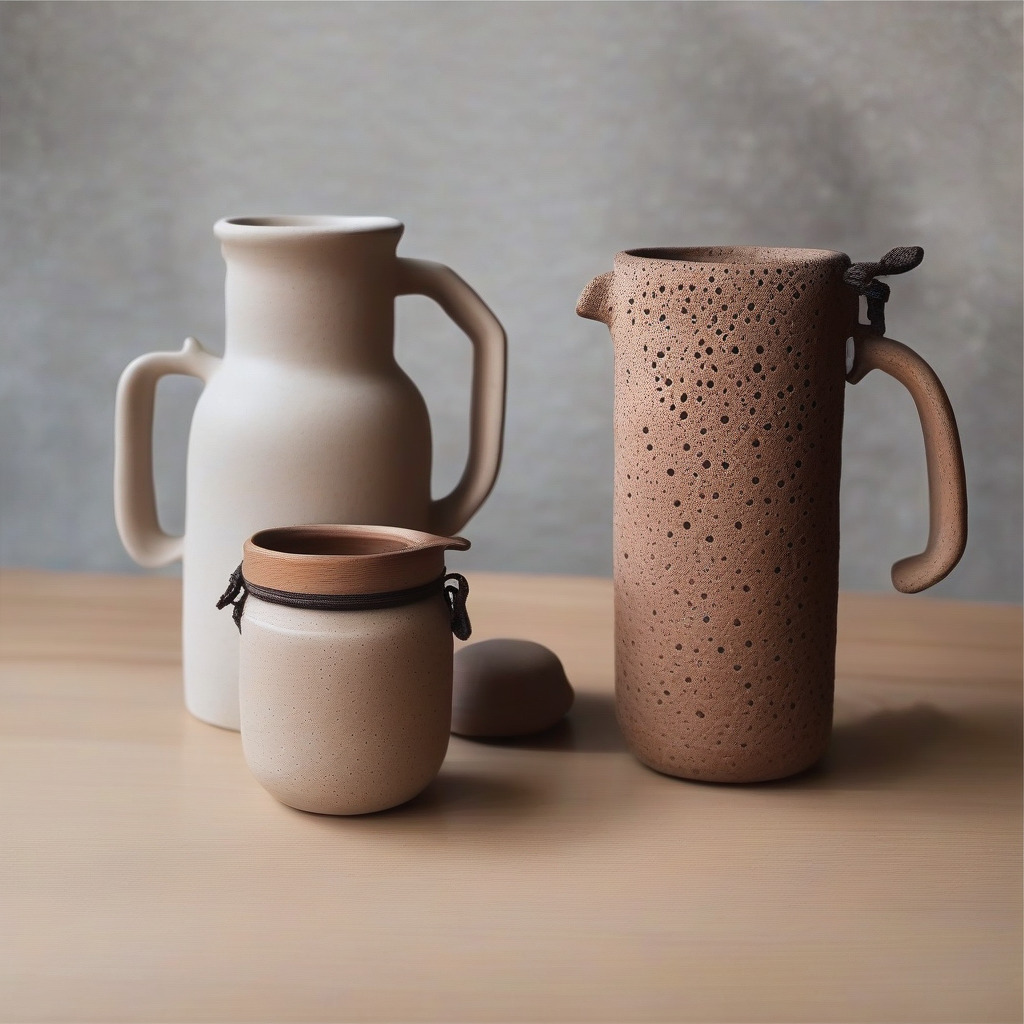} & 
        \includegraphics[width=0.16\textwidth]{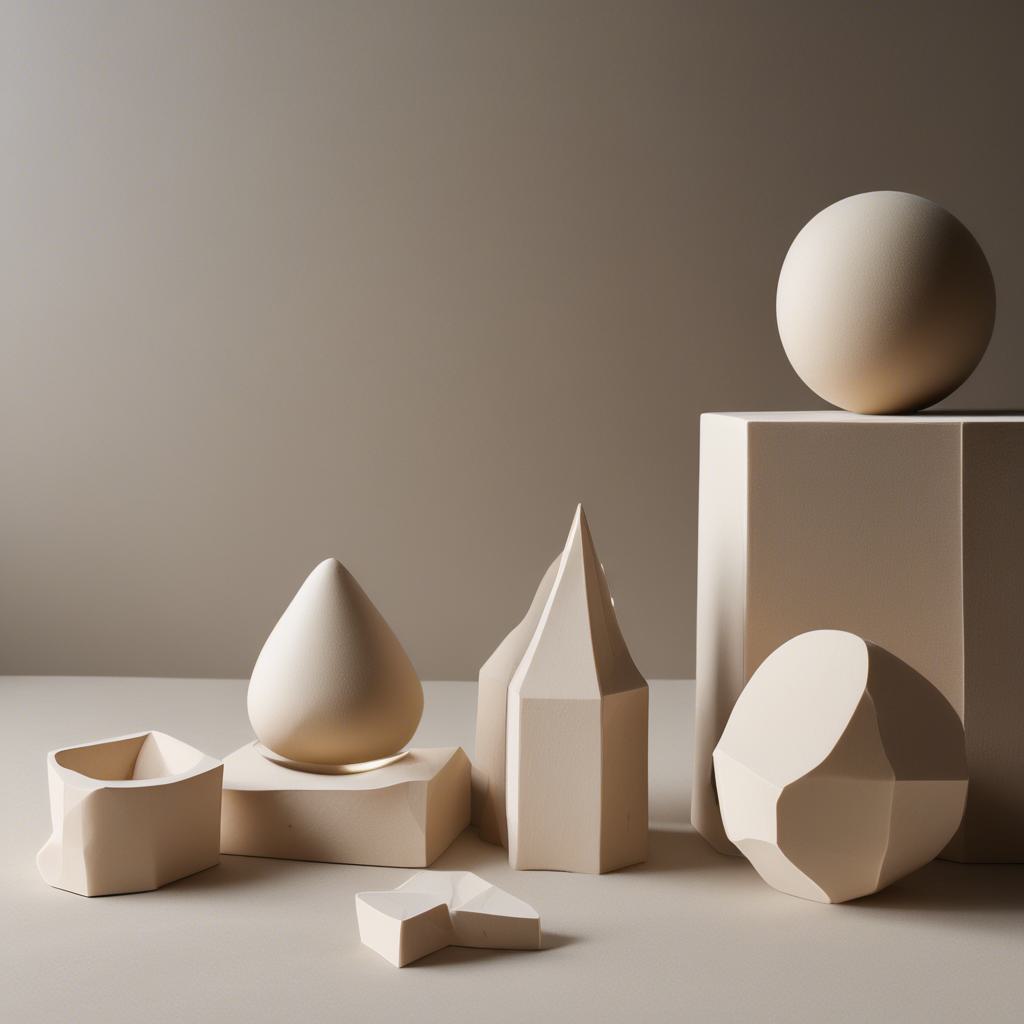} &
        \includegraphics[width=0.16\textwidth]{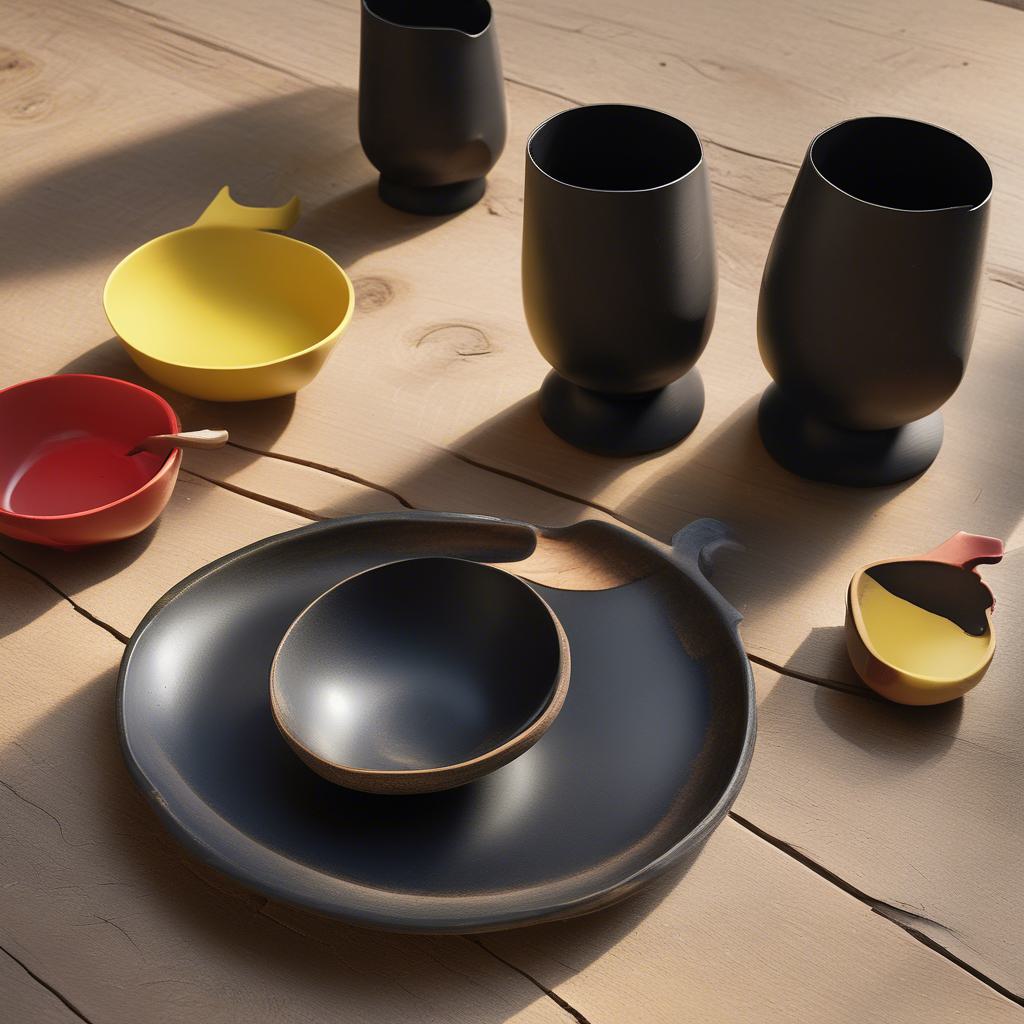} &
        \includegraphics[width=0.16\textwidth]{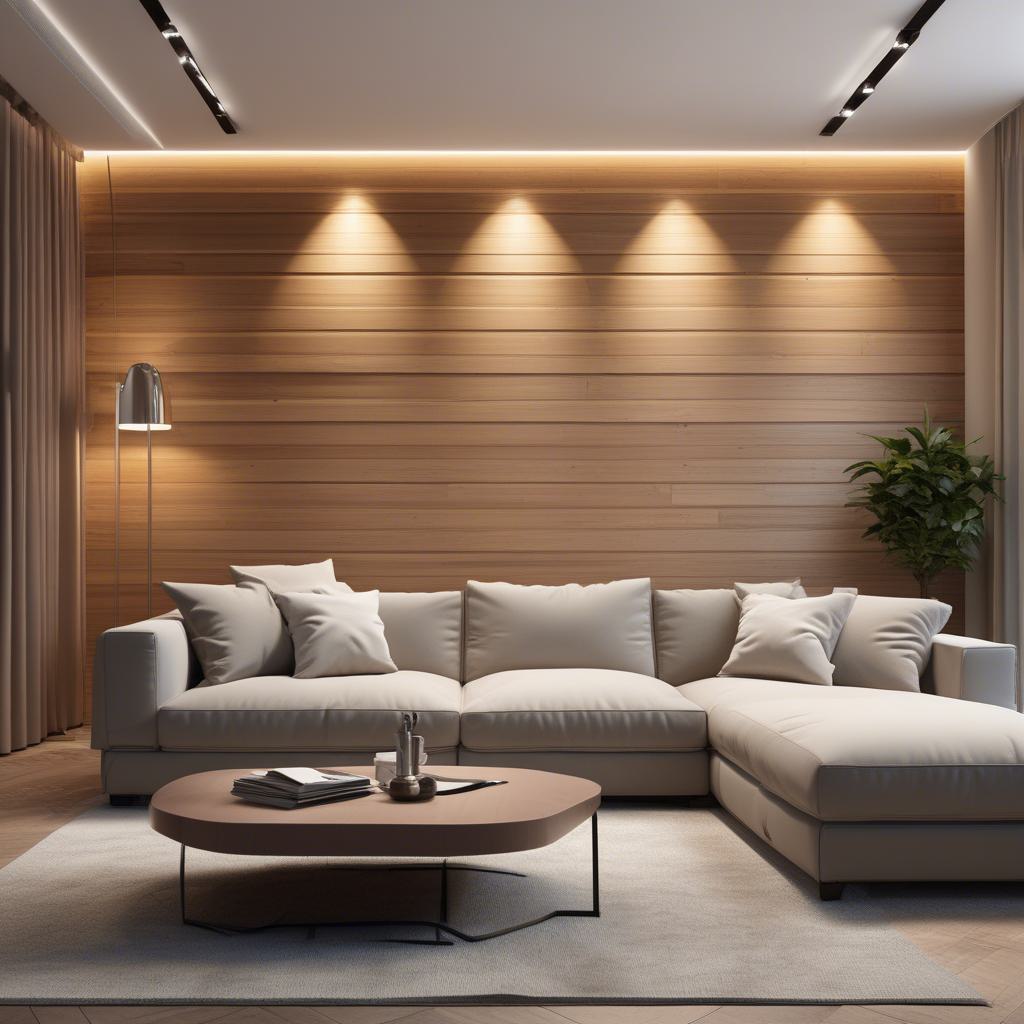} &
        \includegraphics[width=0.16\textwidth]{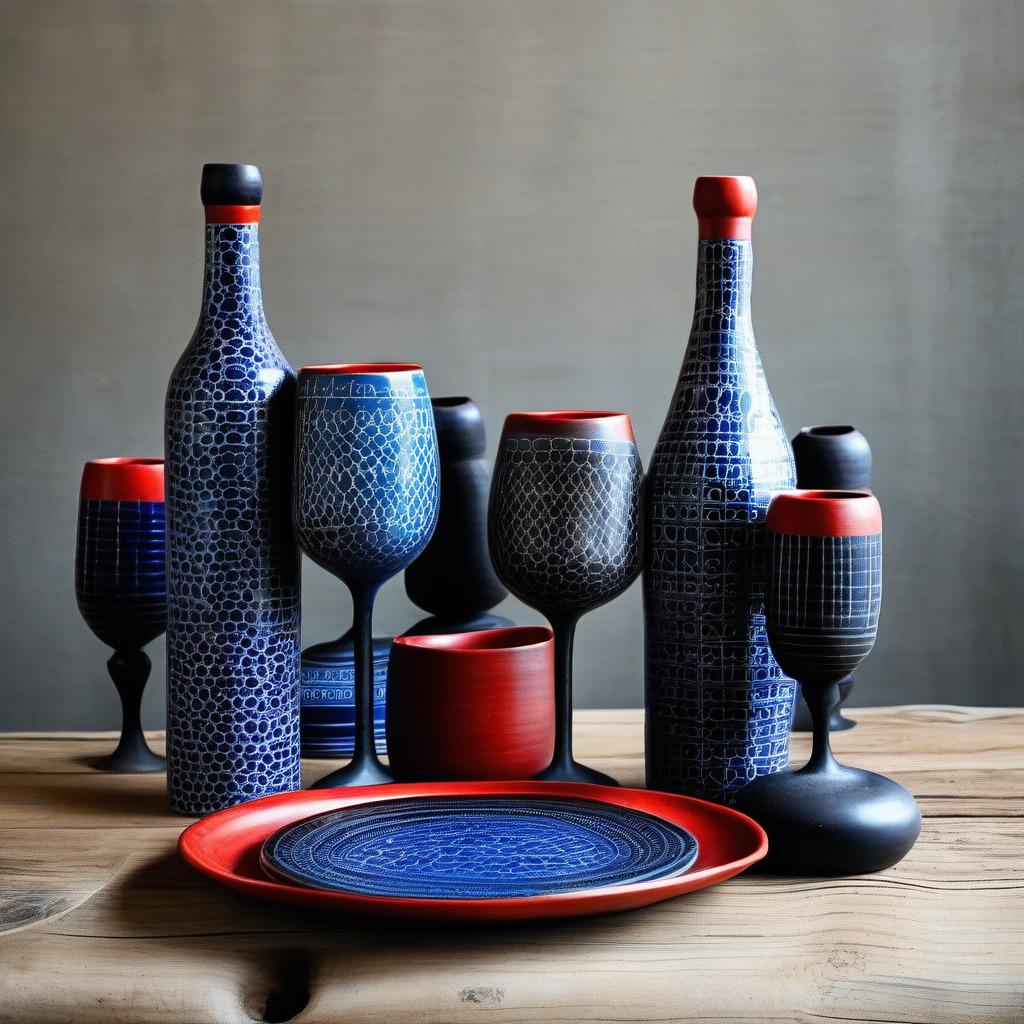} &
        \includegraphics[width=0.16\textwidth]{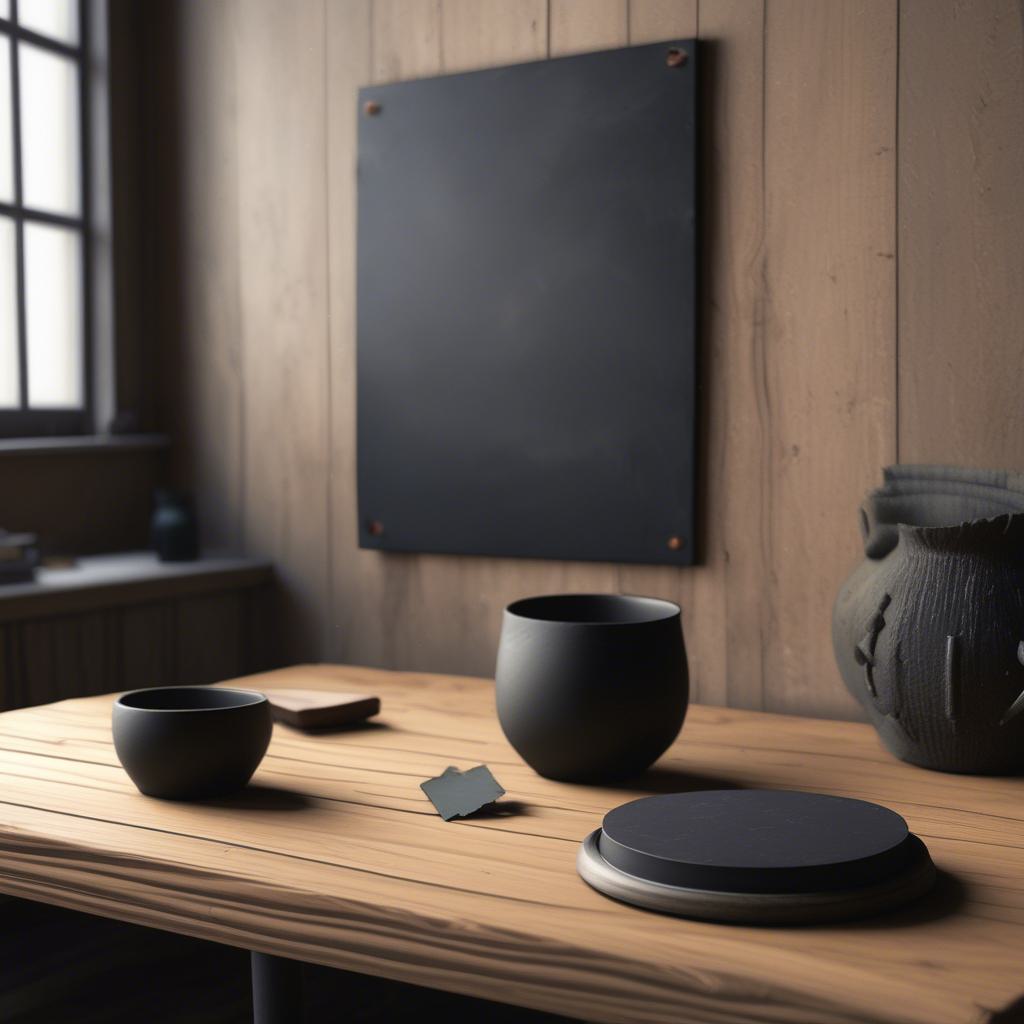} \\
        
        \includegraphics[width=0.16\textwidth]{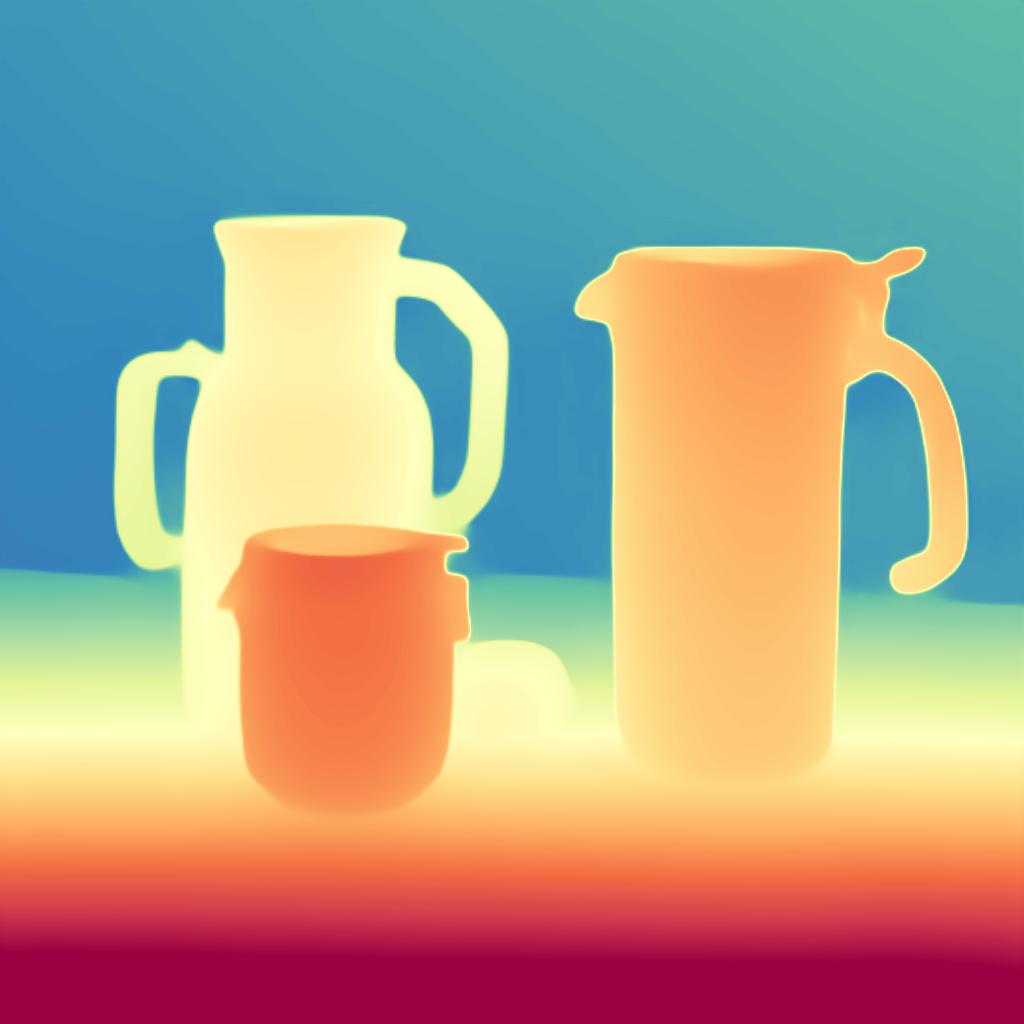} & 
        \includegraphics[width=0.16\textwidth]{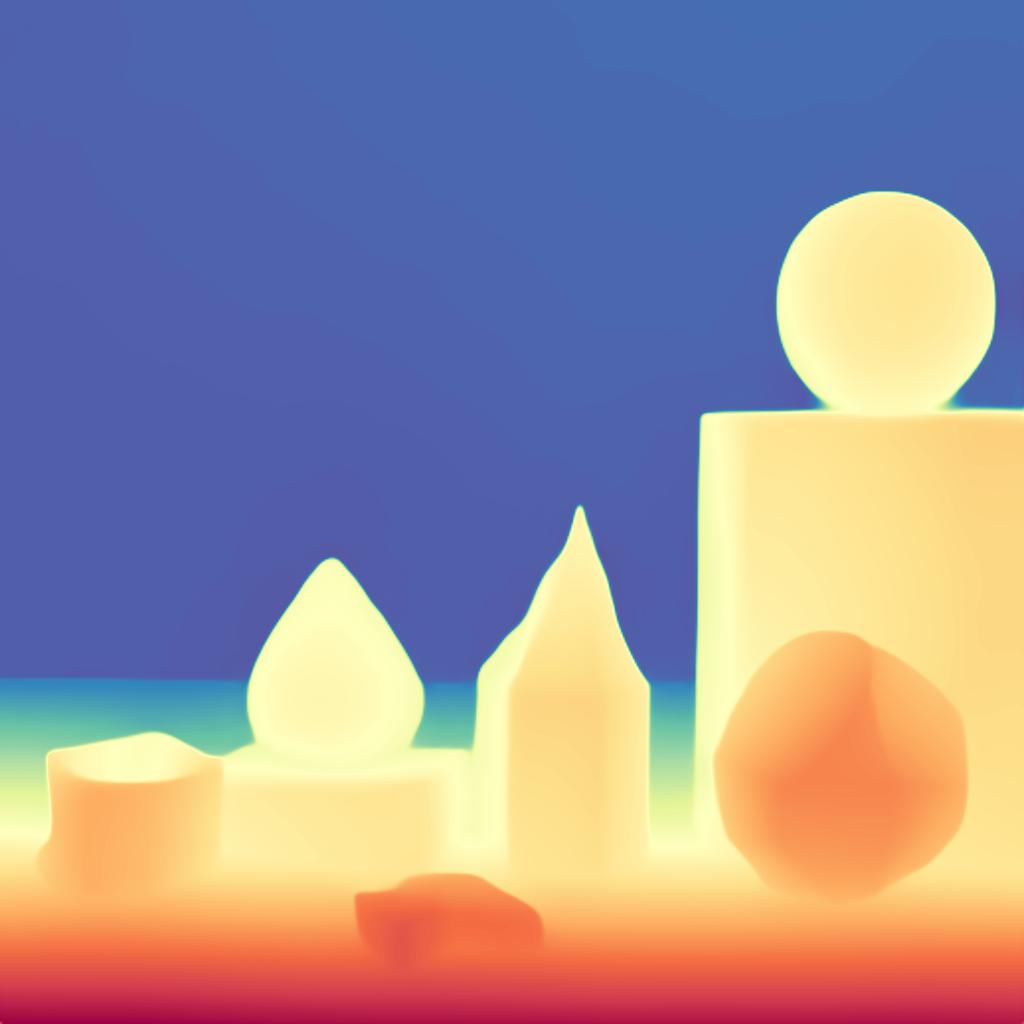} &
        \includegraphics[width=0.16\textwidth]{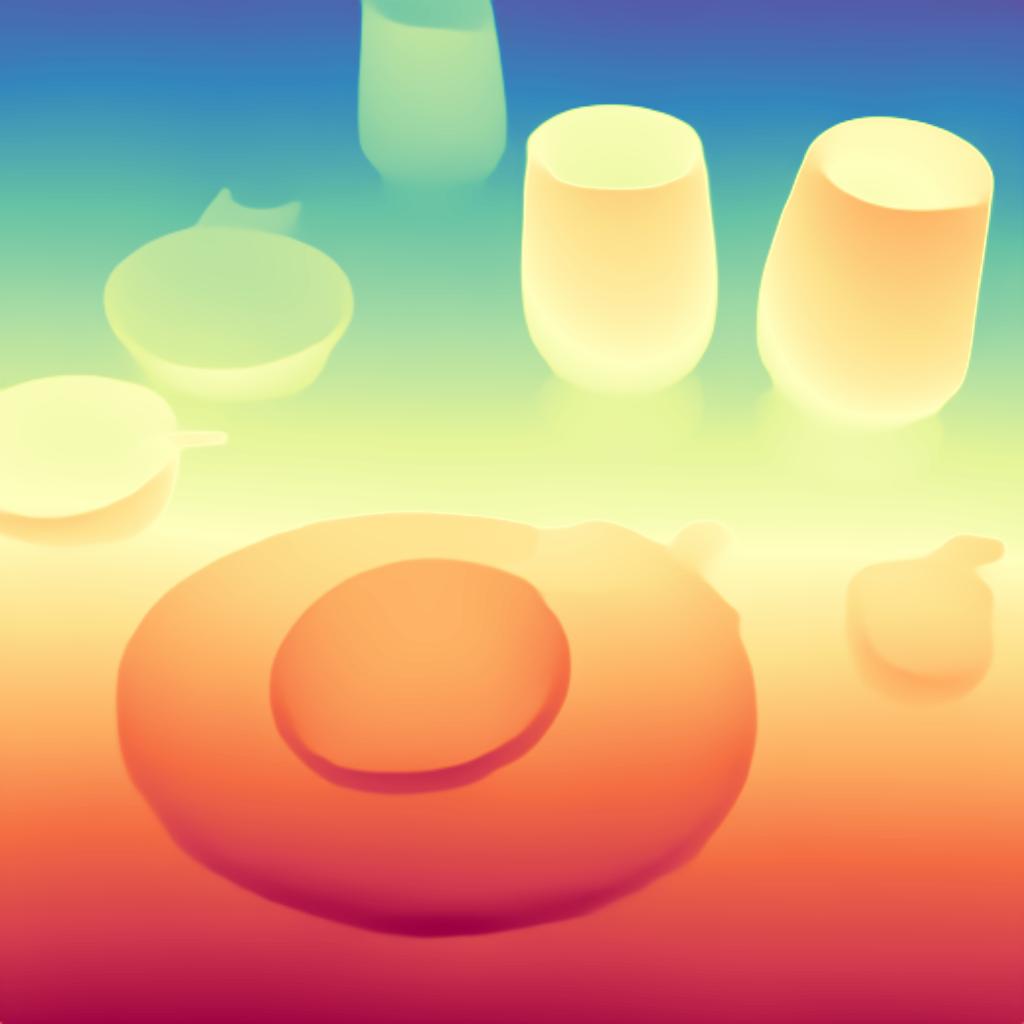} &
        \includegraphics[width=0.16\textwidth]{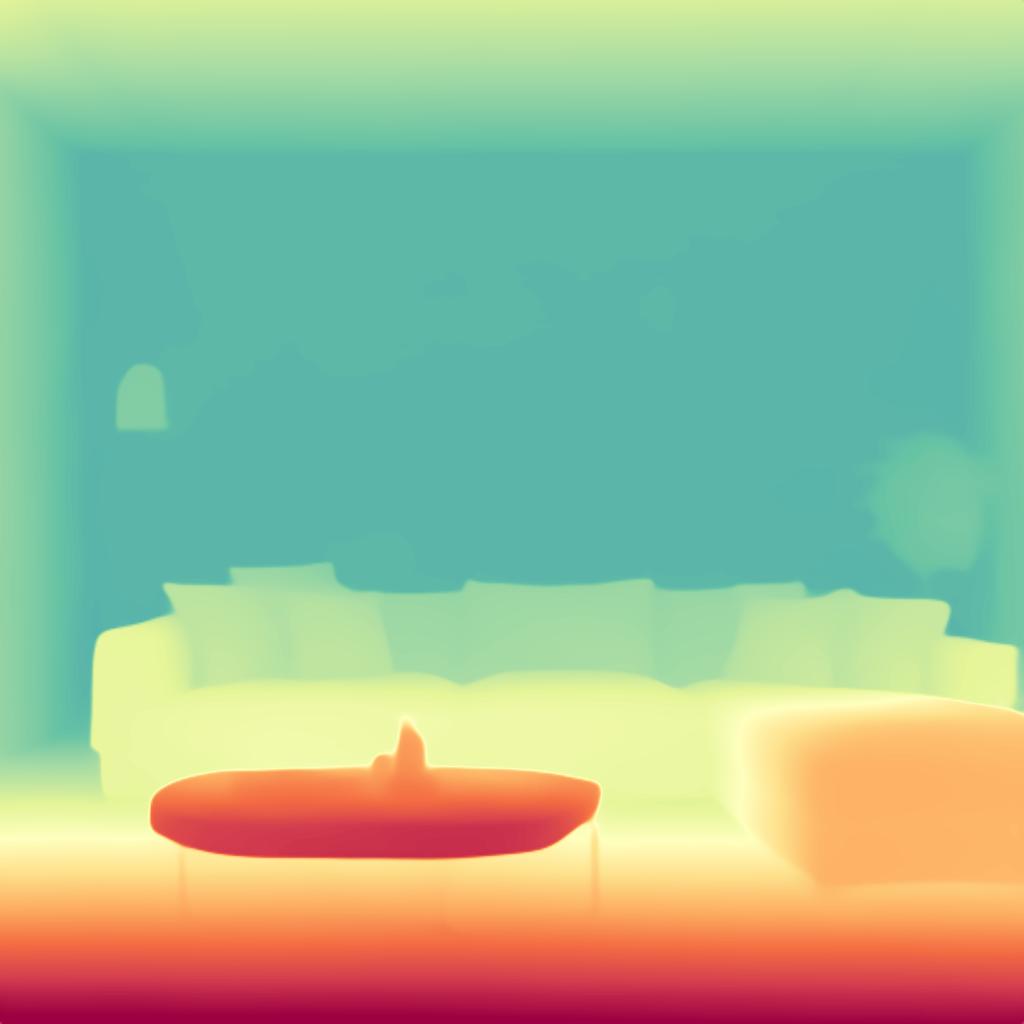} &
        \includegraphics[width=0.16\textwidth]{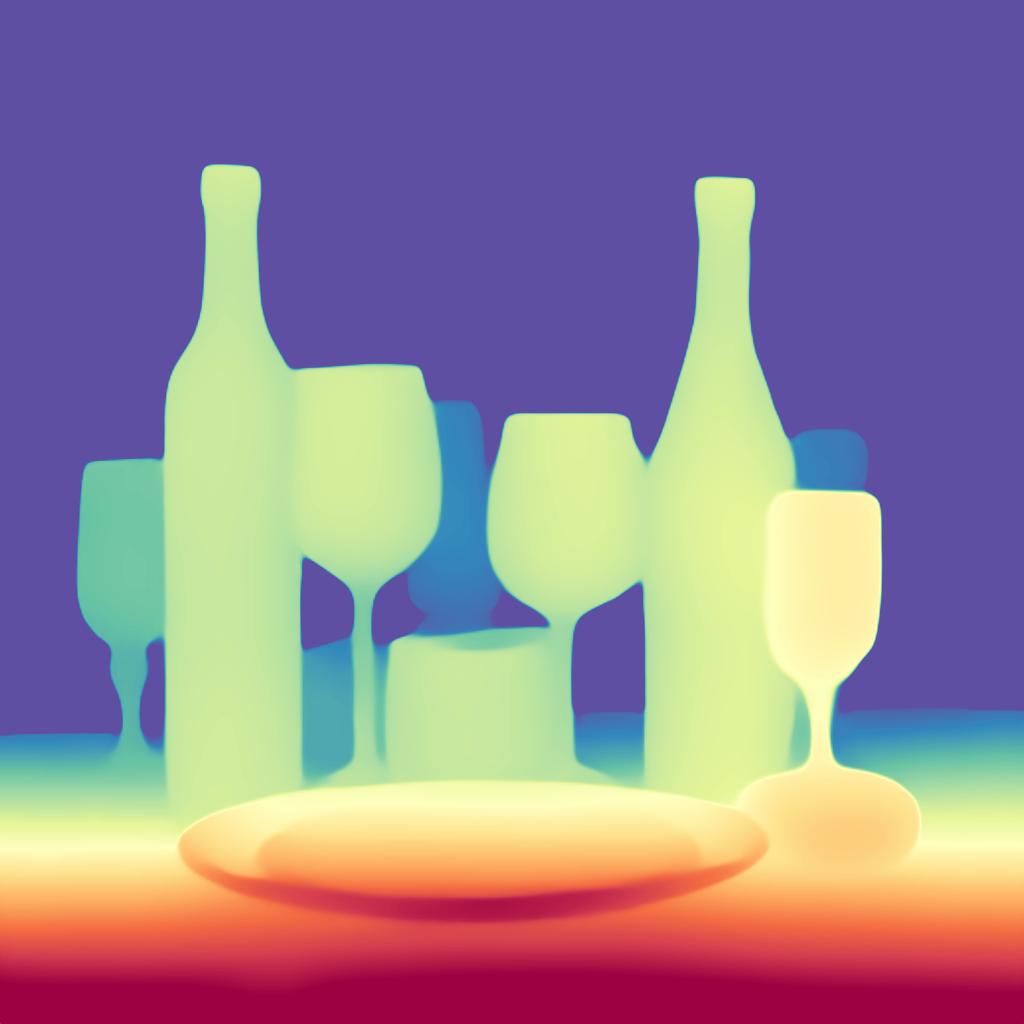} &
        \includegraphics[width=0.16\textwidth]{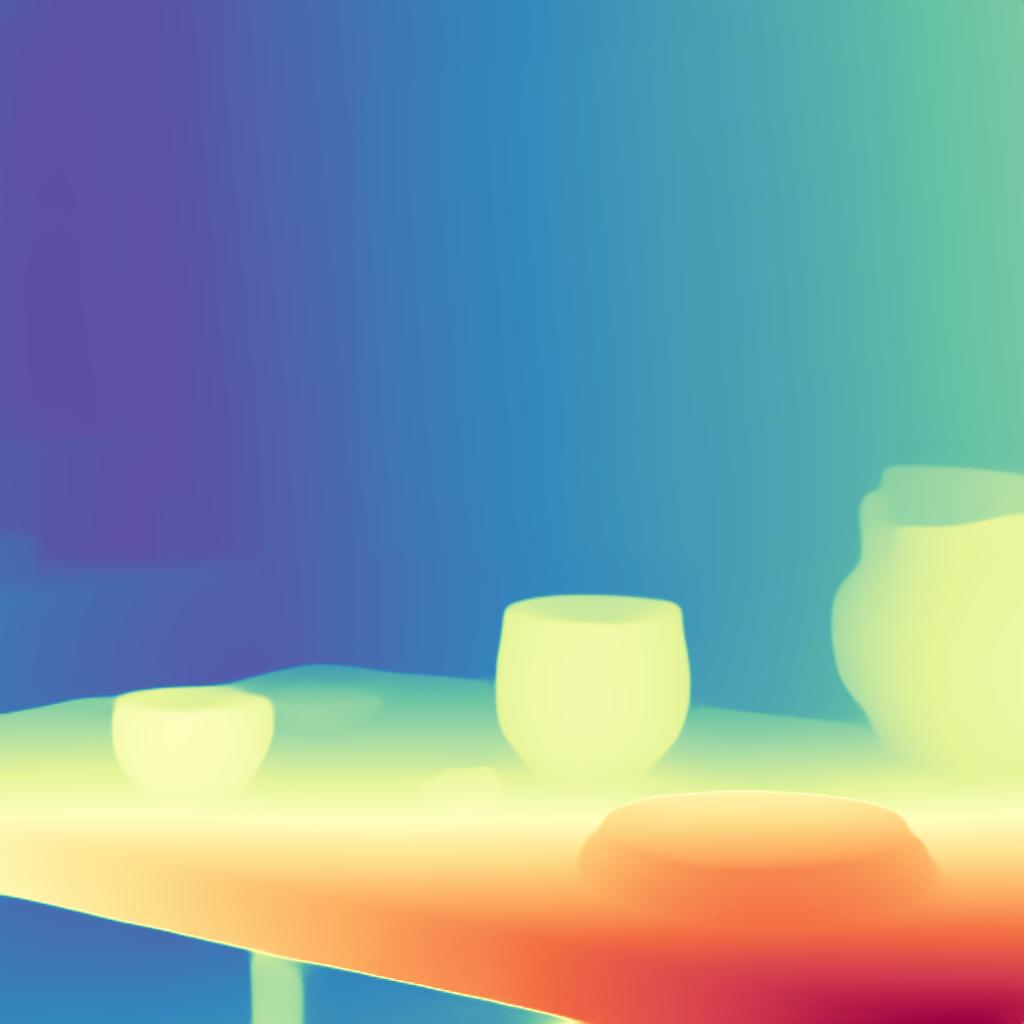} \\

        \includegraphics[width=0.16\textwidth]{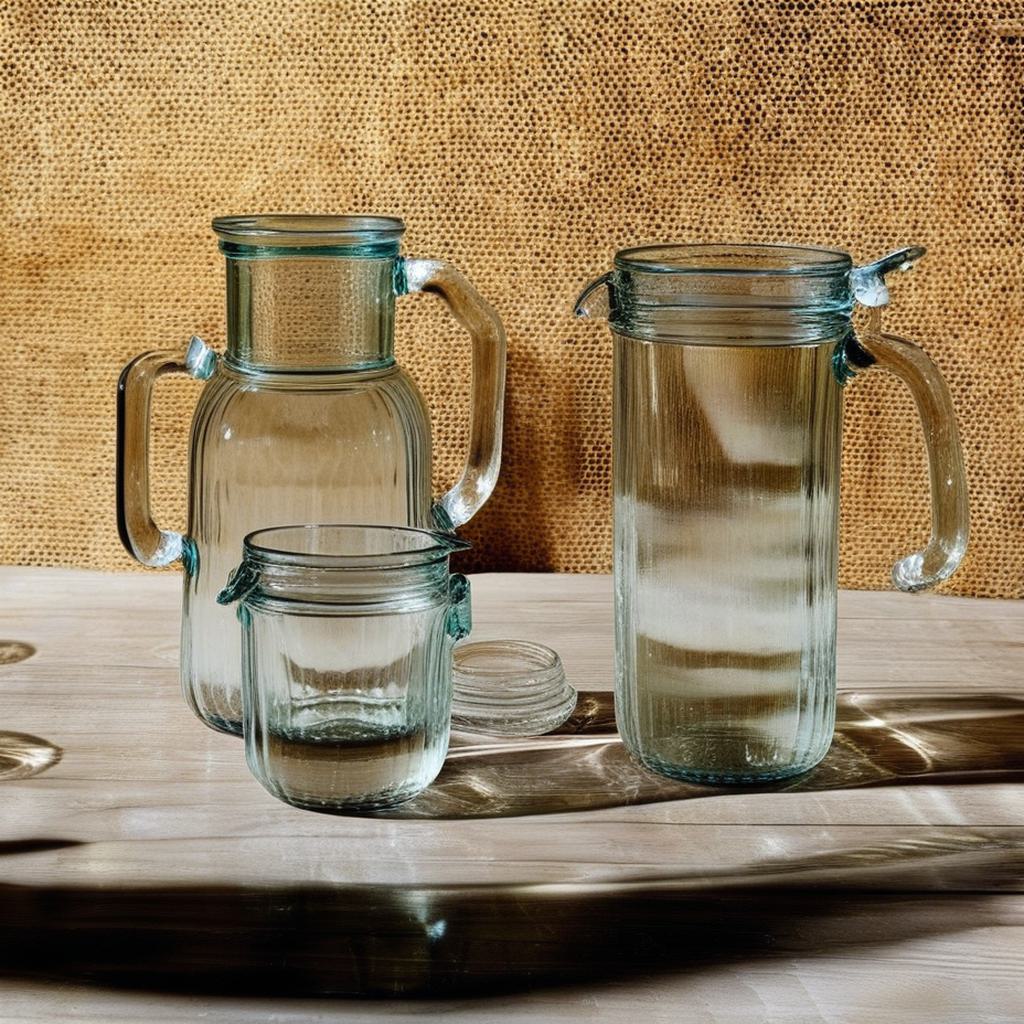} & 
        \includegraphics[width=0.16\textwidth]{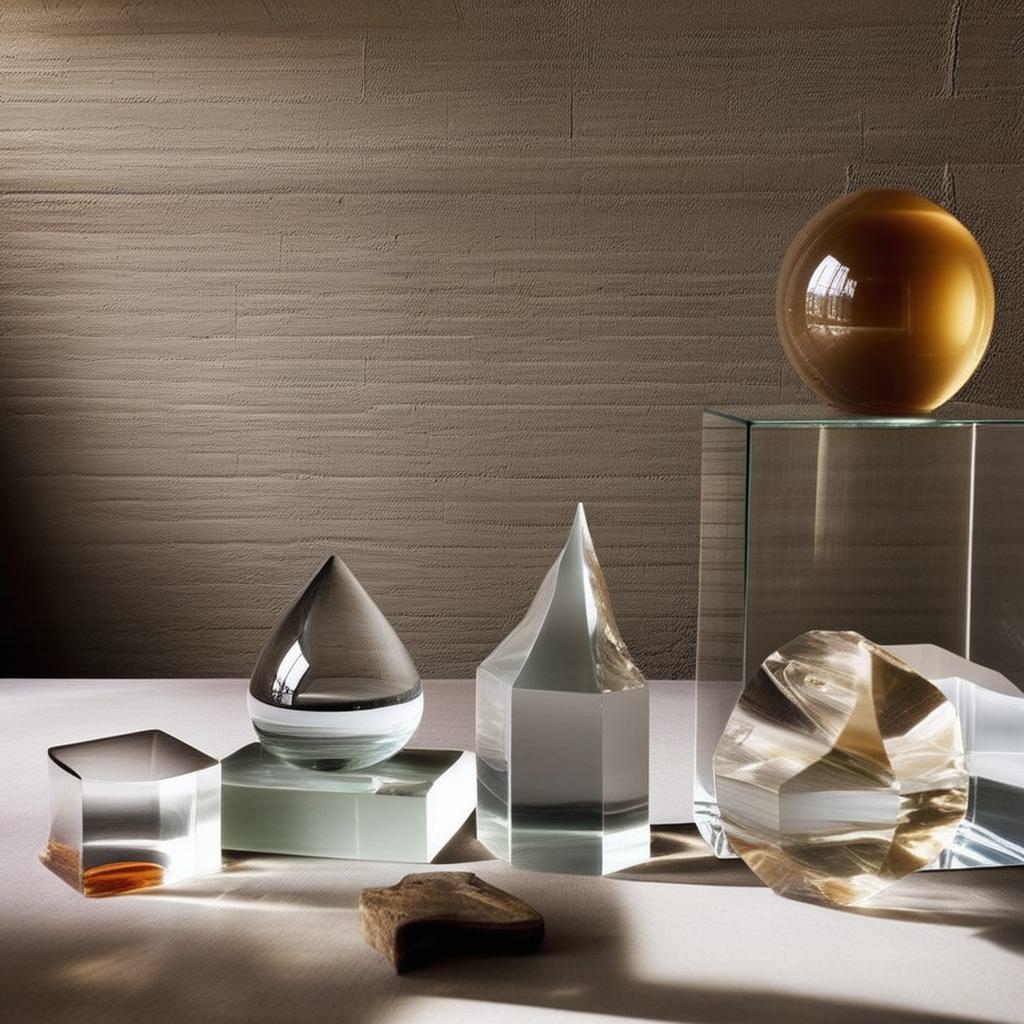} &
        \includegraphics[width=0.16\textwidth]{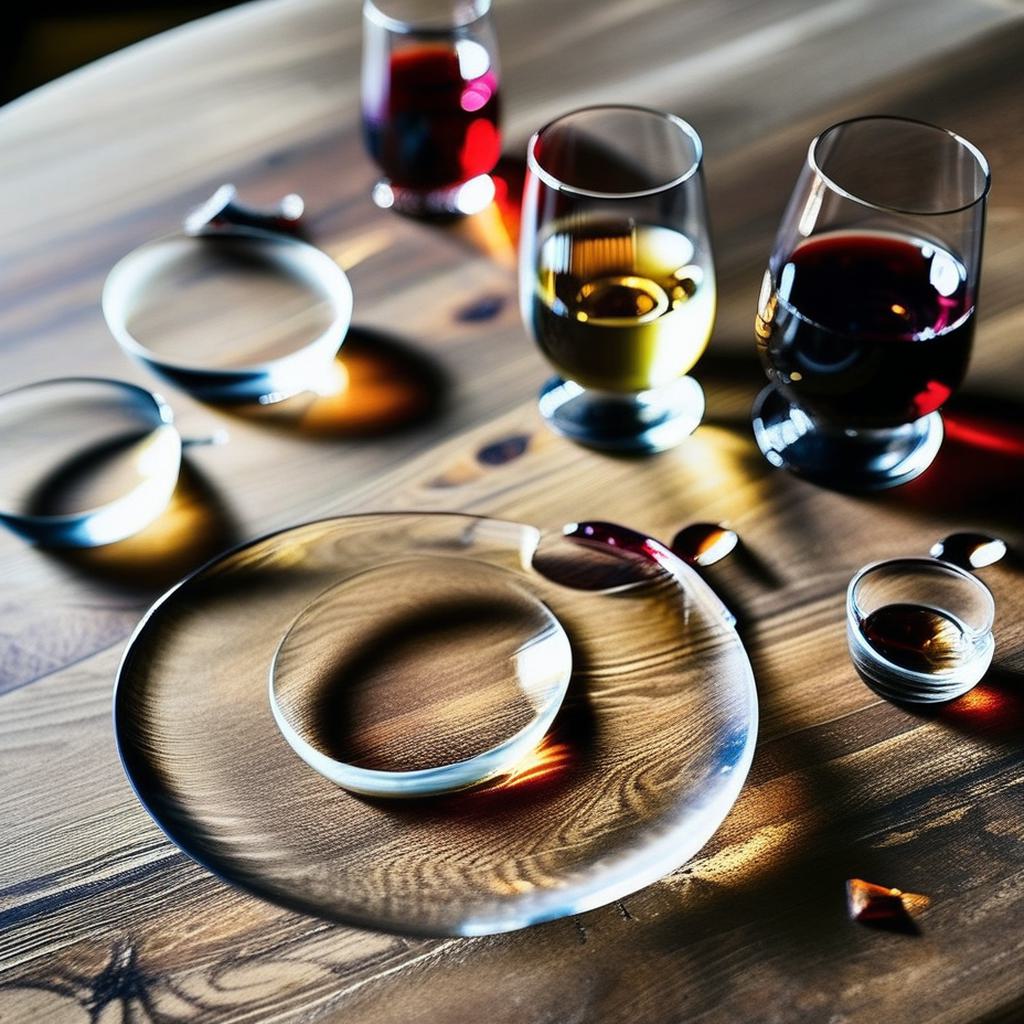} &
        \includegraphics[width=0.16\textwidth]{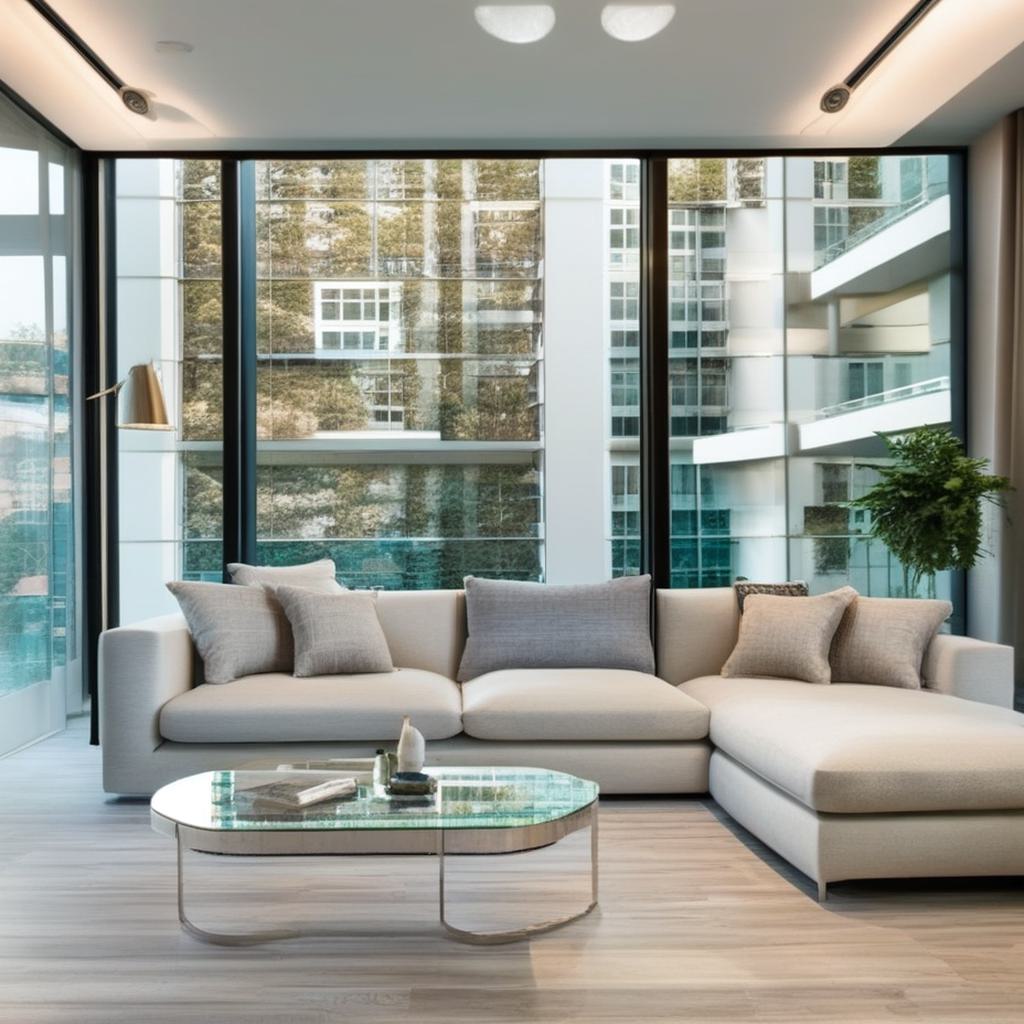} &
        \includegraphics[width=0.16\textwidth]{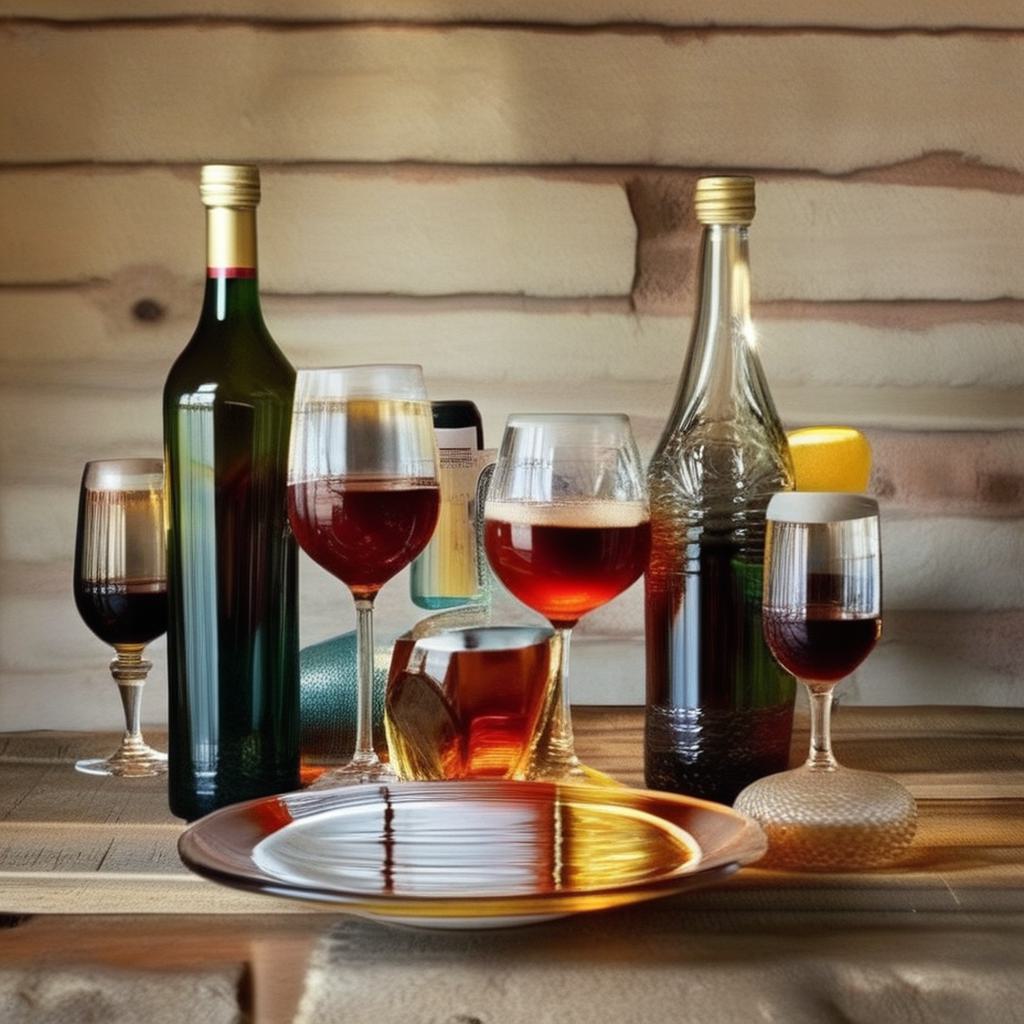} &
        \includegraphics[width=0.16\textwidth]{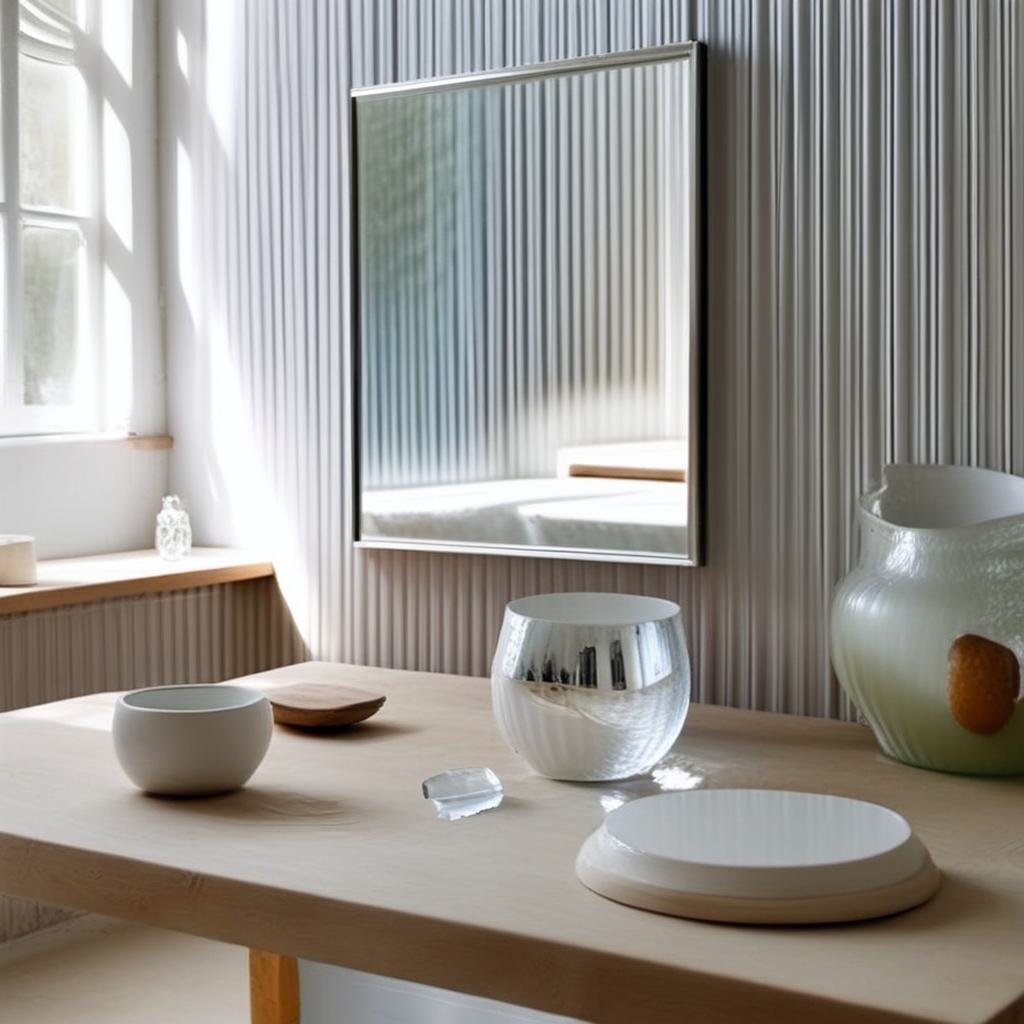} \\
        
    \end{tabular}
    \caption{\textbf{Generated Images -- ToM Objects.} From top to bottom: \textit{easy} scenes from Stable Diffusion \cite{stable-diffusion-xl}, depth from Depth Anything \cite{yang2024depth}, transformed scenes using \cite{mou2023t2i}.}
    \label{fig:ToM_generation}
\end{figure}

\subsection{Diffusion Distilled Data}
\label{sec:diffusion_data}
To create challenging settings from \textit{easy} images, we employ the original code and pre-trained weights provided by T2I-Adapter \cite{mou2023t2i}. 
To ensure fairness with \cite{gasperini_morbitzer2023md4all} on the nuScenes and Robotcar datasets, we exclusively generate night and rain images from the original \textit{easy} ones. This choice is motivated by the fact that these datasets only feature challenging scenarios involving such conditions. It is important to note, however, that the diffusion model can generate data in any challenging scenario, including snow, sun glare, fog, and more.

For data synthesis, we provide the diffusion model with different, random text prompts for night and rain scenarios to ensure diverse environmental variation. 
Consequently, we generate a comparable number of images to \cite{gasperini_morbitzer2023md4all}, totaling 30,258 images for nuScenes and 17,790 images for Robotcar, which we utilize during training. 

In our exploration of challenging surface materials, we also want to highlight the flexibility of diffusion models in generating such representations without relying on real RGB images. To this end, we first use Stable Diffusion \cite{stable-diffusion-xl}, guided by textual prompts, to generate approximately 20K images that resemble Lambertian surfaces such as wooden bottles, ceramic vessels and so on. Then, using \cite{mou2023t2i} along with specific textual prompts designed to transform these standard materials into highly challenging ones, we modify the images to represent a range of non-Lambertian objects. An example of this process is illustrated in Fig. \ref{fig:ToM_generation}. This underlines the adaptability of our approach, and the capability of generative processes to simulate a variety of materials, even in the absence of dedicated datasets. Refer to the \textbf{supplementary material} for additional details. 

\subsection{Training Details}

All experiments are conducted on a single 3090 NVIDIA GPU. For fair comparison, we follow the training/testing protocols and frameworks of \cite{gasperini_morbitzer2023md4all} for adverse weather conditions, and \cite{costanzino2023iccv} for transparent and reflective objects, employing the same monocular networks (md4all for \cite{gasperini_morbitzer2023md4all}, and DPT-Large for \cite{costanzino2023iccv}). We integrate their codebase, substituting their datasets with ours generated using a diffusion model. Furthermore, for extensive experiments involving other networks, such as ZoeDepth \cite{bhat2023zoedepth}, MiDaS \cite{Ranftl2021}, and DPT \cite{Ranftl2022}, we fine-tune for 30K iterations, with an initial learning rate of $10^{-6}$, reduced to $10^{-7}$ after 25K iterations. For Depth Anything \cite{yang2024depth}, we fine-tune for 5K iterations, reducing the learning rate at 4.5K iterations. We use a batch size of 8 by default, except for ZoeDepth, which is set to 3. The AdamW \cite{loshchilov2017decoupled} optimizer is used for all networks. mages are padded, cropped, and resized to maintain 384 pixels for either the long or short side, except for Depth-Anything which uses 518 pixels, preserving aspect ratio with square cropping. We apply data augmentation techniques including color jitter, RGB shift, and horizontal flip, among others. 

\subsection{Adverse Weather Conditions}

\textbf{Improving the Baselines.} \cref{table:baseline_improvement} evaluates monocular depth networks, highlighting their performance across diverse nuScenes \cite{caesar2020nuscenes} scenarios. As baselines, we examine four state-of-the-art methods -- DPT \cite{Ranftl2021}, MiDaS \cite{Ranftl2022}, ZoeDepth \cite{bhat2023zoedepth}, and Depth Anything \cite{yang2024depth} -- known for their strong generalization capabilities, under varying atmospheric conditions: \textit{day-clear}, \textit{night}, and \textit{day-rain}. 

In the table, we present the improvements that our approach yields for each pre-trained model. Specifically, we fine-tune each of the four baselines using our framework and internal protocol. For this purpose, we employ \cite{mou2023t2i} and the baseline depth network itself to create challenging images for the tuning phase. Crucially, the pre-computed depth pseudo-labels of each specific baseline depth network, derived from day-clear images and adopted for their challenging counterpart, remain unchanged and are used directly to minimize loss during subsequent fine-tuning.
It is worth emphasizing again that our methodology relies solely on the availability of simple daytime samples to randomly generate challenging conditions, without relying on any prior knowledge of the target image characteristics of the considered condition (e.g., rain, night). Examining the experimental results reported in \cref{table:baseline_improvement}, it is clear that despite being trained on extensive datasets to generalize across scenarios, the baselines face significant challenges in achieving optimal performance in adverse \textit{day-rain} and \textit{night} scenarios, while performing effectively in the simpler \textit{day-clear} setting. Significantly, our approach consistently outperforms the baselines across all metrics and atmospheric conditions, including both simple and adverse scenarios. This highlights the effectiveness of our methodology in mitigating the complexities associated with this task on state-of-the-art monocular depth networks designed for strong generalization across domains.

\begin{table*}[t]
\centering
\caption{\textbf{Evaluation of monocular networks on the nuScenes \cite{caesar2020nuscenes} validation set.}
Original networks \cite{Ranftl2021, Ranftl2022,bhat2023zoedepth,yang2024depth} versus their fine-tuned versions. 
}
\setlength{\tabcolsep}{12pt}
\scalebox{0.52}{
\begin{tabular}{lrrrrrrrrrr}

\toprule
\multirow{2}{*}{Method}  &\multicolumn{3}{c}{\textit{day-clear}} & \multicolumn{3}{c}{\textit{\textbf{night}}} & \multicolumn{3}{c}{\textit{\textbf{day-rain}}} \\
\cmidrule(lr){2-4} \cmidrule(lr){5-7} \cmidrule(lr){8-10}
& absRel $\downarrow$ & RMSE $\downarrow$ & $\delta_1$ $\uparrow$ & absRel $\downarrow$ & RMSE $\downarrow$ & $\delta_1$ $\uparrow$ & absRel $\downarrow$ & RMSE $\downarrow$ & $\delta_1$ $\uparrow$\\
\hline

MiDaS \cite{Ranftl2021} & 0.171 & 7.703 & 76.75 & 0.261 & 9.729 & 54.66 & 0.218 & 8.823 & 69.39\\

\rowcolor{salmon} MiDaS \cite{Ranftl2021} ft.  Ours  & \textbf{0.168} & \textbf{7.563} & \textbf{76.86} & \textbf{0.254} & \textbf{9.692} & \textbf{63.47} & \textbf{0.195} & \textbf{8.278} & \textbf{72.37}\\

\hline

DPT \cite{Ranftl2022} & 0.189 & 8.094 & 75.39 & 0.354 & 12.875 & 60.97 & 0.237 & 8.780 & 66.96 \\

\rowcolor{salmon} DPT \cite{Ranftl2022} ft. Ours  &  \textbf{0.184} & \textbf{7.839} & \textbf{75.50} & \textbf{0.224} & \textbf{8.375} & \textbf{67.87} & \textbf{0.199} & \textbf{8.079} & \textbf{72.85} \\
\hline

Depth Anything \cite{yang2024depth} & 0.137 & 7.063 & 82.23 & 0.291 & 11.804 & 67.10 & 0.167 & 7.867 & 75.17\\
\rowcolor{salmon} Depth Anything \cite{yang2024depth} ft.  Ours  & \textbf{0.134} & \textbf{6.792} & \textbf{82.53} & \textbf{0.219} & \textbf{ 9.140} & \textbf{70.26} & \textbf{0.157} & \textbf{7.570} & \textbf{77.42}\\

\hline

ZoeDepth \cite{bhat2023zoedepth} & \textbf{0.181} & \textbf{8.517} & \textbf{71.71} & 0.258 &  9.863 & 54.12 & 0.217 & 9.263 & 65.57\\
\rowcolor{salmon} ZoeDepth \cite{bhat2023zoedepth} ft.  Ours  & \textbf{0.181} & 8.946 & 71.43 & \textbf{0.211} & \textbf{9.551} & \textbf{65.69} & \textbf{0.199} & \textbf{9.212} & \textbf{67.65}\\

\hline
\end{tabular}
}

\label{table:baseline_improvement}
\end{table*}

\begin{table*}[t]
\centering
\caption{\textbf{Evaluation of monocular networks on the nuScenes \cite{caesar2020nuscenes} validation set.} Supervisions (sup.): M: monocular videos, S: singe-view images, $*$: test-time median-scaling via LiDAR, v: weak velocity, r: weak radar. Training data (tr. data): \textit{d: \textit{day-clear}, T: Translated in, n: night} (including \textit{night-rain}), r: \textit{day-rain}, a: all. Target Condition (T. Cond.): target atmospheric condition images known in advance. \textdagger: 
depth networks exclusively used within the diffusion model and not employed in the fine-tuning phase. We highlight the \textbf{1st} and \underline{2nd} absolute bests.}
\setlength{\tabcolsep}{6pt}
\scalebox{0.50}{
\begin{tabular}{lcrrrrrrrrrrr}
\toprule
\multirow{2}{*}{Method} & \multirow{2}{*}{T. Cond.} & \multirow{2}{*}{sup.} & \multirow{2}{*}{tr. data} & \multicolumn{3}{c}{\textit{day-clear}} & \multicolumn{3}{c}{\textit{\textbf{night}}} & \multicolumn{3}{c}{\textit{\textbf{day-rain}}} \\
\cmidrule(lr){5-7} \cmidrule(lr){8-10} \cmidrule(lr){11-13}
 & & & & absRel $\downarrow$ & RMSE $\downarrow$ & $\delta_1$ $\uparrow$ & absRel $\downarrow$ & RMSE $\downarrow$ & $\delta_1$ $\uparrow$ & absRel $\downarrow$ & RMSE $\downarrow$ & $\delta_1$ $\uparrow$ \\
\hline
\multicolumn{13}{c}{In-Domain} \\
\hline
\addlinespace
R4Dyn w/o r in \cite{gasperini2021r4dyn} & \xmark & Mvr & d & 0.130 & 6.536 & 85.76 & 0.273 & 12.430 & 52.85 & 0.147 & 7.533 & 80.59 \\
R4Dyn \cite{gasperini2021r4dyn}  (radar) & \xmark & Mvr & d & \textbf{0.126} & \textbf{6.434} & \textbf{86.97} & \textbf{0.219} & \textbf{10.542} & \textbf{62.28} & \textbf{0.134} & \textbf{7.131} & \textbf{83.91} \\
\addlinespace
\hdashline
\addlinespace
Monodepth2 \cite{monodepth2} & \xmark & M* & d & 0.137 & 6.692 & 85.00 & 0.283 & 9.729 & 51.83 & 0.173 & 7.743 & 77.57 \\
PackNet-SfM \cite{guizilini20203d} & \xmark & Mv & d & 0.157 & 7.230 & 82.64 & 0.262 & 11.063 & 56.64 & 0.165 & 8.288 & 77.07 \\
md4all (\textbf{baseline}) \cite{gasperini_morbitzer2023md4all} & \xmark & Mv & d & \underline{0.133} & 6.459 & \textbf{85.88} & 0.242 & 10.922 & 58.17 & 0.157 & 7.453 & 79.49 \\
\rowcolor{salmon} md4all-DD \cite{gasperini_morbitzer2023md4all} ft. Ours & \xmark & Mv & dT(nr) & 0.137 & \textbf{6.318}  & \underline{85.05} & \textbf{0.188} & \textbf{8.432} & \underline{69.94} & 0.147 & 7.345 & 79.59\\
\rowcolor{salmon} md4all-DD \cite{gasperini_morbitzer2023md4all} ft. Ours (DPT\textdagger) & \xmark & Mv & dT(nr) & 0.140 & 6.573 & 83.51 & 0.197 & 8.826 & 69.65 & \underline{0.143} & \underline{7.317} & \underline{80.28}\\

\rowcolor{salmon} md4all-DD \cite{gasperini_morbitzer2023md4all} ft. Ours (Depth Anything  \textdagger) & \xmark & Mv & dT(nr) & \textbf{0.128} & \underline{6.449}  & 84.03 & \underline{0.191} & \underline{8.433} & \textbf{71.14} & \textbf{0.139} & \textbf{7.129} & \textbf{81.36}\\

\addlinespace
\hdashline
\addlinespace
Monodepth2 \cite{monodepth2} & \checkmark & M* & a: dnr & \underline{0.148} & \underline{6.771} & \textbf{85.25} & 2.333 & 32.940 & 10.54 & 0.411 & 9.442 & 60.58 \\
RNW \cite{wang2021regularizing} & \checkmark & M* & dn & 0.287 & 9.185 & 56.21 & 0.333 & 10.098 & 43.72 & 0.295 & 9.341 & 57.21 \\
md4all-AD \cite{gasperini_morbitzer2023md4all} ft. Gasperini et al.\cite{gasperini_morbitzer2023md4all}  & \checkmark & Mv & dT(nr) & 0.152 & 6.853 & 83.11 & \underline{0.219} & \underline{9.003} & \underline{68.84} & \underline{0.160} & \underline{7.832} & \underline{78.97} \\
md4all-DD \cite{gasperini_morbitzer2023md4all} ft. Gasperini et al.\cite{gasperini_morbitzer2023md4all}  & \checkmark & Mv & dT(nr) & \textbf{0.137} & \textbf{6.452} & \underline{84.61} & \textbf{0.192} & \textbf{8.507} & \textbf{71.07} & \textbf{0.141} & \textbf{7.228} & \textbf{80.98} \\
\addlinespace

\bottomrule
\end{tabular}
}
\label{table:depth_estimation_nuscenes}
\end{table*}

\begin{table*}[t]
\centering
\caption{\textbf{Evaluation of monocular depth frameworks on the RobotCar \cite{maddern20171} test set.} We follow the same notation provided in \cref{table:depth_estimation_nuscenes}.}
\setlength{\tabcolsep}{4pt}
\scalebox{0.55}{
\begin{tabular}{lccccrrrrrrrr}
\toprule
\multirow{2}{*}{Method} & \multirow{2}{*}{T. Cond} & \multirow{2}{*}{Source} & \multirow{2}{*}{Sup.} & \multirow{2}{*}{Tr. Data} & \multicolumn{4}{c}{\textit{day-clear} } & \multicolumn{4}{c}{\textit{\textbf{night}} } \\
\cmidrule(lr){6-9} \cmidrule(lr){10-13}
 & & & & &  absRel $\downarrow$ & sqRel $\downarrow$ & RMSE $\downarrow$ & $\delta_1$ $\uparrow$ & absRel $\downarrow$ & sqRel $\downarrow$ & RMSE $\downarrow$ & $\delta_1$ $\uparrow$ \\
\midrule
DeFeatNet \cite{spencer2020defeat} & \checkmark  & \cite{spencer2020defeat} & M* & a: dn & 0.247 & 2.980 & 7.884 & 65.00 & 0.334 & 4.589 & 8.606 & 58.60 \\
ADIDS \cite{liu2021self} & \checkmark & \cite{vankadari2023sun} & M* & a: dn & 0.239 & 2.089 & 6.743 & 61.40 & 0.287 & 2.569 & 7.985 & 49.00 \\
RNW \cite{wang2021regularizing} & \checkmark &  \cite{vankadari2023sun} & M* & a: dn & 0.297 & 2.608 & 7.996 & 43.10 & 0.185 & 1.710 & 6.549 & 73.30 \\
WSGD \cite{vankadari2023sun} & \checkmark & \cite{vankadari2023sun} & M* & a: dn & \underline{0.176} & \underline{1.603} & \underline{6.036} & \underline{75.00} & \underline{0.174} & \underline{1.637} & \underline{6.302} & \underline{75.40} \\
md4all-DD \cite{gasperini_morbitzer2023md4all} ft. Gasperini et al.\cite{gasperini_morbitzer2023md4all}  & \checkmark & \cite{gasperini_morbitzer2023md4all} & Mv & dT(n) & \textbf{0.113} & \textbf{0.648} & \textbf{3.206} & \textbf{87.13} & \textbf{0.122} & \textbf{0.739} & \textbf{3.604} & \underline{\textbf{84.86}} \\
\addlinespace
\hdashline
\addlinespace
Monodepth2 \cite{monodepth2} & \xmark & \cite{gasperini_morbitzer2023md4all} & M* & d & \textbf{0.112} & \textbf{0.670} & \textbf{3.164} & 86.38 & 0.303 & 1.724 & 5.038 & 45.88 \\
md4all (\textbf{baseline}) \cite{gasperini_morbitzer2023md4all} & \xmark & \cite{gasperini_morbitzer2023md4all} & Mv & d & 0.121 & 0.723 & 3.335 & 86.61 & 0.391 & 3.547 & 8.227 & 22.51 \\

\rowcolor{salmon} md4all-DD \cite{gasperini_morbitzer2023md4all} ft. Ours & \xmark & \cite{gasperini_morbitzer2023md4all} & Mv & dT(n) &  \underline{0.119} & \underline{0.676} &  \underline{3.239} & \textbf{87.20} & 0.139 & \textbf{0.739} & \underline{3.700} & 82.46 \\
\rowcolor{salmon} md4all-DD \cite{gasperini_morbitzer2023md4all} ft. Ours (DPT \textdagger) & \xmark & \cite{gasperini_morbitzer2023md4all} & Mv & dT(n) & 0.123 & 0.724 & 3.333 & 86.62 & \underline{0.133} & 0.824 & 3.712 & \textbf{83.95} \\
\rowcolor{salmon} md4all-DD \cite{gasperini_morbitzer2023md4all} ft. Ours (Depth Anything \textdagger) & \xmark & \cite{gasperini_morbitzer2023md4all} & Mv & dT(n) &  \underline{0.119} & 0.728 &  3.287  & \underline{87.17} & \textbf{0.129} & \underline{0.751} & \textbf{3.661} & \underline{83.68} \\

\bottomrule

\end{tabular}
}

\label{table:depth_estimation_robotcar}

\end{table*}

\textbf{Comparison with Existing Methods.} In Tables \ref{table:depth_estimation_nuscenes} and \ref{table:depth_estimation_robotcar}, we compare our approach to existing methods for single-image depth estimation, particularly those addressing challenging conditions. We categorize these methods based on their reliance on prior knowledge of the target image characteristics for a specific condition (\textit{T. Cond.} flag in the tables).
Monodepth2 \cite{monodepth2} and PackNet \cite{guizilini20203d} struggle to achieve satisfactory results if trained solely on \textit{day-clear} images or incorporating all atmospheric conditions in nuScenes. Conversely, approaches that augment the inputs with additional information, such as R4Dyn \cite{gasperini2021r4dyn} using radar data, show improvements, while methods such as that of Gasperini et al. \cite{gasperini_morbitzer2023md4all} -- which uses a ForkGAN \cite{zheng2020forkgan} to transform all \textit{day-clear} training samples into rainy and nocturnal environments -- show superior performance in specific configurations (md4all-AD and md4all-DD) over the self-supervised md4all baseline model \cite{gasperini_morbitzer2023md4all}. Nevertheless, it is important to emphasize that Gasperini et al. \cite{gasperini_morbitzer2023md4all}, in order to achieve such results, rely on prior knowledge of specific image properties (such as noise, luminosity, etc.) for the considered atmospheric conditions in the target dataset. This underscores the stringent requirement for their approach to have images in the desired adverse environment. Our method, instead, significantly enhances the effectiveness of the self-supervised md4all baseline architecture, attaining comparable or even superior performance to Gasperini et al. \cite{gasperini_morbitzer2023md4all}, relying exclusively on the availability of easy samples.
Importantly, our approach allows challenging images to be generated for fine-tuning by T2I-Adapter \cite{mou2023t2i} using user text prompts and depth maps from various sources: md4all, advanced networks like DPT, or Depth Anything applied to \textit{day-clear} images. This choice can be exploited based on T2I-Adapter's built-in compatibility with depth maps derived from these models. Nevertheless, the depth pseudo-labels are consistently derived from the md4all-baseline network during the fine-tuning process, and thus do not take advantage of other external depth annotations.  
We argue using DPT or Depth Anything for the conditional diffusion model does not yield an unfair advantage to our method: according to \cref{table:baseline_improvement}, these networks, at best, achieve (in generalization) comparable results to those trained specifically on nuScenes, like md4all-baseline. Notably, our results show improvements across challenging conditions regardless of the depth maps used for image generation. 
This trend is consistent in \cref{table:depth_estimation_robotcar}: our approach, without prior knowledge of the challenging images, outperforms others specialized in day-to-night translation and is comparable to md4all-DD \cite{gasperini_morbitzer2023md4all}.

\begin{table}[t]
\centering

\caption{\textbf{Performance comparison of DPT \cite{Ranftl2022} and Depth Anything \cite{yang2024depth} pre- and post-fine-tuning.} Challenging images generated via \cite{mou2023t2i} from samples in datasets with only \textit{easy} conditions: Mapillary, Cityscapes, KITTI, and Apolloscapes.}
\setlength{\tabcolsep}{12pt}
\scalebox{0.6}{
\begin{tabular}{lcccccccc}
\toprule
\multirow{2}{*}{} &  \multicolumn{2}{c}{DrivingStereo \cite{yang2019drivingstereo}} & \multicolumn{4}{c}{nuScenes \cite{caesar2020nuscenes}} & \multicolumn{2}{c}{RobotCar \cite{maddern20171}} \\

\multirow{2}{*}{Method} & \multicolumn{2}{c}{\textit{\textbf{day-rain}}} & \multicolumn{2}{c}{\textit{\textbf{night}} } & \multicolumn{2}{c}{\textit{\textbf{day-rain}}} & \multicolumn{2}{c}{\textit{\textbf{night}}} \\
\cmidrule(lr){2-3} \cmidrule(lr){4-5} \cmidrule(lr){6-7} \cmidrule(lr){8-9}
 & absRel $\downarrow$  & $\delta_1$ $\uparrow$ & absRel $\downarrow$ & $\delta_1$ $\uparrow$ & absRel $\downarrow$ & $\delta_1$ $\uparrow$ & absRel $\downarrow$ & $\delta_1$ $\uparrow$\\ \hline
 DPT (\textbf{baseline}) \cite{Ranftl2022}  & 0.188 & 0.700 & 0.354 & 60.97  & 0.237 &  66.96 & 0.154 & 83.40 \\
 ft. Gasperini et al.\cite{gasperini_morbitzer2023md4all}& \textcolor{red}{\xmark} & \textcolor{red}{\xmark} & \textcolor{red}{\xmark} & \textcolor{red}{\xmark} & \textcolor{red}{\xmark} & \textcolor{red}{\xmark} & \textcolor{red}{\xmark} & \textcolor{red}{\xmark} \\

 \rowcolor{salmon} ft. Ours & \textbf{0.124} & \textbf{0.836} & \textbf{0.263} & \textbf{67.39} & \textbf{0.202} & \textbf{70.38} & \textbf{0.130} & \textbf{86.60}\\

 \hline

Depth Anything (\textbf{baseline}) \cite{yang2024depth}  & 0.112 & 0.854 & 0.291 & 67.10  & 0.167 & 75.17 & 0.125 & 87.15 \\
ft. Gasperini et al.\cite{gasperini_morbitzer2023md4all} & \textcolor{red}{\xmark} & \textcolor{red}{\xmark} & \textcolor{red}{\xmark} & \textcolor{red}{\xmark} & \textcolor{red}{\xmark} & \textcolor{red}{\xmark} & \textcolor{red}{\xmark} & \textcolor{red}{\xmark} \\
\rowcolor{salmon} ft. Ours & \textbf{0.110} & \textbf{0.868} & \textbf{0.250} & \textbf{70.38} & \textbf{0.154} & \textbf{78.86} & \textbf{0.117} & \textbf{88.18}\\
\hline
\end{tabular}
}
\label{tab:generalization_experiment}
\end{table}

\begin{figure}[t]
    \centering
    \renewcommand{\tabcolsep}{1pt}
    \begin{tabular}{cccc}
        \scriptsize DrivingStereo (day-rain)
        &
        \scriptsize nuScenes (day-rain)
        &
        \scriptsize nuScenes (night) 
        &
        \scriptsize RobotCar (night)
        \\
        \includegraphics[trim=2cm 0cm 2cm 0cm,clip,height=0.125\textwidth]{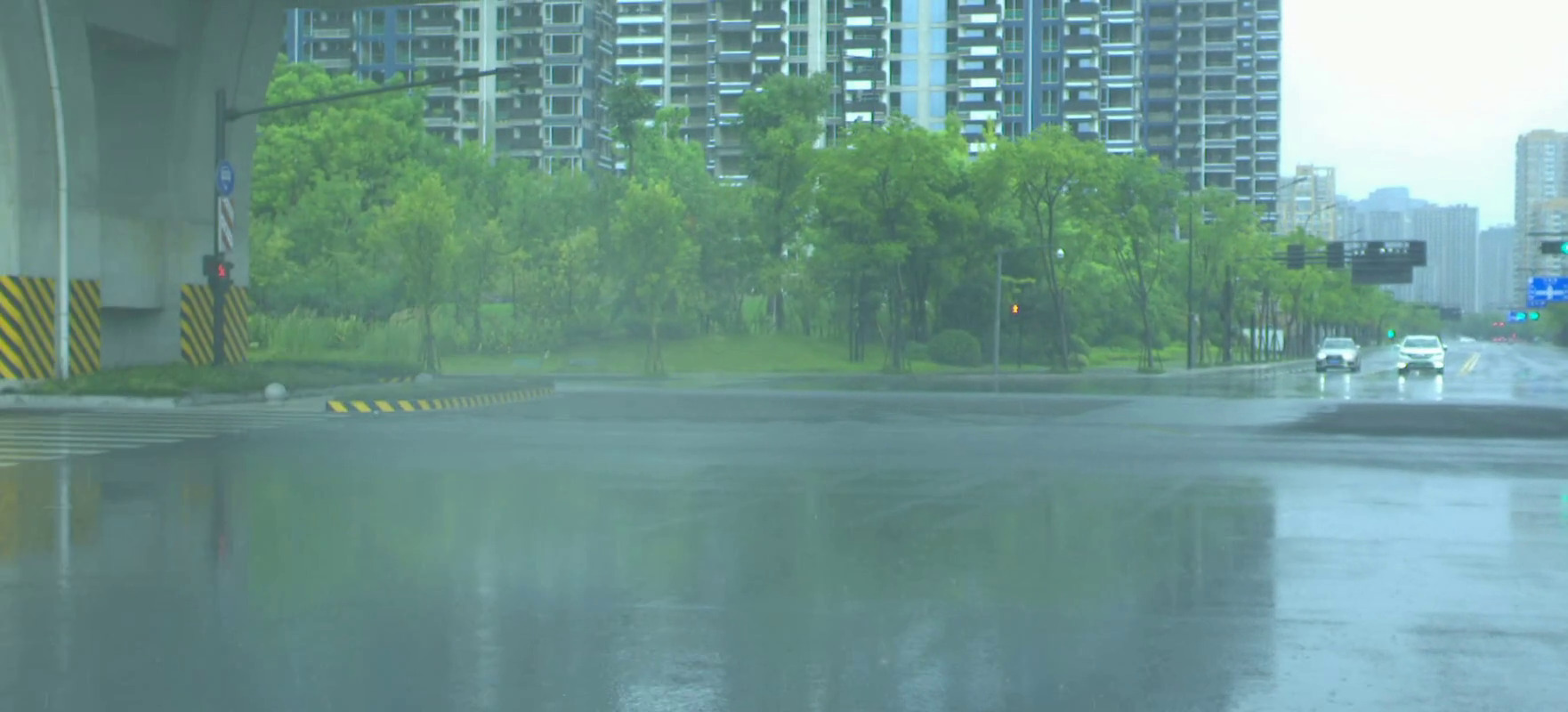} &
        \includegraphics[height=0.125\textwidth]{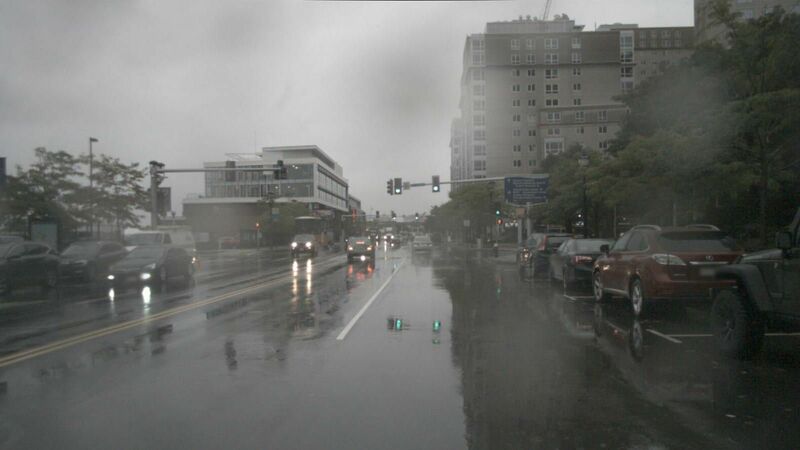} &
        \includegraphics[height=0.125\textwidth]{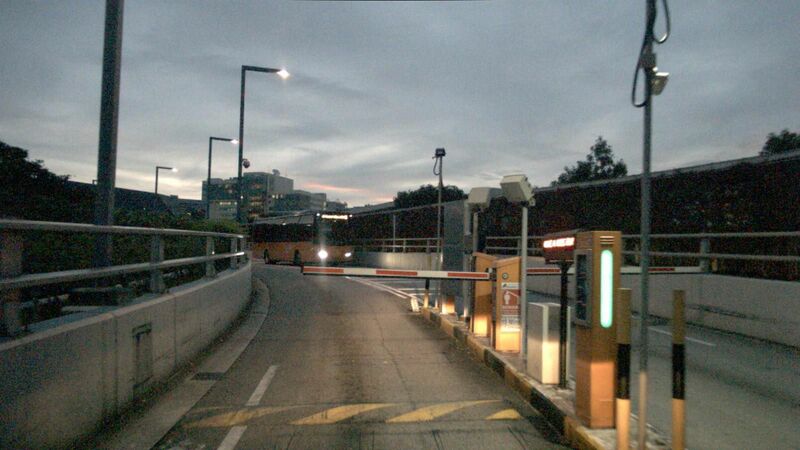} &
        \includegraphics[height=0.125\textwidth]{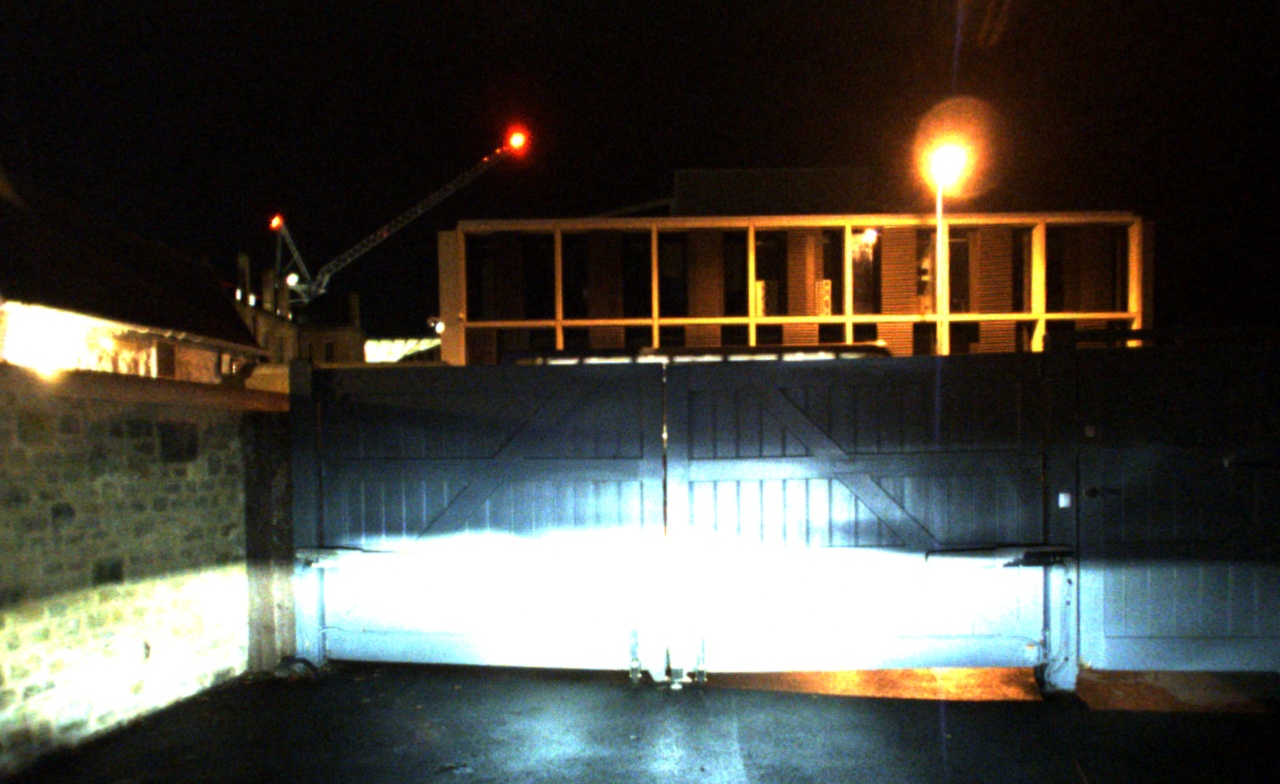} \\
        
        \includegraphics[trim=1.5cm 0cm 1.5cm 0cm,clip,height=0.125\textwidth]{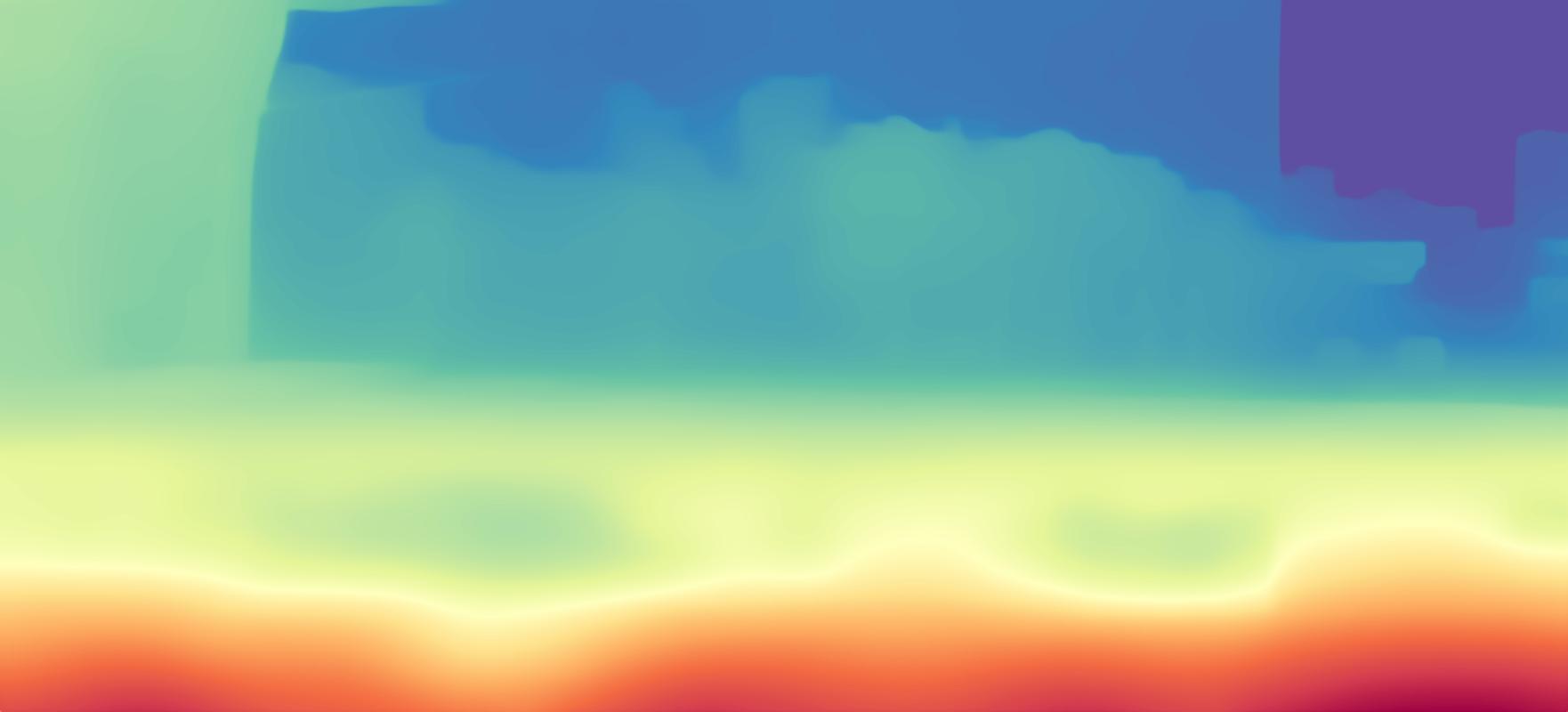} &
        \includegraphics[height=0.125\textwidth]{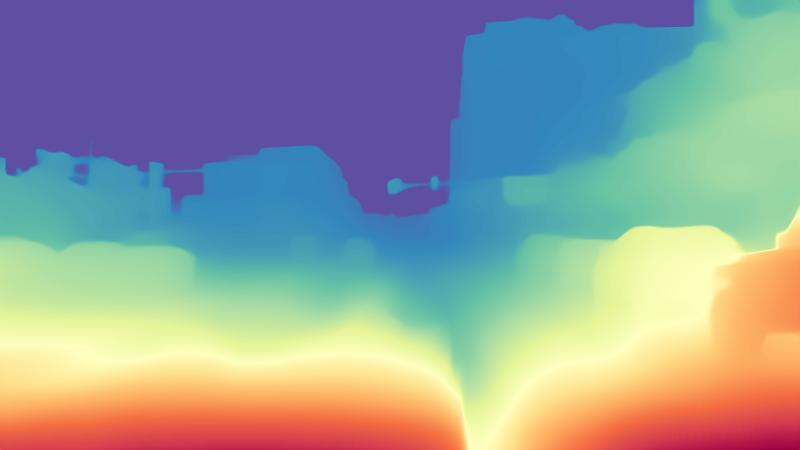} &
        \includegraphics[height=0.125\textwidth]{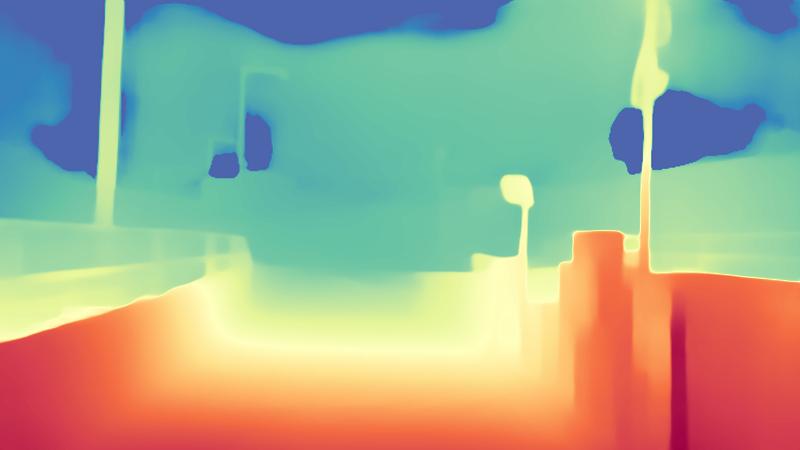} &
        \includegraphics[height=0.125\textwidth]{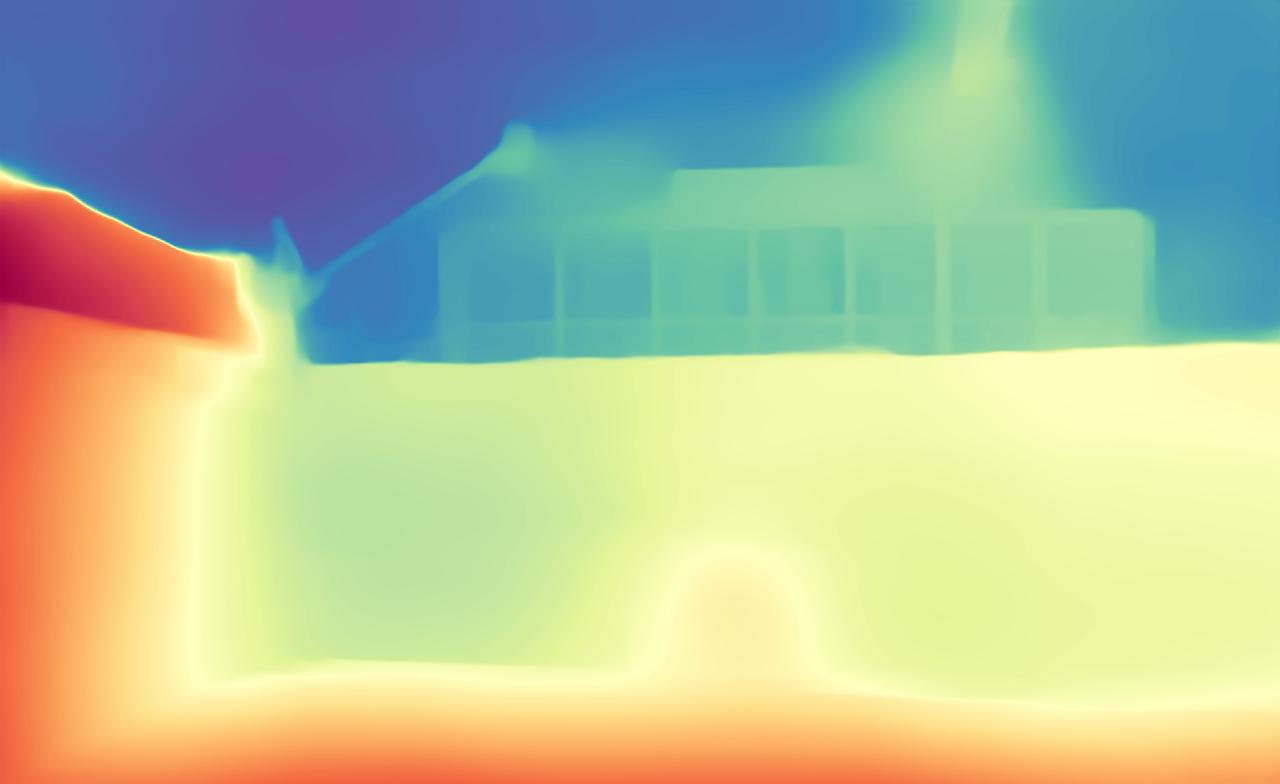} \\

        \includegraphics[trim=1.5cm 0cm 1.5cm 0cm,clip,height=0.125\textwidth]{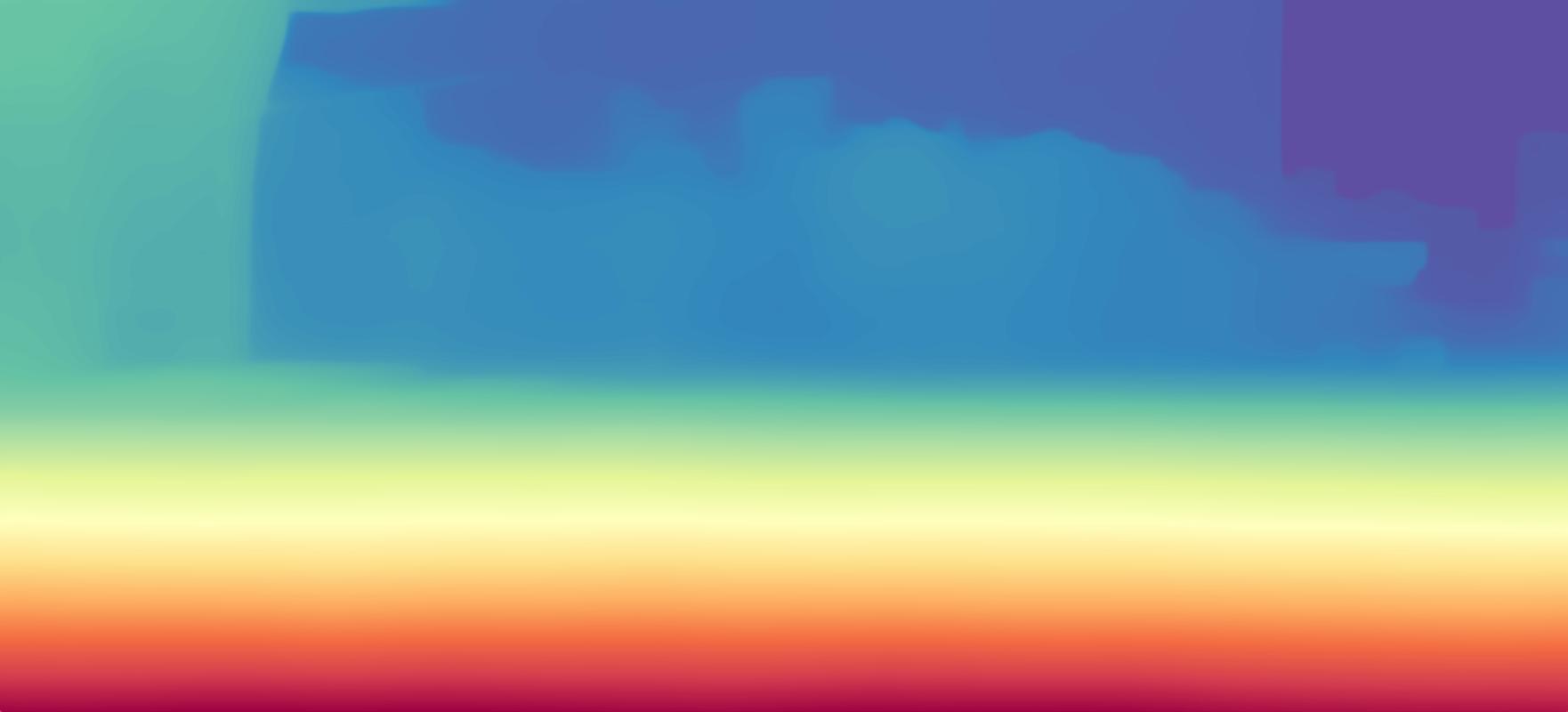} &
        \includegraphics[height=0.125\textwidth]{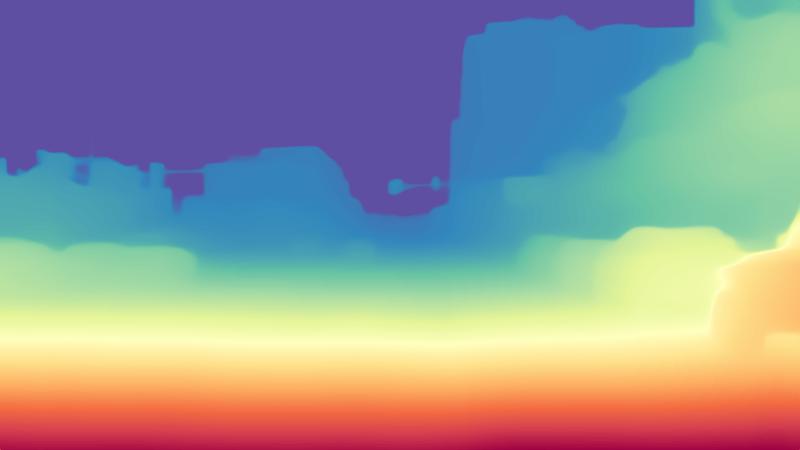} &
        \includegraphics[height=0.125\textwidth]{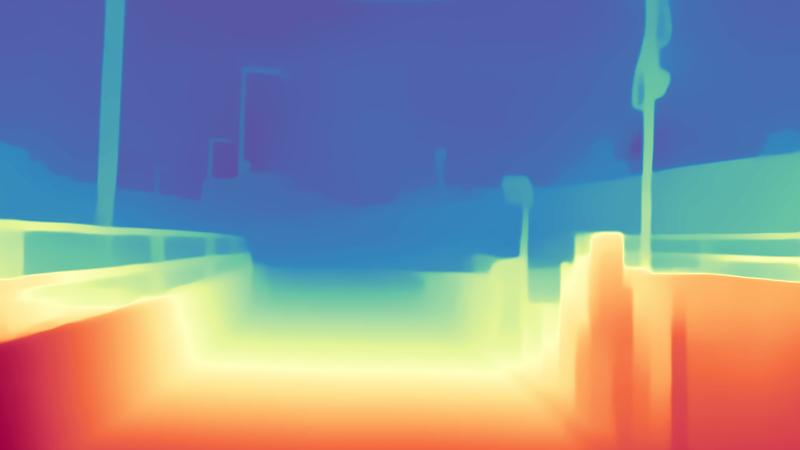} &
        \includegraphics[height=0.125\textwidth]{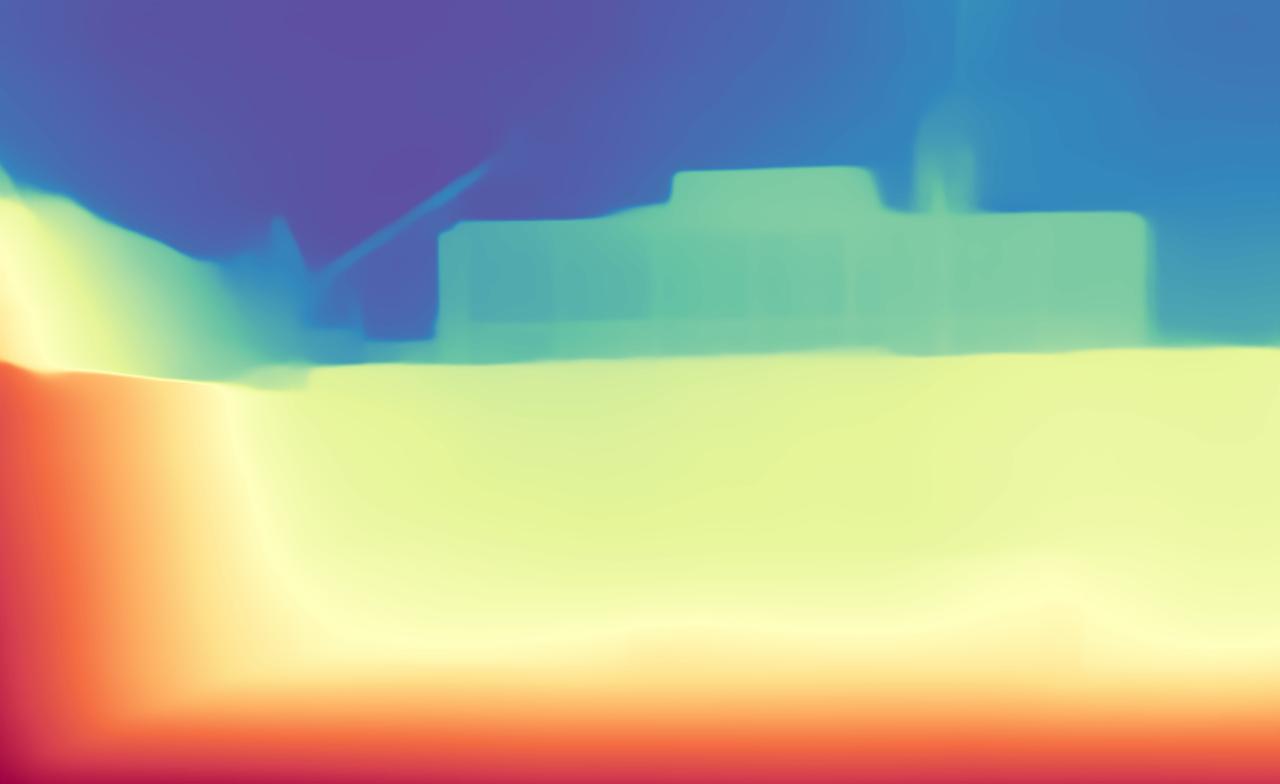} \\
        \multicolumn{2}{c}{\scriptsize Depth Anything \cite{yang2024depth}} & \multicolumn{2}{c}{\scriptsize DPT \cite{Ranftl2021}} \\
        
    \end{tabular}
    \caption{\textbf{Qualitative Results.} 
    From top to bottom: RGB images, depth maps predicted by the original models and the fine-tuned versions using our method.}
    \label{fig:DPT_large_generalization}
\end{figure}

\textbf{Method Applicability with Daytime-only Datasets.} In this experiment, we intend to demonstrate our distinct capability of generating adverse images from datasets that provide \textit{easy} images only. To prove this, in \cref{tab:generalization_experiment}, we present the results obtained by fine-tuning DPT and Depth Anything networks using five different datasets (Mapillary \cite{neuhold2017mapillary}, Cityscapes \cite{Cordts2016Cityscapes}, KITTI 2012 \cite{Geiger2012CVPR}, KITTI 2015 \cite{Menze2015CVPR}, and Apolloscapes \cite{huang2019apolloscape}, totaling approximately 33K images). These datasets contain only \textit{easy} conditions and are employed specifically for testing the ability to generalize across \textit{challenging} conditions on the DrivingStereo \cite{yang2019drivingstereo}, nuScenes \cite{caesar2020nuscenes}, and RobotCar \cite{maddern20171} datasets, without prior exposure to them. 
Importantly, other methodologies, such as \cite{gasperini_morbitzer2023md4all}, cannot be applied in this scenario due to the absence of both the \textit{easy} and \textit{challenging} images. 
Our methodology significantly enhances baseline results, as also evident in Fig. \ref{fig:DPT_large_generalization}.

\begin{table*}[t]
\centering
\caption{\textbf{Fine-tuning for ToM objects.} Results on Booster \cite{zamaramirez2022booster} train set and ClearGrasp \cite{sajjan2020clear}, at quarter and full resolution respectively.  R. ToM: a-priori availability of a real dataset for non-Lambertian objects, GT Seg.: use of ground truth segmentation masks for ToM objects during training. All models start from the official weights \cite{Ranftl2022}.}
\setlength{\tabcolsep}{2pt}
\scalebox{0.52}{
\begin{tabular}{llccccccccccccccc}

\toprule

 &&&& \multicolumn{6}{c}{Booster \cite{zamaramirez2022booster}} & \multicolumn{6}{c}{ClearGrasp \cite{sajjan2020clear}} \\
  \cmidrule(lr){5-10} \cmidrule(lr){11-16}
 Category & Method &  R. ToM & GT Seg. & $\delta$ $<$ 1.25 &  $\delta$ $<$ 1.15 & $\delta$ $<$ 1.05 & MAE & absRel & RMSE & $\delta$ $<$ 1.25 & $\delta$ $<$ 1.15 & $\delta$ $<$ 1.05 & MAE & absRel & RMSE \\
 &&&& $\uparrow$ (\%) & $\uparrow$ (\%) & $\uparrow$ (\%) & $\downarrow$ (mm) & $\downarrow$ & $\downarrow$ (mm) & $\uparrow$ (\%) & $\uparrow$ (\%) & $\uparrow$ (\%) & $\downarrow$ (mm) & $\downarrow$ & $\downarrow$ (mm) \\
 \midrule

 \multicolumn{16}{c}{DPT-Large \cite{Ranftl2021}} \\

 \hline
 
  \textbf{All} & DPT \cite{Ranftl2022} (\textbf{baseline}) &  &  & 96.79 & 89.71 & 56.26 & 75.35 & 0.06 & 100.68 & 98.71 & 94.68 & 64.95 & 32.77 & 0.05 & 45.31 \\

 \rowcolor{salmon} & ft. Ours &  &  & \textbf{98.23} & \textbf{93.66} & \textbf{60.90} & 65.29 & \textbf{0.05} & \textbf{85.48} & 98.61 & 95.67 & 66.23 & 30.64 & 0.05 & \textbf{41.71} \\

  & ft. Costanzino et al. \cite{costanzino2023iccv} & \checkmark & \checkmark & 97.99 & 93.55 & 60.46 & \textbf{64.93} & \textbf{0.05} & 85.93 & \textbf{98.97} & \textbf{96.29} & \textbf{66.30} & \textbf{30.59} & \textbf{0.04} & 41.75 \\
 
 \midrule
  \textbf{ToM} & DPT \cite{Ranftl2022} (\textbf{baseline}) &  &  & 92.77 & 80.98 & 37.70 & 113.14 & 0.10 & 136.28 & 96.50 & 87.39 & 45.63 & 41.04 & 0.07 & 47.85 \\

 \rowcolor{salmon} & ft. Ours &  &  & 96.17 & \textbf{92.54} & 52.88 & 79.64 & 0.07 & 92.56 & \textbf{98.38} & \textbf{93.69} & 57.15 & \textbf{31.32} & 0.06 & \textbf{37.41} \\

  & ft. Costanzino et al. \cite{costanzino2023iccv} & \checkmark & \checkmark & \textbf{96.68} & 92.23 & \textbf{54.67} & \textbf{70.68} & \textbf{0.06} & \textbf{83.06} & 97.46 & 92.85 & \textbf{58.53} & 31.55 & \textbf{0.05} & 37.45 \\

 \midrule
  \textbf{Other} & DPT \cite{Ranftl2022} (\textbf{baseline}) &  &  &  97.10 &  90.08  & 57.31 & 73.19 & 0.06 & 95.63 & 98.86 & 95.17 & 66.10 & 32.25 & 0.05 & 44.42 \\

 \rowcolor{salmon} & ft. Ours  &  &  & \textbf{98.35} & \textbf{96.79} & 61.14 & 65.17 & \textbf{0.05} & \textbf{85.08} & 98.65 & 95.83 & 66.79 & 30.58 & 0.05 & \textbf{41.42} \\

  & ft. Costanzino et al. \cite{costanzino2023iccv} & \checkmark& \checkmark & 98.07 & 93.52 & \textbf{61.19} & \textbf{64.70} & \textbf{0.05} & 85.57 & \textbf{99.05} &  \textbf{96.50} & \textbf{66.81} & \textbf{30.54} & \textbf{0.04} & 41.62 \\

 \bottomrule 

 \multicolumn{16}{c}{Depth Anything \cite{yang2024depth}} \\

 \hline

\textbf{All} & Depth Anything \cite{yang2024depth} (\textbf{baseline}) &  &  & 97.87 & 93.69 & 69.47 & 59.43 & 0.05 & 84.63 & 98.75 & 96.76 & 78.23 & 24.15 & 0.04 & 36.27 \\
   
 \rowcolor{salmon} & ft. Ours  &  & & \textbf{99.44} & \textbf{97.18} & \textbf{76.44} & \textbf{41.50} & \textbf{0.03} & \textbf{56.78} & \textbf{99.89} & \textbf{99.16} & \textbf{79.13} & \textbf{19.73} & \textbf{0.03} & \textbf{26.53} \\
 
 \midrule
  \textbf{ToM} &  Depth Anything \cite{yang2024depth} (\textbf{baseline}) &  &  & 84.23 & 71.10  & 39.94 & 137.96 & 0.13 & 162.62 & 83.46 & 59.26 & 15.84 & 82.22 & 0.15 & 91.88\\
 \rowcolor{salmon} & ft. Ours  &  &  & \textbf{98.91}  & \textbf{93.47} & \textbf{63.04} & \textbf{54.31} & \textbf{0.05} & \textbf{71.51} & \textbf{99.23} & \textbf{94.32} & \textbf{50.65} & \textbf{33.88} & \textbf{0.06} & \textbf{39.71} \\

 \midrule
  \textbf{Other} & Depth Anything \cite{yang2024depth} (\textbf{baseline}) &  &  & 99.05 & 95.48 & 71.90 & 52.13 & 0.04 & 70.14 & 99.74 & 98.92 & 81.34  & 21.08 & \textbf{0.03} & 29.57\\ 
 \rowcolor{salmon}  & ft. Ours  &  &  & \textbf{99.44} & \textbf{97.53} & \textbf{77.33} & \textbf{40.57} & \textbf{0.03} & \textbf{54.46} & \textbf{99.93} & \textbf{99.41} & \textbf{80.72} & \textbf{19.03} & \textbf{0.03} & \textbf{25.17} \\

 \bottomrule 
 \end{tabular}
 }
 \label{tab:ToM}
\end{table*}

\subsection{Challenging Materials}
We now assess the effectiveness of our approach in handling non-Lambertian materials.
Collecting many \textit{easy} samples in this scenario would be complex, requiring the manual setup of scenes with only Lambertian objects. As such, we recall that \cite{gasperini_morbitzer2023md4all}, based on a style-transfer network, is unable to handle this scenario.
To this aim, as described in \cref{sec:diffusion_data}, we first generate \textit{easy} samples using Stable Diffusion \cite{stable-diffusion-xl} from text prompts, and then convert them to challenging ones using \cite{mou2023t2i}.
In \cref{tab:ToM}, we report the performance of DPT \cite{Ranftl2022} with official weights (Baseline) with DPT fine-tuned according to \cite{costanzino2023iccv} (Depth4Tom) or with our approach (Ours). We fine-tune the networks using the framework from \cite{costanzino2023iccv}. While this latter is trained on 
real data featuring transparent or mirror objects \cite{xie2020segmenting,Yang_2019_ICCV}  
with ground truth segmentation maps, our method relies solely on images generated from text prompts. We test the \emph{generalization} performance of all methods on the Booster \cite{zamaramirez2022booster} and ClearGrasp \cite{sajjan2020clear} datasets, following \cite{costanzino2023iccv}.

Additionally, we conducted experiments with Depth Anything, one of the latest and most accurate state-of-the-art monocular depth estimation models, to verify the effectiveness of our method.
As shown in \cref{tab:ToM}, our approach enhances the baseline networks' performance on ToM surfaces across both datasets.
Our method achieves results comparable to \cite{costanzino2023iccv} on both the Booster and ClearGrasp datasets, with only slight variations in performance. Notably, while \cite{costanzino2023iccv} relies on real-world images and manually annotated masks for ToM surfaces, our approach utilizes only text prompts. We argue that collecting a curated dataset with annotations for each setup would be costly and time-consuming. In contrast, our text-prompt-based image generation offers an efficient, cost-effective alternative.  Moreover, we believe that this approach sets a precedent for wider future adoption and scalability in various scenarios.
Lastly, our technique demonstrates superior versatility, adapting to any challenging setting without modifications, whereas Depth4Tom \cite{costanzino2023iccv} is confined to mirrors and glasses.
\section{Conclusion}

In this work, we have introduced a pioneering training paradigm for monocular depth estimation that leverages diffusion models to address out-of-distribution scenarios.  By transforming easy samples into complex ones, we generate diverse data that captures real-world challenges. Our fine-tuning protocol enhances the robustness and generalization capabilities of existing depth networks, enabling them to handle adverse weather and non-Lambertian surfaces without domain-specific data. Extensive experiments across multiple datasets and state-of-the-art architectures demonstrate the effectiveness and versatility of our approach.

\newpage

\noindent\scriptsize{\textbf{Acknowledgements.} 

This study was funded by the European Union – Next Generation EU within the framework of the National Recovery and Resilience Plan NRRP – Mission 4 "Education and Research" – Component 2 - Investment 1.1 "National Research Program and Projects of Significant National Interest Fund (PRIN)" (Call D.D. MUR n. 104/2022) – PRIN2022 – Project reference: "RiverWatch: a citizen-science approach to river pollution monitoring" (ID: 2022MMBA8X, CUP: J53D23002260006).

It was carried out also within the MOST – Sustainable Mobility National Research Center and received funding from the European Union Next Generation EU within the framework of the National Recovery and Resilience Plan NRRP – Mission 4 "Education and Research" – Component 2 - Investment 1.4 (D.D. 1033 17/06/2022, CN00000023). This manuscript reflects only the authors’ views and opinions, neither the European Union nor the European Commission can be considered responsible for them.
}

\bibliographystyle{splncs04}
\bibliography{reference}

\begin{thebibliography}{100}
\providecommand{\url}[1]{\texttt{#1}}
\providecommand{\urlprefix}{URL }
\providecommand{\doi}[1]{https://doi.org/#1}

\bibitem{stable-diffusion-v1-5}
Stable diffusion v1.5 model card (2022), \url{https://huggingface.co/runwayml/stable-diffusion-v1-5}

\bibitem{stable-diffusion-xl}
Stable diffusion xl - sdxl 1.0 model card (2023), \url{https://huggingface.co/stabilityai/stable-diffusion-xl-base-1.0}

\bibitem{alembics-disco-diffusion}
Alembics: Disco diffusion (2022), \url{https://github.com/alembics/disco-diffusion}

\bibitem{aleotti2018generative}
Aleotti, F., Tosi, F., Poggi, M., Mattoccia, S.: Generative adversarial networks for unsupervised monocular depth prediction. In: Proceedings of the European Conference on Computer Vision (ECCV) Workshops. pp.~0--0 (2018)

\bibitem{avrahami2023spatext}
Avrahami, O., Hayes, T., Gafni, O., Gupta, S., Taigman, Y., Parikh, D., Lischinski, D., Fried, O., Yin, X.: Spatext: Spatio-textual representation for controllable image generation. In: Proceedings of the IEEE/CVF Conference on Computer Vision and Pattern Recognition. pp. 18370--18380 (2023)

\bibitem{avrahami2022blended}
Avrahami, O., Lischinski, D., Fried, O.: Blended diffusion for text-driven editing of natural images. In: Proceedings of the IEEE/CVF Conference on Computer Vision and Pattern Recognition. pp. 18208--18218 (2022)

\bibitem{bar2023multidiffusion}
Bar-Tal, O., Yariv, L., Lipman, Y., Dekel, T.: Multidiffusion: Fusing diffusion paths for controlled image generation  (2023)

\bibitem{bashkirova2023masksketch}
Bashkirova, D., Lezama, J., Sohn, K., Saenko, K., Essa, I.: Masksketch: Unpaired structure-guided masked image generation. In: Proceedings of the IEEE/CVF Conference on Computer Vision and Pattern Recognition. pp. 1879--1889 (2023)

\bibitem{bhat2021adabins}
Bhat, S.F., Alhashim, I., Wonka, P.: Adabins: Depth estimation using adaptive bins. In: Proceedings of the IEEE/CVF Conference on Computer Vision and Pattern Recognition. pp. 4009--4018 (2021)

\bibitem{bhat2023zoedepth}
Bhat, S.F., Birkl, R., Wofk, D., Wonka, P., M{\"u}ller, M.: Zoedepth: Zero-shot transfer by combining relative and metric depth. arXiv preprint arXiv:2302.12288  (2023)

\bibitem{bian2019unsupervised}
Bian, J., Li, Z., Wang, N., Zhan, H., Shen, C., Cheng, M.M., Reid, I.: Unsupervised scale-consistent depth and ego-motion learning from monocular video. Advances in neural information processing systems  \textbf{32} (2019)

\bibitem{brooks2023instructpix2pix}
Brooks, T., Holynski, A., Efros, A.A.: Instructpix2pix: Learning to follow image editing instructions. In: Proceedings of the IEEE/CVF Conference on Computer Vision and Pattern Recognition. pp. 18392--18402 (2023)

\bibitem{caesar2020nuscenes}
Caesar, H., Bankiti, V., Lang, A.H., Vora, S., Liong, V.E., Xu, Q., Krishnan, A., Pan, Y., Baldan, G., Beijbom, O.: nuscenes: A multimodal dataset for autonomous driving. In: Proceedings of the IEEE/CVF conference on computer vision and pattern recognition. pp. 11621--11631 (2020)

\bibitem{casser2019depth}
Casser, V., Pirk, S., Mahjourian, R., Angelova, A.: Depth prediction without the sensors: Leveraging structure for unsupervised learning from monocular videos. In: Proceedings of the AAAI conference on artificial intelligence. vol.~33, pp. 8001--8008 (2019)

\bibitem{chen2021attention}
Chen, Y., Zhao, H., Hu, Z., Peng, J.: Attention-based context aggregation network for monocular depth estimation. International Journal of Machine Learning and Cybernetics  \textbf{12},  1583--1596 (2021)

\bibitem{choi2021adaptive}
Choi, H., Lee, H., Kim, S., Kim, S., Kim, S., Sohn, K., Min, D.: Adaptive confidence thresholding for monocular depth estimation. In: Proceedings of the IEEE/CVF International Conference on Computer Vision. pp. 12808--12818 (2021)

\bibitem{Cordts2016Cityscapes}
Cordts, M., Omran, M., Ramos, S., Rehfeld, T., Enzweiler, M., Benenson, R., Franke, U., Roth, S., Schiele, B.: The cityscapes dataset for semantic urban scene understanding. In: Proc. of the IEEE Conference on Computer Vision and Pattern Recognition (CVPR) (2016)

\bibitem{costanzino2023iccv}
Costanzino, A., Zama~Ramirez, P., Poggi, M., Tosi, F., Mattoccia, S., Di~Stefano, L.: Learning depth estimation for transparent and mirror surfaces. In: The IEEE International Conference on Computer Vision (2023), iCCV

\bibitem{creswell2018generative}
Creswell, A., White, T., Dumoulin, V., Arulkumaran, K., Sengupta, B., Bharath, A.A.: Generative adversarial networks: An overview. IEEE signal processing magazine  \textbf{35}(1),  53--65 (2018)

\bibitem{dhariwal2021diffusion}
Dhariwal, P., Nichol, A.: Diffusion models beat gans on image synthesis. Advances in neural information processing systems  \textbf{34},  8780--8794 (2021)

\bibitem{eftekhar2021omnidata}
Eftekhar, A., Sax, A., Malik, J., Zamir, A.: Omnidata: A scalable pipeline for making multi-task mid-level vision datasets from 3d scans. In: Proceedings of the IEEE/CVF International Conference on Computer Vision. pp. 10786--10796 (2021)

\bibitem{eigen2015predicting}
Eigen, D., Fergus, R.: Predicting depth, surface normals and semantic labels with a common multi-scale convolutional architecture. In: Proceedings of the IEEE international conference on computer vision. pp. 2650--2658 (2015)

\bibitem{eigen2014depth}
Eigen, D., Puhrsch, C., Fergus, R.: Depth map prediction from a single image using a multi-scale deep network. Advances in neural information processing systems  \textbf{27} (2014)

\bibitem{gafni2022make}
Gafni, O., Polyak, A., Ashual, O., Sheynin, S., Parikh, D., Taigman, Y.: Make-a-scene: Scene-based text-to-image generation with human priors. In: European Conference on Computer Vision. pp. 89--106. Springer (2022)

\bibitem{garg2016unsupervised}
Garg, R., Bg, V.K., Carneiro, G., Reid, I.: Unsupervised cnn for single view depth estimation: Geometry to the rescue. In: Computer Vision--ECCV 2016: 14th European Conference, Amsterdam, The Netherlands, October 11-14, 2016, Proceedings, Part VIII 14. pp. 740--756. Springer (2016)

\bibitem{gasperini2021r4dyn}
Gasperini, S., Koch, P., Dallabetta, V., Navab, N., Busam, B., Tombari, F.: R4dyn: Exploring radar for self-supervised monocular depth estimation of dynamic scenes. In: 2021 International Conference on 3D Vision (3DV). pp. 751--760. IEEE (2021)

\bibitem{gasperini_morbitzer2023md4all}
Gasperini, S., Morbitzer, N., Jung, H., Navab, N., Tombari, F.: Robust monocular depth estimation under challenging conditions. In: Proceedings of the IEEE/CVF International Conference on Computer Vision (2023)

\bibitem{Geiger2012CVPR}
Geiger, A., Lenz, P., Urtasun, R.: Are we ready for autonomous driving? the kitti vision benchmark suite. In: Conference on Computer Vision and Pattern Recognition (CVPR) (2012)

\bibitem{monodepth17}
Godard, C., {Mac Aodha}, O., Brostow, G.J.: Unsupervised monocular depth estimation with left-right consistency. In: CVPR (2017)

\bibitem{monodepth2}
Godard, C., {Mac Aodha}, O., Firman, M., Brostow, G.J.: Digging into self-supervised monocular depth prediction. In: The International Conference on Computer Vision (ICCV) (October 2019)

\bibitem{gordon2019depth}
Gordon, A., Li, H., Jonschkowski, R., Angelova, A.: Depth from videos in the wild: Unsupervised monocular depth learning from unknown cameras. In: Proceedings of the IEEE/CVF International Conference on Computer Vision. pp. 8977--8986 (2019)

\bibitem{guizilini20203d}
Guizilini, V., Ambrus, R., Pillai, S., Raventos, A., Gaidon, A.: 3d packing for self-supervised monocular depth estimation. In: Proceedings of the IEEE/CVF conference on computer vision and pattern recognition. pp. 2485--2494 (2020)

\bibitem{guizilini2020semantically}
Guizilini, V., Hou, R., Li, J., Ambrus, R., Gaidon, A.: Semantically-guided representation learning for self-supervised monocular depth. arXiv preprint arXiv:2002.12319  (2020)

\bibitem{guizilini2023towards}
Guizilini, V., Vasiljevic, I., Chen, D., Ambruș, R., Gaidon, A.: Towards zero-shot scale-aware monocular depth estimation. In: Proceedings of the IEEE/CVF International Conference on Computer Vision. pp. 9233--9243 (2023)

\bibitem{hertz2022prompt}
Hertz, A., Mokady, R., Tenenbaum, J., Aberman, K., Pritch, Y., Cohen-Or, D.: Prompt-to-prompt image editing with cross attention control. arXiv preprint arXiv:2208.01626  (2022)

\bibitem{hoiem2005automatic}
Hoiem, D., Efros, A.A., Hebert, M.: Automatic photo pop-up. In: ACM SIGGRAPH 2005 Papers, pp. 577--584 (2005)

\bibitem{hornauer2022gradient}
Hornauer, J., Belagiannis, V.: Gradient-based uncertainty for monocular depth estimation. In: European Conference on Computer Vision. pp. 613--630. Springer (2022)

\bibitem{hu2023cocktail}
Hu, M., Zheng, J., Liu, D., Zheng, C., Wang, C., Tao, D., Cham, T.J.: Cocktail: Mixing multi-modality control for text-conditional image generation. In: Thirty-seventh Conference on Neural Information Processing Systems (2023)

\bibitem{huang2023composer}
Huang, L., Chen, D., Liu, Y., Shen, Y., Zhao, D., Zhou, J.: Composer: Creative and controllable image synthesis with composable conditions. arXiv preprint arXiv:2302.09778  (2023)

\bibitem{huang2019apolloscape}
Huang, X., Wang, P., Cheng, X., Zhou, D., Geng, Q., Yang, R.: The apolloscape open dataset for autonomous driving and its application. IEEE transactions on pattern analysis and machine intelligence  \textbf{42}(10),  2702--2719 (2019)

\bibitem{kawar2023imagic}
Kawar, B., Zada, S., Lang, O., Tov, O., Chang, H., Dekel, T., Mosseri, I., Irani, M.: Imagic: Text-based real image editing with diffusion models. In: Proceedings of the IEEE/CVF Conference on Computer Vision and Pattern Recognition. pp. 6007--6017 (2023)

\bibitem{ke2024repurposing}
Ke, B., Obukhov, A., Huang, S., Metzger, N., Daudt, R.C., Schindler, K.: Repurposing diffusion-based image generators for monocular depth estimation. In: Proceedings of the IEEE/CVF Conference on Computer Vision and Pattern Recognition. pp. 9492--9502 (2024)

\bibitem{kim2022diffusionclip}
Kim, G., Kwon, T., Ye, J.C.: Diffusionclip: Text-guided diffusion models for robust image manipulation. In: Proceedings of the IEEE/CVF Conference on Computer Vision and Pattern Recognition. pp. 2426--2435 (2022)

\bibitem{kingma2021variational}
Kingma, D., Salimans, T., Poole, B., Ho, J.: Variational diffusion models. Advances in neural information processing systems  \textbf{34},  21696--21707 (2021)

\bibitem{klingner2020self}
Klingner, M., Term{\"o}hlen, J.A., Mikolajczyk, J., Fingscheidt, T.: Self-supervised monocular depth estimation: Solving the dynamic object problem by semantic guidance. In: Computer Vision--ECCV 2020: 16th European Conference, Glasgow, UK, August 23--28, 2020, Proceedings, Part XX 16. pp. 582--600. Springer (2020)

\bibitem{lee2019big}
Lee, J.H., Han, M.K., Ko, D.W., Suh, I.H.: From big to small: Multi-scale local planar guidance for monocular depth estimation. arXiv preprint arXiv:1907.10326  (2019)

\bibitem{li2015depth}
Li, B., Shen, C., Dai, Y., Van Den~Hengel, A., He, M.: Depth and surface normal estimation from monocular images using regression on deep features and hierarchical crfs. In: Proceedings of the IEEE conference on computer vision and pattern recognition. pp. 1119--1127 (2015)

\bibitem{liu2015learning}
Liu, F., Shen, C., Lin, G., Reid, I.: Learning depth from single monocular images using deep convolutional neural fields. IEEE transactions on pattern analysis and machine intelligence  \textbf{38}(10),  2024--2039 (2015)

\bibitem{liu2021self}
Liu, L., Song, X., Wang, M., Liu, Y., Zhang, L.: Self-supervised monocular depth estimation for all day images using domain separation. In: Proceedings of the IEEE/CVF International Conference on Computer Vision. pp. 12737--12746 (2021)

\bibitem{loshchilov2017decoupled}
Loshchilov, I., Hutter, F.: Decoupled weight decay regularization. arXiv preprint arXiv:1711.05101  (2017)

\bibitem{luo2018single}
Luo, Y., Ren, J., Lin, M., Pang, J., Sun, W., Li, H., Lin, L.: Single view stereo matching. In: Proceedings of the IEEE Conference on Computer Vision and Pattern Recognition. pp. 155--163 (2018)

\bibitem{maddern20171}
Maddern, W., Pascoe, G., Linegar, C., Newman, P.: 1 year, 1000 km: The oxford robotcar dataset. The International Journal of Robotics Research  \textbf{36}(1),  3--15 (2017)

\bibitem{mahjourian2018unsupervised}
Mahjourian, R., Wicke, M., Angelova, A.: Unsupervised learning of depth and ego-motion from monocular video using 3d geometric constraints. In: Proceedings of the IEEE conference on computer vision and pattern recognition. pp. 5667--5675 (2018)

\bibitem{Menze2015CVPR}
Menze, M., Geiger, A.: Object scene flow for autonomous vehicles. In: Conference on Computer Vision and Pattern Recognition (CVPR) (2015)

\bibitem{mou2024t2i}
Mou, C., Wang, X., Xie, L., Wu, Y., Zhang, J., Qi, Z., Shan, Y.: T2i-adapter: Learning adapters to dig out more controllable ability for text-to-image diffusion models. In: Proceedings of the AAAI Conference on Artificial Intelligence. vol.~38, pp. 4296--4304 (2024)

\bibitem{mou2023t2i}
Mou, C., Wang, X., Xie, L., Zhang, J., Qi, Z., Shan, Y., Qie, X.: T2i-adapter: Learning adapters to dig out more controllable ability for text-to-image diffusion models. arXiv preprint arXiv:2302.08453  (2023)

\bibitem{neuhold2017mapillary}
Neuhold, G., Ollmann, T., Rota~Bulo, S., Kontschieder, P.: The mapillary vistas dataset for semantic understanding of street scenes. In: Proceedings of the IEEE international conference on computer vision. pp. 4990--4999 (2017)

\bibitem{nichol2021glide}
Nichol, A., Dhariwal, P., Ramesh, A., Shyam, P., Mishkin, P., McGrew, B., Sutskever, I., Chen, M.: Glide: Towards photorealistic image generation and editing with text-guided diffusion models. arXiv preprint arXiv:2112.10741  (2021)

\bibitem{openai-dall-e-2}
OpenAI: Dall-e 2 (2023), \url{https://openai.com/product/dall-e-2}

\bibitem{parmar2023zero}
Parmar, G., Kumar~Singh, K., Zhang, R., Li, Y., Lu, J., Zhu, J.Y.: Zero-shot image-to-image translation. In: ACM SIGGRAPH 2023 Conference Proceedings. pp. 1--11 (2023)

\bibitem{patil2022p3depth}
Patil, V., Sakaridis, C., Liniger, A., Van~Gool, L.: P3depth: Monocular depth estimation with a piecewise planarity prior. In: Proceedings of the IEEE/CVF Conference on Computer Vision and Pattern Recognition. pp. 1610--1621 (2022)

\bibitem{peng2021excavating}
Peng, R., Wang, R., Lai, Y., Tang, L., Cai, Y.: Excavating the potential capacity of self-supervised monocular depth estimation. In: Proceedings of the IEEE/CVF International Conference on Computer Vision. pp. 15560--15569 (2021)

\bibitem{pilzer2018unsupervised}
Pilzer, A., Xu, D., Puscas, M., Ricci, E., Sebe, N.: Unsupervised adversarial depth estimation using cycled generative networks. In: 2018 international conference on 3D vision (3DV). pp. 587--595. IEEE (2018)

\bibitem{poggi2020uncertainty}
Poggi, M., Aleotti, F., Tosi, F., Mattoccia, S.: On the uncertainty of self-supervised monocular depth estimation. In: Proceedings of the IEEE/CVF Conference on Computer Vision and Pattern Recognition. pp. 3227--3237 (2020)

\bibitem{poggi2018learning}
Poggi, M., Tosi, F., Mattoccia, S.: Learning monocular depth estimation with unsupervised trinocular assumptions. In: 2018 International conference on 3d vision (3DV). pp. 324--333. IEEE (2018)

\bibitem{ramesh2022hierarchical}
Ramesh, A., Dhariwal, P., Nichol, A., Chu, C., Chen, M.: Hierarchical text-conditional image generation with clip latents. arXiv preprint arXiv:2204.06125  \textbf{1}(2), ~3 (2022)

\bibitem{ramirez2023booster}
Ramirez, P.Z., Costanzino, A., Tosi, F., Poggi, M., Salti, S., Mattoccia, S., Di~Stefano, L.: Booster: a benchmark for depth from images of specular and transparent surfaces. IEEE Transactions on Pattern Analysis and Machine Intelligence  (2023)

\bibitem{Ramirez_2024_CVPR}
Ramirez, P.Z., Tosi, F., Di~Stefano, L., Timofte, R., Costanzino, A., Poggi, M., Salti, S., Mattoccia, S., Zhang, Y., Wu, C., He, Z., Yin, S., Dong, J., Liu, Y., Jiang, H., Shi, J., A, Y., Jin, Y., Li, D., Ke, B., Obukhov, A., Wang, T., Metzger, N., Huang, S., Schindler, K., Huang, Y., Li, J., Zhang, J., Wang, Y., Huang, Z., Liu, T., Cao, Z., Li, P., Wang, J.L., Zhu, W., Geng, H., Zhang, Y., Lan, L., Xu, K., Sun, T., Xu, Q., Saini, S., Gupta, A., Mistry, S.K., Shukla, A., Jakhetiya, V., Jaiswal, S., Sun, Y., Zheng, Z., Ning, Y., Cheng, J.H., Liu, H.I., Huang, H.W., Yang, C.Y., Jiang, Z., Peng, Y.H., Huang, A., Hwang, J.N.: Ntire 2024 challenge on hr depth from images of specular and transparent surfaces. In: Proceedings of the IEEE/CVF Conference on Computer Vision and Pattern Recognition (CVPR) Workshops. pp. 6499--6512 (June 2024)

\bibitem{Ranftl2021}
Ranftl, R., Bochkovskiy, A., Koltun, V.: Vision transformers for dense prediction. ICCV  (2021)

\bibitem{Ranftl2022}
Ranftl, R., Lasinger, K., Hafner, D., Schindler, K., Koltun, V.: Towards robust monocular depth estimation: Mixing datasets for zero-shot cross-dataset transfer. IEEE Transactions on Pattern Analysis and Machine Intelligence  \textbf{44}(3) (2022)

\bibitem{ranjan2019competitive}
Ranjan, A., Jampani, V., Balles, L., Kim, K., Sun, D., Wulff, J., Black, M.J.: Competitive collaboration: Joint unsupervised learning of depth, camera motion, optical flow and motion segmentation. In: Proceedings of the IEEE/CVF conference on computer vision and pattern recognition. pp. 12240--12249 (2019)

\bibitem{rombach2022high}
Rombach, R., Blattmann, A., Lorenz, D., Esser, P., Ommer, B.: High-resolution image synthesis with latent diffusion models. In: Proceedings of the IEEE/CVF conference on computer vision and pattern recognition. pp. 10684--10695 (2022)

\bibitem{ronneberger2015u}
Ronneberger, O., Fischer, P., Brox, T.: U-net: Convolutional networks for biomedical image segmentation. In: Medical Image Computing and Computer-Assisted Intervention--MICCAI 2015: 18th International Conference, Munich, Germany, October 5-9, 2015, Proceedings, Part III 18. pp. 234--241. Springer (2015)

\bibitem{sajjan2020clear}
Sajjan, S., Moore, M., Pan, M., Nagaraja, G., Lee, J., Zeng, A., Song, S.: Clear grasp: 3d shape estimation of transparent objects for manipulation. In: 2020 IEEE International Conference on Robotics and Automation (ICRA). pp. 3634--3642. IEEE (2020)

\bibitem{saxena2005learning}
Saxena, A., Chung, S., Ng, A.: Learning depth from single monocular images. Advances in neural information processing systems  \textbf{18} (2005)

\bibitem{saxena2008make3d}
Saxena, A., Sun, M., Ng, A.Y.: Make3d: Learning 3d scene structure from a single still image. IEEE transactions on pattern analysis and machine intelligence  \textbf{31}(5),  824--840 (2008)

\bibitem{saxena2023surprising}
Saxena, S., Herrmann, C., Hur, J., Kar, A., Norouzi, M., Sun, D., Fleet, D.J.: The surprising effectiveness of diffusion models for optical flow and monocular depth estimation. arXiv preprint arXiv:2306.01923  (2023)

\bibitem{saxena2023monocular}
Saxena, S., Kar, A., Norouzi, M., Fleet, D.J.: Monocular depth estimation using diffusion models. arXiv preprint arXiv:2302.14816  (2023)

\bibitem{sohl2015deep}
Sohl-Dickstein, J., Weiss, E., Maheswaranathan, N., Ganguli, S.: Deep unsupervised learning using nonequilibrium thermodynamics. In: International conference on machine learning. pp. 2256--2265. PMLR (2015)

\bibitem{spencer2020defeat}
Spencer, J., Bowden, R., Hadfield, S.: Defeat-net: General monocular depth via simultaneous unsupervised representation learning. In: Proceedings of the IEEE/CVF Conference on Computer Vision and Pattern Recognition. pp. 14402--14413 (2020)

\bibitem{spencer2023monocular}
Spencer, J., Qian, C.S., Russell, C., Hadfield, S., Graf, E., Adams, W., Schofield, A.J., Elder, J.H., Bowden, R., Cong, H., et~al.: The monocular depth estimation challenge. In: Proceedings of the IEEE/CVF Winter Conference on Applications of Computer Vision. pp. 623--632 (2023)

\bibitem{spencer2023second}
Spencer, J., Qian, C.S., Trescakova, M., Russell, C., Hadfield, S., Graf, E.W., Adams, W.J., Schofield, A.J., Elder, J., Bowden, R., et~al.: The second monocular depth estimation challenge. In: Proceedings of the IEEE/CVF Conference on Computer Vision and Pattern Recognition. pp. 3064--3076 (2023)

\bibitem{spencer2024third}
Spencer, J., Tosi, F., Poggi, M., Arora, R.S., Russell, C., Hadfield, S., Bowden, R., Zhou, G., Li, Z., Rao, Q., et~al.: The third monocular depth estimation challenge. In: Proceedings of the IEEE/CVF Conference on Computer Vision and Pattern Recognition. pp. 1--14 (2024)

\bibitem{sun2021unsupervised}
Sun, Q., Tang, Y., Zhang, C., Zhao, C., Qian, F., Kurths, J.: Unsupervised estimation of monocular depth and vo in dynamic environments via hybrid masks. IEEE Transactions on Neural Networks and Learning Systems  \textbf{33}(5),  2023--2033 (2021)

\bibitem{tosi2019learning}
Tosi, F., Aleotti, F., Poggi, M., Mattoccia, S.: Learning monocular depth estimation infusing traditional stereo knowledge. In: Proceedings of the IEEE/CVF Conference on Computer Vision and Pattern Recognition. pp. 9799--9809 (2019)

\bibitem{tosi2020distilled}
Tosi, F., Aleotti, F., Ramirez, P.Z., Poggi, M., Salti, S., Stefano, L.D., Mattoccia, S.: Distilled semantics for comprehensive scene understanding from videos. In: Proceedings of the IEEE/CVF conference on computer vision and pattern recognition. pp. 4654--4665 (2020)

\bibitem{vankadari2020unsupervised}
Vankadari, M., Garg, S., Majumder, A., Kumar, S., Behera, A.: Unsupervised monocular depth estimation for night-time images using adversarial domain feature adaptation. In: Computer Vision--ECCV 2020: 16th European Conference, Glasgow, UK, August 23--28, 2020, Proceedings, Part XXVIII 16. pp. 443--459. Springer (2020)

\bibitem{vankadari2023sun}
Vankadari, M., Golodetz, S., Garg, S., Shin, S., Markham, A., Trigoni, N.: When the sun goes down: Repairing photometric losses for all-day depth estimation. In: Conference on Robot Learning. pp. 1992--2003. PMLR (2023)

\bibitem{voynov2023sketch}
Voynov, A., Aberman, K., Cohen-Or, D.: Sketch-guided text-to-image diffusion models. In: ACM SIGGRAPH 2023 Conference Proceedings. pp. 1--11 (2023)

\bibitem{wang2018learning}
Wang, C., Buenaposada, J.M., Zhu, R., Lucey, S.: Learning depth from monocular videos using direct methods. In: Proceedings of the IEEE conference on computer vision and pattern recognition. pp. 2022--2030 (2018)

\bibitem{wang2021regularizing}
Wang, K., Zhang, Z., Yan, Z., Li, X., Xu, B., Li, J., Yang, J.: Regularizing nighttime weirdness: Efficient self-supervised monocular depth estimation in the dark. In: Proceedings of the IEEE/CVF International Conference on Computer Vision. pp. 16055--16064 (2021)

\bibitem{watson2019self}
Watson, J., Firman, M., Brostow, G.J., Turmukhambetov, D.: Self-supervised monocular depth hints. In: Proceedings of the IEEE/CVF International Conference on Computer Vision. pp. 2162--2171 (2019)

\bibitem{wu2022toward}
Wu, C.Y., Wang, J., Hall, M., Neumann, U., Su, S.: Toward practical monocular indoor depth estimation. In: Proceedings of the IEEE/CVF Conference on Computer Vision and Pattern Recognition. pp. 3814--3824 (2022)

\bibitem{xie2020segmenting}
Xie, E., Wang, W., Wang, W., Ding, M., Shen, C., Luo, P.: Segmenting transparent objects in the wild. arXiv preprint arXiv:2003.13948  (2020)

\bibitem{yang2019drivingstereo}
Yang, G., Song, X., Huang, C., Deng, Z., Shi, J., Zhou, B.: Drivingstereo: A large-scale dataset for stereo matching in autonomous driving scenarios. In: Proceedings of the IEEE/CVF Conference on Computer Vision and Pattern Recognition. pp. 899--908 (2019)

\bibitem{yang2024depth}
Yang, L., Kang, B., Huang, Z., Xu, X., Feng, J., Zhao, H.: Depth anything: Unleashing the power of large-scale unlabeled data. In: Proceedings of the IEEE/CVF Conference on Computer Vision and Pattern Recognition. pp. 10371--10381 (2024)

\bibitem{Yang_2019_ICCV}
Yang, X., Mei, H., Xu, K., Wei, X., Yin, B., Lau, R.W.: Where is my mirror? In: The IEEE International Conference on Computer Vision (ICCV) (October 2019)

\bibitem{yin2019enforcing}
Yin, W., Liu, Y., Shen, C., Yan, Y.: Enforcing geometric constraints of virtual normal for depth prediction. In: Proceedings of the IEEE/CVF International Conference on Computer Vision. pp. 5684--5693 (2019)

\bibitem{yin2023metric3d}
Yin, W., Zhang, C., Chen, H., Cai, Z., Yu, G., Wang, K., Chen, X., Shen, C.: Metric3d: Towards zero-shot metric 3d prediction from a single image. In: Proceedings of the IEEE/CVF International Conference on Computer Vision. pp. 9043--9053 (2023)

\bibitem{yin2018geonet}
Yin, Z., Shi, J.: Geonet: Unsupervised learning of dense depth, optical flow and camera pose. In: Proceedings of the IEEE conference on computer vision and pattern recognition. pp. 1983--1992 (2018)

\bibitem{yuan2022neural}
Yuan, W., Gu, X., Dai, Z., Zhu, S., Tan, P.: Neural window fully-connected crfs for monocular depth estimation. In: Proceedings of the IEEE/CVF Conference on Computer Vision and Pattern Recognition. pp. 3916--3925 (2022)

\bibitem{Ramirez2023b}
Zama~Ramirez, P., Fabio, T., Di~Stefano, L., Timofte, R., Costanzino, A., Poggi, M., Salti, S., Mattoccia, S., Shi, J., Zhang, D., A, Y., Jin, Y., Li, D., Li, C., Liu, Z., Zhang, Q., Wang, Y., Yin, S.: {NTIRE} 2023 challenge on {HR} depth from images of specular and transparent surfaces. In: Proceedings of the IEEE/CVF Conference on Computer Vision and Pattern Recognition Workshops (2023)

\bibitem{zama2019geometry}
Zama~Ramirez, P., Poggi, M., Tosi, F., Mattoccia, S., Di~Stefano, L.: Geometry meets semantics for semi-supervised monocular depth estimation. In: Computer Vision--ACCV 2018: 14th Asian Conference on Computer Vision, Perth, Australia, December 2--6, 2018, Revised Selected Papers, Part III 14. pp. 298--313. Springer (2019)

\bibitem{zamaramirez2022booster}
Zama~Ramirez, P., Tosi, F., Poggi, M., Salti, S., Di~Stefano, L., Mattoccia, S.: Open challenges in deep stereo: the booster dataset. In: Proceedings of the IEEE conference on computer vision and pattern recognition (2022), cVPR

\bibitem{zavadski2023controlnet}
Zavadski, D., Feiden, J.F., Rother, C.: Controlnet-xs: Designing an efficient and effective architecture for controlling text-to-image diffusion models. arXiv preprint arXiv:2312.06573  (2023)

\bibitem{zhang2023adding}
Zhang, L., Rao, A., Agrawala, M.: Adding conditional control to text-to-image diffusion models (2023)

\bibitem{zhao2020monocular}
Zhao, C., Sun, Q., Zhang, C., Tang, Y., Qian, F.: Monocular depth estimation based on deep learning: An overview. Science China Technological Sciences  \textbf{63}(9),  1612--1627 (2020)

\bibitem{zhao2022unsupervised}
Zhao, C., Tang, Y., Sun, Q.: Unsupervised monocular depth estimation in highly complex environments. IEEE Transactions on Emerging Topics in Computational Intelligence  \textbf{6}(5),  1237--1246 (2022)

\bibitem{zhao2022monovit}
Zhao, C., Zhang, Y., Poggi, M., Tosi, F., Guo, X., Zhu, Z., Huang, G., Tang, Y., Mattoccia, S.: Monovit: Self-supervised monocular depth estimation with a vision transformer. arXiv preprint arXiv:2208.03543  (2022)

\bibitem{zhao2024uni}
Zhao, S., Chen, D., Chen, Y.C., Bao, J., Hao, S., Yuan, L., Wong, K.Y.K.: Uni-controlnet: All-in-one control to text-to-image diffusion models. Advances in Neural Information Processing Systems  \textbf{36} (2024)

\bibitem{zheng2020forkgan}
Zheng, Z., Wu, Y., Han, X., Shi, J.: Forkgan: Seeing into the rainy night. In: Computer Vision--ECCV 2020: 16th European Conference, Glasgow, UK, August 23--28, 2020, Proceedings, Part III 16. pp. 155--170. Springer (2020)

\bibitem{zhou2017unsupervised}
Zhou, T., Brown, M., Snavely, N., Lowe, D.G.: Unsupervised learning of depth and ego-motion from video. In: Proceedings of the IEEE conference on computer vision and pattern recognition. pp. 1851--1858 (2017)

\bibitem{zou2018df}
Zou, Y., Luo, Z., Huang, J.B.: Df-net: Unsupervised joint learning of depth and flow using cross-task consistency. In: Proceedings of the European conference on computer vision (ECCV). pp. 36--53 (2018)

\end{thebibliography}

\end{document}